\documentclass[twoside]{article}

\usepackage{amsmath,bm}
\usepackage[capitalise]{cleveref}
\usepackage{amssymb,mathrsfs, amsthm}
\usepackage{xspace}
\usepackage{enumitem}
\usepackage[ruled]{algorithm2e}
\SetKwInput{KwInput}{Input}   
\SetKwInput{KwOutput}{Output} 
\SetKwProg{init}{Initialize}{}{}
\usepackage{lipsum}
\usepackage{xpatch}
\usepackage{mathtools}
\usepackage{pgf,tikz}
\usetikzlibrary{positioning}
\usepackage{caption}
\usepackage{graphicx}
\usepackage{textcomp} %for showing special ' in french names
\def\ntk{{k_{\mathrm{NN}}}}
\def\cntk{{k_{\mathrm{CNN}}}}

\def\mNTK{{\bm{K}_{\mathrm{NN}}}}

\def\ntkalg{{\textsc{NTK-UCB}}\xspace}
\def \supntkalg{{\textsc{SupNTK-UCB}}\xspace}
\def\nnalg{{\textsc{NN-UCB}}\xspace}
\def\supnnalg{{\textsc{SupNN-UCB}}\xspace}
\def\cntkalg{{\textsc{CNTK-UCB}}\xspace}

\def\cnnalg{{\textsc{CNN-UCB}}\xspace}
\def\supcnnalg{{\textsc{SupCNN-UCB}}\xspace}
\def\mnist{{\textsc{MNIST}}\xspace}

\def\CNN{{C(d,L)}}
\def\tCNN{{C_{\mathrm{NN}}(d,L)}}

\newtheorem{theorem}{Theorem}[section]
\newtheorem{lemma}[theorem]{Lemma}
\newtheorem{corollary}[theorem]{Corollary}
\theoremstyle{definition}

\newtheorem{assumption}[theorem]{Assumption}
\newtheorem{condition}[theorem]{Condition}
\newtheorem{prop}[theorem]{Proposition}
\theoremstyle{remark}

\newtheorem*{remark*}{Remark}

\def\1{\bm{1}}

\def\vtheta{{\bm{\theta}}}
\def\vphi{{\bm{\phi}}}
\def\va{{\bm{a}}}
\def\vb{{\bm{b}}}

\def\vf{{\bm{f}}}
\def\vg{{\bm{g}}}

\def\vs{{\bm{s}}}

\def\vv{{\bm{v}}}
\def\vw{{\bm{w}}}
\def\vx{{\bm{x}}}
\def\vy{{\bm{y}}}
\def\vz{{\bm{z}}}

\def\mA{{\bm{A}}}

\def\mG{{\bm{G}}}
\def\mH{{\bm{H}}}
\def\mI{{\bm{I}}}

\def\mK{{\bm{K}}}

\def\mP{{\bm{P}}}
\def\mQ{{\bm{Q}}}

\def\mW{{\bm{W}}}
\def\mX{{\bm{X}}}

\def\mZ{{\bm{Z}}}

\DeclareMathAlphabet{\mathsfit}{\encodingdefault}{\sfdefault}{m}{sl}
\SetMathAlphabet{\mathsfit}{bold}{\encodingdefault}{\sfdefault}{bx}{n}

\def\gA{{\mathcal{A}}}

\def\gC{{\mathcal{C}}}

\def\gF{{\mathcal{F}}}

\def\gH{{\mathcal{H}}}

\def\gL{{\mathcal{L}}}

\def\gO{{\mathcal{O}}}

\def\sR{{\mathbb{R}}}
\def\sS{{\mathbb{S}}}

\newcommand{\Ls}{\mathcal{L}}
\newcommand{\R}{\mathbb{R}}

\newcommand{\Cov}{\mathrm{Cov}}
\DeclareMathOperator*{\argmax}{arg\,max}

\DeclarePairedDelimiter\abs{\lvert}{\rvert}%
\DeclarePairedDelimiter\norm{\lVert}{\rVert}%

\makeatletter
\let\oldabs\abs
\def\abs{\@ifstar{\oldabs}{\oldabs*}}
\let\oldnorm\norm
\def\norm{\@ifstar{\oldnorm}{\oldnorm*}}
\makeatother

\makeatletter
\newcommand{\pushright}[1]{\ifmeasuring@#1\else\omit\hfill$\displaystyle#1$\fi\ignorespaces}
\newcommand{\pushleft}[1]{\ifmeasuring@#1\else\omit$\displaystyle#1$\hfill\fi\ignorespaces}
\makeatother

\usepackage[accepted]{aistats2022}
\usepackage[round]{natbib}

\begin{document}

% If your paper is accepted and the title of your paper is very long,
% the style will print as headings an error message. Use the following
% command to supply a shorter title of your paper so that it can be
% used as headings.
%
%\runningtitle{I use this title instead because the last one was very long}

% If your paper is accepted and the number of authors is large, the
% style will print as headings an error message. Use the following
% command to supply a shorter version of the authors names so that
% they can be used as headings (for example, use only the surnames)
%
%\runningauthor{Surname 1, Surname 2, Surname 3, ...., Surname n}
\twocolumn[

\aistatstitle{Neural Contextual Bandits without Regret}

\aistatsauthor{ Parnian Kassraie \And Andreas Krause }
\aistatsaddress{ ETH Zurich \\ \texttt{pkassraie@ethz.ch} \And  ETH Zurich\\ \texttt{krausea@ethz.ch}} ]

\begin{abstract}
  %Many applications of Machine Learning can be modeled as online optimization of an unknown reward function from noisy samples.
Contextual bandits are a rich model for sequential decision making given side information, with important applications, e.g., in recommender systems. We propose novel algorithms for contextual bandits harnessing neural networks to approximate the unknown reward function. We resolve the open problem of proving sublinear regret bounds in this setting for general context sequences, considering both fully-connected and convolutional networks. 
 To this end, we first analyze \ntkalg, a kernelized bandit optimization algorithm employing the Neural Tangent Kernel (NTK), and bound its regret in terms of the NTK maximum information gain $\gamma_T$, a complexity parameter capturing the difficulty of learning. Our bounds on $\gamma_T$ for the NTK may be of independent interest.
 We then introduce our neural network based algorithm \nnalg, and show that its regret closely tracks that of \ntkalg. Under broad non-parametric assumptions about the reward function, our approach converges to the optimal policy at a $\tilde{\mathcal{O}}(T^{-1/2d})$ rate,  where $d$ is the dimension of the context. 

\end{abstract}

\section{Introduction}\label{sect:intro}
Contextual bandits are a model for sequential decision making based on noisy observations. At every step, the agent is presented with a context vector and picks an action, based on which it receives a noisy reward. Learning about the reward function with few samples (exploration), while simultaneously maximizing its cumulative payoff (exploitation), are the agent's two competing objectives. 
Our goal is to develop an algorithm whose action selection policy attains sublinear regret, which implies convergence to an optimal policy as the number of observations grows.
%, i.e. satisfies the \emph{no-regret} condition.  
A celebrated approach for regret minimization is the {\em optimism} principle: establishing upper confidence bounds (UCB) on the reward, and always selecting a plausibly optimal action.
%solving this problem is estimating both the reward function and its variance, then picking an action that maximizes a linear combination of the two. \emph{Upper Confidence Bound} (UCB) methods are a family of algorithms that follow this policy but use different linear or kernelized models for the estimation part. 
Prior work has developed UCB-based contextual bandit approaches, such as linear or kernelized bandits, for increasingly rich models of reward functions \citep{abbasi2011improved,auer2002using,chowdhury2017kernelized,srinivas2009gaussian}. There are also several recent attempts to harness deep neural networks for contextual bandit tasks.  While these perform well in practice \citep{gu2021batched,riquelme2018deep, zahavy2019deep,zhang2020neural, zhou2019neural}, there is a lack of theoretical understanding of neural network-based bandit approaches.
%, no explicit convergence guarantee has been given for a neural network based contextual algorithm . 

%However, the field of contextual bandits has not yet fully harnessed the advances of deep neural networks 

%there is a lack of contextual bandit methods that harness the computational advances in deep learning, while still yielding regret bounds. 
%While methods are effective for simple problems, they fail to scale to more complicated applications. One issue is that computations become too heavy when the context vector is rich, e.g., making decisions upon images. In addition, kernel methods are outperformed by neural networks for estimating functions that reside in complex spaces and act on high-dimensional input domains \citep{ghorbani2020neural}. The gap is more severe when the input domain consists of images and these methods are compared against convolutional networks.
%A natural solution to the scalability problem is to adhere to the UCB policy, but use a deep model for estimating the reward and its variance. Along this line, there are previous attempts that perform well in practice, but no explicit convergence guarantee has been given for a neural network based algorithm \citep{riquelme2018deep, zahavy2019deep, zhou2019neural}. 

We introduce two optimistic contextual bandit algorithms that employ neural networks to estimate the reward and its uncertainty, \nnalg and its convolutional variant \cnnalg. 
Under the assumption that the unknown reward function $f$ resides in a Reproducing Kernel Hilbert Space (RKHS) with a bounded norm, we prove that both algorithms converge to the optimal policy, if the network is sufficiently wide, or has many channels. 
To prove this bound, we take a two-step approach. We begin by bounding the regret for \ntkalg, which simply estimates mean and variance of the reward via Gaussian process (GP) inference. Here, the covariance function of the GP is set to $\ntk$, the Neural Tangent Kernel associated with the given architecture. We then exploit the fact that neural networks trained with gradient descent approximate the posterior mean of this GP \citep{arora2019exact}, and generalize our analysis of \ntkalg to bound \nnalg's regret.
By drawing a connection between fully-connected and 2-layer convolutional networks, we extend our analysis to include \cntkalg and \cnnalg, the convolutional variants of the algorithms.  A key contribution of our work is bounding the {\em NTK maximum information gain}, a parameter that measures the difficulty of learning about a reward function when it is a sample from $\mathrm{GP}(0, \ntk)$. This result may be of independent interest, as this quantity is integral to sequential decision making approaches.
%For online decision making applications, this bound motivates the extension of other kernelized algorithms to neural network based variants.
%In applications of Bayesian optimization or sequential decision making, analysis of the regret are often in terms of this parameter. Our information gain bound motivates the extension of kernelized algorithms in this field to neural network based variants. 

\paragraph{Related Work} 
\looseness -1 This work is inspired by \citet{zhou2019neural} who introduce the idea of training a neural network within a UCB style algorithm. They analyze \textsc{Neural-UCB}, which bears many similarities to \nnalg. 
Relevant treatments of the regret are given by \citet{gu2021batched,yang2020optimism}, \citet{zhang2020neural}, and \citet{ban2021convolutional} for other neural contextual bandit algorithms. 
However, as discussed in Section \ref{sect:nn}, these approaches do not generally guarantee sublinear regret, unless further restrictive assumptions about the context are made.
%This matter is . 
In addition, there is a large literature on kernelized contextual bandits. Closely related to our work are \citet{krause2011contextual} and \citet{valko2013finite} who provide regret bounds for kernelized UCB methods, with Bayesian and Frequentist perspectives respectively. \citet{srinivas2009gaussian} are the first to tackle the kernelized bandit problem with a UCB based method. Many have then proposed variants of this algorithm, or improved its convergence guarantees under a variety of settings \citep{berkenkamp2016bayesian,bogunovic2020corruption,calandriello2019gaussian,chowdhury2017kernelized, djolonga2013high,kandasamy2019multi,mutny2019efficient,scarlett2018tight}. 
The majority of the bounds in this field are expressed in terms of the {\em maximum information gain}, and \citet{srinivas2009gaussian} establish a priori bounds on this parameter. Their analysis only holds for smooth kernel classes, but has since been extended to cover more complex kernels \citep{janz2020bandit, scarlett2017lower,shekhar2018gaussian, vakili2020information}. In particular, \citet{vakili2020information} introduce a technique that applies to smooth Mercer kernels, which we use as a basis for our analysis of the NTK's maximum information gain. In parallel to UCB methods, online decision making via Thompson Sampling is also extensively studied following \citet{russo2016information}.

Our work further builds on the literature linking wide neural networks and Neural Tangent Kernels. \citet{cao2019generalization} provide important results on training wide fully-connected networks with gradient descent, which we extend to 2-layer convolutional neural networks (CNNs). Through a non-asymptotic bound, \citet{arora2019exact} approximate a trained neural network by the posterior mean of a GP with the NTK as its covariance function. \citet{bietti2020deep} study the Mercer decomposition of the NTK and calculate the decay rate of its eigenvalues, which plays an integral role in our analysis. Little is known about the properties of the Convolutional Neural Tangent Kernel (CNTK), and the extent to which it can be used for approximating trained CNNs. \citet{bietti2021approximation} and \citet{mei2021learning} are among the first to study this kernel by investigating its invariance towards certain groups of transformations, which we draw inspiration from.

\paragraph{Contributions} Our main contributions are: 
\begin{itemize}
\item  \looseness -1 To our knowledge, we are the first to give an explicit sublinear regret bound for a neural network based UCB algorithm. We show that \nnalg's cumulative regret after a total of $T$ steps is $\tilde{\gO}(T^{(2d-1)/2d})$, for any arbitrary context sequence on the  $d$-dimensional hyper-sphere. (\cref{thm:nucbregret})
\item We introduce \cnnalg, and prove that when the number of channels is large enough, it converges to the optimal policy at the same rate as \nnalg. (\cref{thm:cnnucbregret})
\end{itemize}
The $\tilde{\mathcal{O}}$ notation omits the terms of order $\log T$ or slower. Along the way, we present intermediate results that may be of independent interest. In \cref{thm:info_gain} we prove that $\gamma_T$, the maximum information gain for the NTK after $T$ observations, is $\tilde{\mathcal{O}}(T^{(d-1)/d})$. 
We introduce and analyze \ntkalg and \cntkalg, two kernelized methods with sublinear regret (Theorems~\ref{thm:reg1}~\&~\ref{thm:rkhsregret}) that can be used in practice or as a theoretical tool. 
Theorems~\ref{thm:info_gain}~through~\ref{thm:rkhsregret} may provide an avenue for extending other kernelized algorithms to neural network based methods. 

 %%%%%%%%%%%%%%%%%% OLD Contributions$$$$$$$$$$$$
%  \item We prove that $\gamma_T$, the maximum information gain for the NTK after $T$ observations, is $\tilde{\mathcal{O}}(\sqrt{T})$. The $\tilde{\mathcal{O}}$ notation omits the terms of order $\log T$ or slower. (Theorem \ref{thm:info_gain})
% \item We analyze the \ntkalg{} when the reward is sampled from $GP(0,\ntk)$ or is a norm-bounded member of the RKHS for which NTK is reproducing. We show that the cumulative regret $R_T$ after a total of $T$ steps, is $\tilde{\mathcal{O}}(\sqrt{\gamma_TT})$. (Theorem \ref{thm:reg1}  and Theorem \ref{thm:rkhsregret})
% \item As for the \nnalg, if the width of the network is set to be of polynomial order with $T$, we guarantee that its regret enjoys the same rate as \ntkalg. (Theorem \ref{thm:nucbregret})
% \item We extend our analysis to cover the case of a 2-layer CNN and its corresponding CNTK. (Theorem \ref{thm:info_gain_cnn}, Corollary \ref{cor:cntk_regret}, and Theorem \ref{thm:cnnucbregret})
%     \item We establish that under the RKHS assumption, the regret for all four proposed algorithms grows with $\tilde{\mathcal{O}}(T^{3/4})$ for any arbitrary sequence of context vectors. This rate ensures that $R_T/T \rightarrow 0$ as $T \rightarrow \infty$, fulfilling the no-regret condition.

\section{Problem Statement} \label{sect:problem}
%We lay out the setting to the bandit problem. 
\looseness -1 Contextual bandits are a model of sequential decision making over $T$ rounds, where, at step $t$, the learner observes a context matrix $\vz_t$, and picks an action $\va_t$ from $\mathcal{A}$, the finite set of actions. The context matrix consists of a set of vectors, one for each action, i.e., $\vz_t = (\vz_{t,1}, \cdots, \vz_{t,|\mathcal{A}|}) \in \mathbb{R}^{d\times |\mathcal{A}|}$.
The learner then receives a noisy reward $y_t = f(\vx_t) + \epsilon_t$. 
%We allow that a different context vector is observed for each action, i.e., $\vz_t = (\vz_{t,1}, \cdots, \vz_{t,|\mathcal{A}|}) \in \mathbb{R}^{d\times |\mathcal{A}|}$. T
Here, the input to the reward function is the context vector associated with the chosen action, i.e.,  $\vx_t = \vz_t\va_t \in \sR^d$, where $\va_t$ is represented as a one-hot vector of length $|\mathcal{A}|$. Then the reward function is defined as $f: \mathcal{X}\rightarrow \sR$, where $\mathcal{X}\subseteq \mathbb{R}^d$ denotes the input space. Observation noise is modeled with $\epsilon_t$, an i.i.d. sample from a zero-mean sub-Gaussian distribution with variance proxy $\sigma^2 >0$. The goal is to choose actions that maximize the cumulative reward over $T$ time steps. This is analogous to minimizing the {\em cumulative regret}, the difference between the maximum possible (context-dependent) reward and the actual reward received,
$
R_T = \sum_{t=1}^T f(\vx_t^*) - f(\vx_t)
$,
 where $\vx_t$ is the learner's pick and $\vx^*_t$ is the maximizer of the reward function at step $t$
\[
\vx^*_t = \argmax_{\vx = \vz_t\va,\va \in \mathcal{A}} f(\vx).
\]
The learner's goal is to select actions such that $R_T/T \rightarrow 0$ as $T \rightarrow \infty$. This property implies that the learner's policy will converge to the optimal policy.
%and is always picking the correct action. 

\subsection{Assumptions}\label{sect:f_assumptions}
Our regret bounds require some assumptions on the reward function $f$ and the input space $\mathcal{X}$. Throughout this work, we assume that $\mathcal{A}$ is finite and $\mathcal{X}$ is a subset of $\mathbb{S}^{d-1}$ the $d$-dimensional unit hyper-sphere. We consider two sets of assumptions on $f$,
\begin{itemize}
    \item {\em Frequentist Setting}: We assume that $f$ is an arbitrary function residing in $\mathcal{H}_\ntk$ the RKHS that is reproducing for the NTK, and has a bounded kernel norm, $\norm{f}_\ntk \leq B$.
    \item {\em Bayesian Setting}: We assume that $f$ is a sample from a zero-mean Gaussian process, that uses the NTK as its covariance function, $\text{GP}(0, \ntk)$.
\end{itemize}
These assumptions are broad, non-parametric and imply that $f$ is continuous on the hyper-sphere. 
Both the Bayesian and the Frequentist setting impose certain smoothness properties on $f$ via $\ntk$. Technically, the function class addressed by each assumption has an empty intersection with the other.
Appendix \ref{app:gp_vs_rkhs} provides more insight into the connection between the two assumptions. We require a mild {\em Sufficient Exploration} assumption on the kernel matrix, exclusive to the results in Sections~\ref{sect:nn}~and~\ref{sect:cnn}. This is presented later, under Assumption~\ref{cond:suf_exp}.

\subsection{The Neural Tangent Kernel} \label{sect:ntk_intro}
\looseness -1 We review important properties of the NTK as relied upon in this work. Training very wide neural networks has similarities to estimation with kernel methods using the NTK. For now, we will focus on fully-connected feed-forward ReLU networks and their corresponding NTK. In Section~\ref{sect:cnn}, we extend our result to networks with one convolutional layer. Let $f(\vx; \vtheta): \mathbb{R}^d \rightarrow \mathbb{R}$ be a fully-connected network, with $L$ hidden layers of equal width $m$, and ReLU activations, recursively defined as follows,
\begin{equation*}
    %\label{eq:def_nn}
    \begin{split}
        f^{(1)}(\vx) &= \mW^{(1)}\vx,\\
     f^{(l)}(\vx) &= \sqrt{\frac{2}{m}} \mW^{(l)} \sigma_{\text{relu}}\big(f^{(l-1)}(\vx)\big) \in \mathbb{R}^m, \,\, 1<l\leq L\\
     f(\vx; \vtheta) &= \sqrt{2} \mW^{(L+1)} \sigma_{\text{relu}}\big(f^{(L)}(\vx)\big)\in \sR.  
    \end{split}
\end{equation*}
The weights $\mW^{(i)}$ are initialized to random matrices with standard normal i.i.d.~entries, and $\vtheta = (\mW^{(i)})_{i\leq L+1}$. Let $\vg(\vx; \vtheta)=\nabla_\vtheta f(\vx; \vtheta)$ be the gradient of $f$. Assume that given a fixed dataset, the network is trained with gradient descent using an infinitesimally small learning rate. For networks with large width $m$, training causes little change in the parameters and, respectively, the gradient vector. For any $\vx,\, \vx' \in \mathcal{X}$, and as $m$ tends to infinity, a limiting behavior emerges: $\langle\vg(\vx; \vtheta),\vg(\vx'; \vtheta)\rangle/m$, the inner product of the gradients, remains constant during training and converges to $\ntk(\vx; \vx')$, a deterministic kernel function \citep{arora2019exact,jacot2018neural}. This kernel satisfies the conditions of Mercer's Theorem over $\sS^{d-1}$ with the uniform measure \citep{cao2019towards} and has the following Mercer decomposition,
\begin{equation} \label{eq:mercer_decompostion}
    \ntk(\vx, \vx') = \sum_{k=0}^\infty \mu_k \sum_{j=1}^{N(d,k)} Y_{j,k}(\vx)Y_{j,k}(\vx'),
\end{equation}
where $Y_{j,k}$ is the $j$-th spherical harmonic polynomial of degree $k$, and $N(d, k)$ denotes the algebraic multiplicity of $\mu_k$. In other words, each $\mu_k$ corresponds to a $N(d,k)$ dimensional eigenspace, where $N(d,k)$ grows with $k^{d-2}$. Without loss of generality, assume that the distinct eigenvalues $\mu_k$ are in descending order. \citet{bietti2020deep} show that there exists an absolute constant $\CNN$ such that
\begin{equation} \label{eq:ntk_decayrate}
    \mu_k \simeq \CNN k^{-d}.
\end{equation}
 Taking the algebraic multiplicity into account, we obtain that the decay rate for the complete spectrum of eigenvalues is of polynomial rate $k^{-1-1/(d-1)}$. This decay is slower than that of the kernels commonly used for kernel methods. The eigen-decay for the squared exponential kernel is $O(\exp(-k^{1/d}))$ \citep{belkin2018approximation}, and Mat\'ern kernels with smoothness $\nu > 1/2$ have a $O(k^{-1-2\nu/d})$ decay rate \citep{santin2016approximation}. 
The RKHS associated with $\ntk$ is then given by
\begin{align}
%\begin{equation} 
%\begin{split}
        \mathcal{H}_\ntk = \Big\{f: \, f = \sum_{k\geq 0} \sum_{j=1}^{N(d,k)} & \beta_{j,k}Y_{j,k},\label{eq:ntk_rkhs} \\
        & \sum_{k\geq 0} \sum_{j=1}^{N(d,k)}\frac{\beta_{j,k}^2}{\mu_k}< \infty  \Big\}\nonumber.
%\end{split}
%\end{equation}
\end{align}
Equation \ref{eq:ntk_rkhs} explains how the eigen-decay of $k$ controls the complexity of $\mathcal{H}_k$. Only functions whose coefficients $\beta_{j,k}$ decay at a faster rate than the kernel's eigenvalues are contained in the RKHS. Therefore, if the eigenvalues of $k$ decay rapidly, $\gH_k$ is more limited. The slow decay of the NTK's eigenvalues implies that the assumptions on the reward function given in Section \ref{sect:f_assumptions} are less restrictive compared to what is often studied in the kernelized contextual bandit literature.

\section{Warm-up: \ntkalg\ -- Kernelized Contextual Bandits with the NTK} \label{sect:ntk}
Our first step will be to analyze kernelized bandit algorithms that employ the NTK as the kernel. In particular, we focus on the \emph{Upper Confidence Bound} (UCB) exploration policy \citep{srinivas2009gaussian}. Kernelized bandits can be interpreted as modeling the reward function $f$ via a Bayesian prior, namely a Gaussian process $\text{GP}(0, k)$ with covariance function given by $k$. At each step $t$, we calculate the posterior mean and variance $\mu_{t-1}(\cdot)$ and $\sigma_{t-1}(\cdot)$, using the samples observed at previous steps. For i.i.d.~ $\mathcal{N}(0, \sigma^2)$~noise, the posterior GP has a closed form expression,
\begin{align}
    \mu_{t-1} (\vx)& = \bm{k}_{t-1}^T(\vx)(\bm{K}_{t-1}+\sigma^2\bm{I})^{-1}\vy_{t-1} \label{eq:GPposteriors} \\
    \sigma^2_{t-1}(\vx) & = k(\vx, \vx) - \bm{k}^T_{t-1}(\vx)(\bm{K}_{t-1}+\sigma^2\bm{I})^{-1}\bm{k}_{t-1}(\vx) \nonumber
\end{align}
where $\vy_{t-1} = [y_i]_{i < t}$ is the vector of received rewards, $\bm{k}_{t-1}(\vx) = [k(\vx, \vx_i)]_{i < t}$, and $\bm{K}_{t-1} = [k(\vx_i, \vx_j)]_{i,j < t}$ is the kernel matrix. We then select the action by maximizing the UCB,
\begin{equation}\label{eq:UCB_policy}
\vx_t = \argmax_{\vx = \vz_t\va,\,\va \in \mathcal{A}} \mu_{t-1}(\vx) + \sqrt{\beta_t}\sigma_{t-1}(\vx).
\end{equation}
The acquisition function balances exploring uncertain actions and exploiting the gained information via parameter $\beta_t$ which will be detailed later. %This policy roughly translates to picking an action that optimistically would give the maximum reward.  
Our method \ntkalg, adopts the UCB approach, and uses $\ntk$ as the covariance kernel function of the GP for calculating the posteriors in Equation \ref{eq:GPposteriors}.
\subsection{Information Gain} \label{sect:info_gain}
The UCB policy seeks to learn about $f$ quickly, while picking actions that also give big rewards. The speed at which we learn about $f$ is quantified by the {\em maximum information gain}.
Assume that for a sequence of inputs $X_T = (\vx_1, \cdots, \vx_T)$, the learner observes noisy rewards $\vy_T=(y_1,\dots,y_T)$, and let $\vf_T=(f(\vx_1),\dots,f(\vx_T))$ be the corresponding true rewards. Then the information gain is defined as the mutual information between these random vectors, $I(\vy_T; \vf_T) := H(\vy_T) - H(\vy_T\vert\vf_T)$, where $H$ denotes the entropy.
%In Appendix \ref{app:infogain_complexity} we elaborate on how $I(\vy_T; \vf_T)$ can also be viewed as a measure of model complexity.
Assuming the GP prior $f \sim \text{GP}(0, \ntk)$, and in the presence of i.i.d.~Gaussian noise,
\[I(\vy_T; \vf_T) = \frac{1}{2}\log\det(\bm{I} + \sigma^{-2}\mK_T)\]
with the kernel matrix $\mK_T = [\ntk(\vx_i,\vx_j)]_{i,j\leq T}$. 
Following \citet{srinivas2009gaussian}, we will express our regret bounds in terms of the information gain.
%and it is desired that $R_T/T$ convergences to zero, uniformly for any arbitrary sequence $X_T$. 
The information gain depends on the sequence of points observed. 
%, and its growth rate with $T$ is unclear. 
To obtain bounds for arbitrary context sequences, we work with the {\em maximum information gain} defined as $\gamma_T := \max_{X_T} I(\vy_T; \vf_T)$. By bounding $I(\vy_T; \vf_T)$ with $\gamma_T$, we obtain regret bounds that are {\em independent} of the input sequence.

Many regret bounds in this literature, including ours, are of the form $\tilde{\mathcal{O}}(\sqrt{T\gamma_T})$ or $\tilde{\mathcal{O}}(\sqrt{T}\gamma_T)$.
%, in Bayesian and Frequentist setting, respectively. 
For such a bound not to be vacuous, i.e., for it to guarantee convergence to an optimal policy, $\gamma_T$ must grow strictly sub-linearly with $T$. Our first main result is an \textit{a priori} bound on $\gamma_T$ for Neural Tangent Kernels corresponding to fully-connected networks of depth $L$. 
\begin{theorem} \label{thm:info_gain}
%Consider a GP with covariance function $\ntk$, the 
Suppose the observation noise is i.i.d., zero-mean and a Gaussian of variance $\sigma^2>0$, and the input domain $\mathcal{X} \subset \mathbb{S}^{d-1}$. Then the maximum information gain associated with the NTK of a fully-connected ReLU network is bounded by
\resizebox{\columnwidth}{!}{
$
        \gamma_T = \mathcal{O}\bigg(  \Big( \frac{\CNN T}{\log (1+ \frac{T}{\sigma^2})} \Big)^{\frac{d-1}{d}}
         \log\Big( 1
          + \frac{T}{\sigma^{2}}\Big( \frac{\CNN T}{\log (1+ \frac{T}{\sigma^2})} \Big)^{\frac{d-1}{d}} \Big)\bigg)
$
}
\end{theorem}

\looseness -1 The parameter $\gamma_T$ arises not only in the bandit setting, but in a broad range of related sequential decision making tasks  \citep{berkenkamp2016bayesian,kandasamy2016gaussian,kirschner2020distributionally,kirschner2019stochastic,sessa2019no, sessa2020contextual,sui18a}. \cref{thm:info_gain} might therefore be of independent interest and facilitate the extension of other kernelized algorithms to neural network based methods. 
When restricted to $\sS^{d-1}$, the growth rate of $\gamma_T$ for the NTK matches the rate for a Mat\'ern kernel with smoothness coefficient of $\nu = 1/2$, since both kernels have the same rate of eigen-decay \citep{chen2020deep}. \citet{srinivas2009gaussian} bound $\gamma_T$ for smooth Mat\'ern kernels with $\nu \geq 1$, and \citet{vakili2020information} extend this result to $\nu > 1/2$. From this perspective, \cref{thm:info_gain} pushes the previous literature one step further by bounding the information gain of a kernel with the same eigen-decay as a Mat\'ern kernel with $\nu = 1/2$.\footnote{Under the assumption that $f \sim \mathrm{GP}(0,k)$ with the covariance function a Mat\'ern $\nu = 1/2$, \citet{shekhar2018gaussian} give a \emph{dimension-type} regret bound for a tree-based bandit algorithm. Their analysis however, is not in terms of the information gain, due to the structure of this algorithm.}
%\footnote{\looseness -1 For any $d\geq 1$ and $\nu \geq 0$, the NTK on $\sS^{d-1}$ has a sharper information gain bound compared to the $\nu$-Mat\'ern kernel on $\sR^d$, for which $\gamma_T$ is lower bounded by $\Omega(T^{(\nu + d)/(2\nu + d)})$, when the domain is restricted to the $d$-dimensional hypercube \citep{scarlett2017lower}. This is mainly due to the fact the NTK is rotation invariant and its eigenfunctions are spherical harmonics, while the Mat\'ern family are translation invariant and are diagonalized by the Fourier basis.}
% Both $1/2$-Mat\'ern kernel and the NTK have the same rate of eigen-decay, the NTK however, has a piecewise constant eigenspectrum. This allows a sharper information gain bound, since $\gamma_T$ depends on the sum eigenvalues.}
\paragraph{Proof Idea} Beyond classical analyses of $\gamma_T$, additional challenges arise when working with the NTK, since it does not have the smoothness properties required in prior works.
As a consequence, we directly use the Mercer decomposition of the NTK (Eq. \ref{eq:mercer_decompostion}) and break it into two terms, one corresponding to a kernel with a finite-dimensional feature map, and a tail sum. We separately bound the information gain caused by each term. 
From the Mat\'ern perspective, we are able to extend the previous results, in particular due to our treatment of the Mercer decomposition tail sum. An integral element of our approach is a fine-grained analysis of the NTK's eigenspectrum over the hyper-sphere, given by \citet{bietti2020deep}. The complete proof is given in Appendix \ref{app:infogainbound}.

\subsection{Regret Bounds}
We now proceed with bounding the regret for \ntkalg, under both Bayesian and Frequentist assumptions, as explained in Section \ref{sect:f_assumptions}. Following \citet{krause2011contextual}, and making adjustments where needed, we obtain a bound for the Bayesian setting. 
\begin{theorem} \label{thm:reg1}
Let $\delta \in (0, 1)$ and suppose $f$ is sampled from $\mathrm{GP}(0, \ntk)$. Values of $f$ are observed with a zero-mean Gaussian noise of variance $\sigma^2$, and the exploration parameter is set to $\beta_t = 2 \log (\vert\mathcal{A} \vert t^2 \pi^2/6\delta)$. Then with probability greater than $1-\delta$, the regret of \ntkalg  satisfies
\begin{equation*}
    R_T \leq C \sqrt{T\beta_T\gamma_T}
\end{equation*}
for any $T\geq 1$, where $C := \sqrt{8\sigma^{-2}/\log (1+\sigma^{-2})}$.
\end{theorem}
Crucially, this bound holds for {\em any} sequence of observed contexts, since $\gamma_T$ is deterministic and only depends on $\sigma$, $T$, the kernel function $\ntk$, and the input domain $\mathcal{X}$. A key ingredient in the proofs of regret bounds, including \cref{thm:reg1}, is a concentration inequality of the form 
\begin{equation}\label{eq:concentration}
\abs{f(\vx_t) - \mu_{t-1}(\vx_t)} \leq \sqrt{\beta_t} \sigma_{t-1}(\vx_t),
\end{equation}
which holds with high probability for every $\vx_t$ and $t$, conditioned on context-reward pairs from the previous steps. This inequality holds naturally under the GP assumption, since $f(\vx)$ is obtained directly from Bayesian inference. Setting $\beta_t$ to grow with $\log t$ satisfies the inequality and results in a $\tilde{\mathcal{O}}(\sqrt{T\gamma_T})$ regret bound. 
However, additional challenges arise under the RKHS assumption. 
For Equation
~\ref{eq:concentration} to hold in this setting, we need $\beta_t$ to grow with $\gamma_t \log t$. The regret would then be $\tilde{\mathcal{O}}(\sqrt{T}\gamma_T)$ \citep{chowdhury2017kernelized}. For \ntkalg, $\gamma_T$ is $\tilde{\mathcal{O}}(T^{(d-1)/d})$, and the $\gamma_T\sqrt{T}$ rate would no longer imply convergence to the optimal policy for $d \geq 2$. To overcome this open problem \citep{vakili2021open}, we analyze a variant of our algorithm -- called the {\em Sup} variant -- that has been successfully applied in the kernelized bandit literature
\citep{auer2002using,chu2011contextual,li2010contextual,  valko2013finite}.
A detailed description of the \supntkalg, along with its pseudo-code and properties is given in Appendix \ref{app:supvar}. Here we give a high-level overview of the Sup variant, and how it resolves the large $\beta_t$ problem. This variant combines \ntkalg policy with Random Exploration (RE). At \ntkalg  steps, the UCB is calculated only using the context-reward pairs observed in the previous RE steps. Moreover, the rewards received during the RE steps are {\em statistically independent} conditioned on the input for those steps. Together with other properties, this allows a choice of $\beta_t$ that grows with $\log T$. We obtain the following:%bound
\begin{theorem} \label{thm:rkhsregret}
Let $\delta \in [0,1]$. Suppose $f$ lies in the RKHS of $\ntk$, with $\vert \vert f \vert \vert_\ntk \leq B$. Samples of $f$ are observed with a zero-mean sub-Gaussian noise with variance proxy $\sigma^2$. Then for a constant $\beta_t = 2\log \big( 2T|\mathcal{A}|/\delta\big)$, with probability greater than $1-\delta$, \supntkalg  algorithm satisfies
\begin{equation*}
    \begin{split}
R(T)  =  \mathcal{O}\Big(\sqrt{T}\Big(& \sqrt{ \gamma_T \sigma^{-2}(\log T)^3\log(T\log T |\mathcal{A}|/\delta)}\\
& + \sigma B\Big) \Big).
    \end{split}
\end{equation*}
\end{theorem}
The first term corresponds to regret of the random exploration steps, and the second term results from the steps at which the actions were taken by the \ntkalg  policy. Proofs of Theorems~\ref{thm:reg1}~and~\ref{thm:rkhsregret} are given in Appendices ~\ref{app:gpregret}~and~\ref{app:ntkrkhsregret} respectively. We finish our analysis of the \ntkalg  with the following conclusion, employing our bound on $\gamma_T$ from \cref{thm:info_gain} in Theorems~\ref{thm:reg1}~and~\ref{thm:rkhsregret}.  
\begin{corollary} \label{cor:NTK_regret}
%It follows from Theorems \ref{thm:info_gain}, \ref{thm:reg1}, and \ref{thm:rkhsregret}, that when 
\looseness - 1 Suppose $f$ satisfies either the GP or RKHS assumption.  Then the \ntkalg  (resp.~its Sup variant) has sublinear regret $R(T) = \tilde{\mathcal{O}}(\tCNN T^{\frac{2d-1}{2d}})$ with high probability. Hereby, $\tCNN$ is a coefficient depending on the eigen-decay of the NTK.
\end{corollary}
\section{Main Result: \nnalg\ -- Neural Contextual Bandits without Regret} \label{sect:nn}
Having analyzed 
%At the first stage of our two-step approach, we analyzed 
\ntkalg, %a kernelized method that uses the NTK, 
we now %. For the second step, we 
present our neural net based algorithm \nnalg, which leverages the connections between NN training and GP regression with the NTK. By design, \nnalg benefits from the favorable properties of the kernel method which helps us with establishing our regret bound. \nnalg  results from approximating the posterior mean and variance functions appearing in the UCB criterion (Equation \ref{eq:UCB_policy}). First, we approximate the posterior mean $\mu_{t-1}$ with $f^{(J)} = f(\vx; \vtheta^{(J)})$, the neural network trained for $J$ steps of gradient descent with some learning rate $\eta$ with respect to the regularized LSE loss
\begin{equation}
    \label{eq:post_var_est}
    \mathcal{L}(\vtheta) = \sum_{i=1}^{t-1} \big( f(\vx_i; \vtheta) - y_i  \big)^2 + m \sigma^2 \big\vert\big\vert\vtheta - \vtheta^{0}\big\vert\big\vert^2_2,
\end{equation}
where $m$ is the width of the network and $\vtheta^{0}$ denotes the network parameters at initialization. This choice is motivated by \citet{arora2019exact}, who show point-wise convergence of
%$f^{(\infty)}(\vx)$, 
the solution of gradient descent on the unregularized LSE loss, to
%$f_{\text{ntk}}(\vx)$, 
the GP posterior mean when there is no observation noise. We adapt their result to our setting where we consider $\ell_2$ regularized loss and noisy samples.
%They show that if $m$ is $\mathrm{Poly}(t, 1/\epsilon)$, then for any test point $\vx$, with high probability $f^{(\infty)}(\vx)$, the solution of gradient descent on the unregularized LSE loss, is closer than $\epsilon$ to $f_{\text{ntk}}(\vx)$, the GP posterior mean when there is no observation noise.
It remains to approximate the posterior variance. Recall from Section \ref{sect:ntk_intro} that the NTK is the limit of $\langle \vg(\vx), \vg(\vx')\rangle/m$ as $m \rightarrow \infty$, where $\vg(\cdot)$ is the gradient of the network at initialization. This property hints that for a wide network, $\vg/\sqrt{m}$ can be viewed as substitute for $\vphi$, the infinite-dimensional feature map of the NTK, since $\ntk(\vx, \vx') = \langle \vphi(\vx), \vphi(\vx')\rangle$. By re-writing $\sigma_{t-1}$ in terms of $\vphi$ and substituting $\vphi$ with $\vg/\sqrt{m}$, we get
\[    \hat\sigma^2_{t-1}(\vx) =  \frac{\vg^T(\vx)}{\sqrt{m}}\Big(\sigma^2 \mI + \sum_{i = 1}^{t-1} \frac{\vg^T(\vx_i)\vg(\vx_i)}{m}\Big)^{-1}\frac{\vg(\vx)}{\sqrt{m}}.\]
\looseness -1 At the beginning, \nnalg  initializes the network parameters to $\vtheta^0$. Then at step $t$, $\hat\sigma_{t-1}(\cdot)$ is calculated using $\vg(\cdot, \vtheta^0)$ and the action is chosen via maximizing the \emph{approximate UCB} 
\[
 \vx_t = \argmax_{\vx = \vz_t\va,\, \va \in \gA} f^{(J)}(\vx; \vtheta_{t-1}) + \sqrt{\beta_t}\hat\sigma_{t-1}(\vx)
\]
where $\vtheta_{t-1}$ is obtained by training $f(\cdot; \vtheta^0)$, for $J$ steps, with gradient descent on the data observed so far. This algorithm essentially trains a neural network for estimating the reward and combines it with a random feature model for estimating the variance of the reward. These random features arise from the gradient of a neural network with random Gaussian parameters. The pseudo-code to \nnalg  is given in Appendix \ref{app:nn_supvar}. %Despite the Upper Confidence Bound, the approximate UCB does not have a clear mathematical interpretation. 
In Appendix \ref{app:interp}, we assess the ability of the approximate UCB criterion to quantify uncertainty in the reward via experiments on the \mnist dataset.

\paragraph{Regret Bound} Similar to \cref{thm:rkhsregret}, we make the RKHS assumption on $f$ and establish a regret bound on the Sup variant of \nnalg. To do so, we need two further technical assumptions. Following \citet{zhou2019neural}, for convenience we assume that $f(\vx; \vtheta^0) = 0$,  for any $\vx = \vz_t\va$ where $t \leq T$ and $\va \in \mathcal{A}$. As explained in Appendix \ref{app:nn_additional_assump}, this requirement can be fulfilled without loss of generality. 
\begin{assumption}[Sufficient Exploration]\label{cond:suf_exp}
The kernel matrix is bounded away from zero, i.e., $\lambda_0\mI \preccurlyeq \mK_T$.
\end{assumption}
This assumption is common within the literature \citep{arora2019exact, cao2019generalization,du2019gradient,zhou2019neural} and is satisfied as long as the learner sufficiently explores the input space, such that no two inputs $\vx_t$ and $\vx_{t'}$ are identical. Further, a weaker version of it is often required to hold for the kernel matrix in the sparse linear bandits literature \citep{bastani2020online,hao2020high, kim2019doubly}. 
\begin{theorem} \label{thm:nucbregret}
Let $\delta \in (1, 0)$. Suppose $f$ lies in the RKHS of $\ntk$ with $\norm{f}_\ntk \leq B$. Samples of $f$ are observed with zero-mean sub-Gaussian noise of variance proxy $\sigma^2$. Set $J>1$ and $\beta_t = 2\log( 2T|\mathcal{A}|/\delta)$ constant. Choose the width such that
\begin{align*}
         m \geq \text{poly}\left(T, L, |\mathcal{A}|, \sigma^{-2},B, \lambda_0^{-1}, \log(1/\delta)\right),
         \end{align*}
and $\eta = C(LmT+m\sigma^2)^{-1}$ with some universal constant $C$. Then, with probability greater than $1-\delta$, the regret of \supnnalg  satisfies
\begin{equation*}
    \begin{split}
R(T)  = \mathcal{O}\Big( \sqrt{T}\Big(&\sqrt{\gamma_T\sigma^{-2}(\log T)^3\log(T\log T|\mathcal{A}|/\delta)}\\
& + \sigma B\Big) \Big).
    \end{split}
\end{equation*}
\end{theorem}
\looseness -1 The pseudo-code of \supnnalg and the proof are given in Appendix \ref{app:nn_supvar}. % and Appendix \ref{app:thm4proof}. 
The key idea there is to show that given samples with noisy rewards, members of $\mathcal{H}_k$ are well estimated by the solution of gradient descent on the $\ell_2$ regularized LSE loss. The following lemma captures this statement. 

\begin{lemma}[Concentration of $f$ and $f^{(J)}$, simplified] \label{lem:gd_gp} Consider the setting of \cref{thm:nucbregret} and further assume that the rewards $\{y_i\}_{i<t}$ are independent conditioned on the contexts $\{\vx_i\}_{i<t}$. Let $0<\delta<1$ and set $m$, $\beta_t$ and $\eta$ according to \cref{thm:nucbregret}. Then, with probability greater than $1-2e^{-\beta_T/2}-\delta$,
\begin{align*}
\vert f^{(J)}(\vx_t) - f(\vx_t) \vert & \leq \hat\sigma_{t-1}(\vx_t) \sqrt{\beta_T} \,\mathrm{Poly}(B, m, t, L, \eta)
\end{align*}
\end{lemma} 
\looseness -1 \cref{thm:nucbregret} shows that \supnnalg  obeys the same regret guarantee as \supntkalg. In the theorem, the asymptotic growth of the regret is given for large enough $m$, and terms that are $o(1)$ with $T$ are neglected. To compare the two algorithms in more detail, we revisit the bound for a fixed $m$. With a probability greater than $1-\delta$,
\begin{equation*} \label{eq:regret}
    \begin{split}
    R_T \leq&  \mathcal{O}\Big(\sqrt{T\gamma_T} \sqrt{\sigma^{-2}(\log T)^3\log\left( T\log T |\mathcal{A}|/\delta\right)}\\
    & +
    \left(1 + \sigma\sqrt{\left(m\log( T\log T |\mathcal{A}|/\delta)\right)^{-1}}\right) \sigma B\sqrt{T}\\
    &  +
    L^3\left(\frac{TB}{m\sigma^2} \right)^{5/3} \sqrt{m^3 \log m}\\
    & + \frac{\sqrt{B}(1-m\eta \sigma^2)^{J/2}}{\sqrt{m\eta \log(T\log T |\mathcal{A}|/\delta)}} \Big).
    \end{split}
\end{equation*}
The last two terms, which vanish for sufficiently large $m$, convey the error of approximating GP inference with NN training: The fourth term is the gradient descent optimization error, and the third term is a consequence of working with the linear first order Taylor approximation of $f(\vx; \vtheta)$. The first two terms, however, come from selecting explorative actions, as in the regret bound of \ntkalg  (\cref{thm:rkhsregret}).  The first term denotes regret from random exploration steps, and the second presents the regret at the steps for which the UCB policy is used to pick actions.
\paragraph{Comparison with Prior Work}\looseness -1 The \textsc{Neural-UCB} algorithm introduced by \citet{zhou2019neural} bears resemblance to our method. At step $t$, \nnalg  approximates the posterior variance via Equation~ \ref{eq:post_var_est} with $\vg(\cdot; \vtheta^{0})$, a fixed feature map. \textsc{Neural-UCB}, however, updates the feature map at every step $t$, by using $\vg(\cdot; \vtheta_{t-1})$, where $\vtheta_{t-1}$ is defined as before.
Effectively, \citeauthor{zhou2019neural} adopt a GP prior that changes with $t$. %, but no reasoning is given for this choice. 
Under additional assumptions on $f$ and for $\sigma \geq \max\{1, B^{-1}\}$, they show that for  \textsc{Neural-UCB}, a guarantee of the following form holds with probability greater than $1-\delta$.
\begin{equation*}
\begin{split}
        R_T \leq \tilde{\mathcal{O}}\Big( \sqrt{T I(\vy_T; \vf_T)} \big[ & \sigma \sqrt{I(\vy_T; \vf_T) + 1 -\log \delta} \\
        & + \sqrt{T} (\sigma + \frac{TL}{\sigma})\big(1-\frac{\sigma^2}{TL}\big)^{J/2} \\
        & + \sigma B\big] \Big)
\end{split}
\end{equation*}
The bound above is {\em data-dependent} via $I(\vy_T; \vf_T)$ and in this setting, the only known way of bounding the information gain is through $\gamma_T$. The treatment of regret given in \citet{yang2020optimism} and \citet{zhang2020neural} also results in a bound of the form $\tilde{\gO}(\sqrt{T}\gamma_T)$. However, the maximum information gain itself grows as $\tilde{\gO}(T^{(d-1)/d})$ for the NTK covariance function. Therefore, without further assumptions on the sequence of contexts, the above bounds are vacuous.
%While \textsc{Neural-UCB} is expected to have sublinear regret, the way this bound is presented is vacuous. 
In contrast, 
%Through our two-step approach, we are able to give an a 
our regret bounds for \nnalg  are {\em sublinear  without any further restrictions on the context sequence}. This follows from \cref{thm:info_gain} and \cref{thm:nucbregret}:
\begin{corollary} \label{cor:nucbregret}
Under the conditions of \cref{thm:nucbregret}, for arbitrary sequences of contexts, with probability greater than $1-\delta$, \supnnalg  satisfies,
\[
R(T) = \tilde{\mathcal{O}}(\tCNN T^{\frac{2d-1}{2d}}).
\]
\end{corollary}
\looseness -1 The coefficient $\tCNN$ in Corollary~\ref{cor:NTK_regret}~and~\ref{cor:nucbregret} denotes the same constant. Figures~\ref{fig:infogain_antkvnucb}~and~\ref{fig:regret_allntks} in \cref{app:experiments} plot the information gain and regret obtained for \nnalg  when used on the task of online \mnist  classification.
%The empirical growth rates closely track our a priori bounds, suggesting that our analysis describes the practical behavior of the algorithms well.
\section{Extensions to Convolutional Neural Networks} \label{sect:cnn}
So far, regret bounds for contextual bandits based on convolutional neural networks have remained elusive.
Below, we %In this section we revisit the bounds given so far,  and 
extend our results to a particular case of 2-layer convolutional networks. 
Consider a cyclic shift $c_l$ that maps $\vx$ to $c_l \cdot \vx = (x_{l+1}, x_{l+2}, \cdots, x_{d}, x_1, \cdots, x_l)$. We can write a 2-layer CNN, with one convolutional and one fully-connected layer, as a 2-layer NN that is averaged over all cyclic shifts of the input
\begin{equation*} \label{eq:def_cnn}
\begin{split}
   f_{\text{CNN}}(\vx; \vtheta) & =\sqrt{2} \sum_{i=1}^m v_i \Big[ \frac{1}{d} \sum_{l = 1}^{d} \sigma_{\mathrm{relu}}(\langle \vw_i, c_l \cdot \vx\rangle)\Big] \\
   & = \frac{1}{d} \sum_{l = 1}^{d} f_{\text{NN}}(c_l \cdot \vx; \vtheta). 
\end{split}
\end{equation*}
\looseness -1 Let $\gC_d$ denote the group of cyclic shifts $\{c_l\}_{l\leq d}$. Then the 2-layer CNN is $\mathcal{C}_d$-invariant, i.e., $ f_{\text{CNN}}(c_l\cdot \vx)=f_{\text{CNN}}(\vx)$, for every $c_l$. The corresponding CNTK is also $\gC_d$-invariant and can be viewed as an averaged NTK
\begin{equation} \label{eq:cntk_def}
\begin{split}
    \cntk(\vx, \vx') & = \frac{1}{d^2}\sum_{l,\, l' = 1}^{d}\ntk(c_l \cdot \vx, c_{l'} \cdot \vx') \\
    & = \frac{1}{d}\sum_{l = 1}^{d} \ntk(\vx, c_l \cdot \vx').
\end{split}
\end{equation}
The second equality holds because $\ntk(\vx, \vx')$ depends on its arguments only through the angle between them. In Appendix \ref{app:cntk_vs_ntk}, we give more intuition about this equation via the random feature kernel formulation \citep{chizatlazy,rahimi2008random}.
Equation \ref{eq:cntk_def} implies that the CNTK is a Mercer kernel and in Lemma~\ref{lem:cnn_ext} we give its Mercer decomposition. The proof is presented in Appendix \ref{app:cntk_mercer}.
\begin{lemma} \label{lem:cnn_ext}
 The Convolutional Neural Tangent Kernel corresponding to $f_{\mathrm{CNN}}(\vx; \vtheta)$, a 2-layer CNN with standard Gaussian weights, can be decomposed as
 \begin{equation*}
    \cntk(\vx, \vx') = \sum_{k=0}^\infty \mu_k \sum_{j=1}^{\bar N(d,k)} \bar Y_{j,k}(\vx)\bar Y_{j,k}(\vx')
\end{equation*}
where $\mu_k \simeq \CNN k^{-d}$. The algebraic multiplicity is $\bar N(d,k) \simeq N(d,k)/d$, and the eigenfunctions $\{\bar Y_{j,k}\}_{j \leq \bar N(d,k)}$ form an orthonormal basis for the space of $\mathcal{C}_d$-invariant degree-$k$ polynomials on $\sS^{d-1}$. 
\end{lemma}

\looseness -1 With this lemma, we show that the 2-layer CNTK has the same distinct eigenvalues as the NTK, while the eigenfunctions and the algebraic multiplicity of each distinct eigenvalue change. The eigenspaces of the NTK are degree-$k$ polynomials, while for the CNTK, they shrink to $\mathcal{C}_d$-invariant degree-$k$ polynomials. This reduction in the dimensionality of eigenspaces results in a smaller algebraic multiplicity for each distinct eigenvalue. 

\paragraph{Information Gain} We begin by bounding the maximum information gain $\bar \gamma_T$, when the reward function is assumed to be a sample from $\text{GP}(0, \cntk)$.
\looseness -1 Proposition~\ref{thm:info_gain_cnn} establishes that the growth rate of $\bar \gamma_T$ matches our result for maximum information gain of the NTK. The dependence on $d$ however, is improved by a factor of $d^{(d-1)/d}$, indicating that the speed of learning about the reward function is potentially $d^{(d-1)/d}$ times faster for methods that use a CNN. The proof is given in Appendix~\ref{app:infogainbound}. 
 \begin{prop}\label{thm:info_gain_cnn}
%Let $f \sim \mathrm{GP}(0,\cntk)$ over the 
Suppose the input domain satisfies $\mathcal{X} \subset \mathbb{S}^{d-1}$, and samples of $f$ are observed with i.i.d.~zero-mean sub-Gaussian noise of variance proxy $\sigma^2>0$. Then the maximum information gain of the {\em convolutional neural tangent kernel $\cntk$} satisfies\\
\resizebox{\columnwidth}{!}{
$
\bar \gamma_T = \mathcal{O}\bigg(\Big(\frac{T\CNN}{d\log(1+ \frac{T}{\sigma^2})}\Big)^{\frac{d-1}{d}} \log\bigg( 1 + \frac{T}{\sigma^2}\Big(\frac{T\CNN}{d\log(1+ \frac{T}{\sigma^2})}\Big)^{\frac{d-1}{d}} \bigg)\bigg).
$
}
\end{prop}
\paragraph{\cntkalg} This algorithm is set to be the convolutional variant of \ntkalg. We take the UCB policy (Equation \ref{eq:UCB_policy}) and plug in $\cntk$ as the covariance function for calculating the posterior mean and variance. 
By Lemma~\ref{lem:cnn_ext}, the rate of eigen-decay, and therefore, the smoothness properties, are identical between NTK and the 2-Layer CNTK. Through this correspondence, Theorems~\ref{thm:reg1}~and~\ref{thm:rkhsregret} carry over to \cntkalg, thus guaranteeing  sublinear regret.
\begin{corollary} \label{cor:cntk_regret}
It follows from \cref{thm:reg1}, \cref{thm:rkhsregret}, and Proposition~\ref{thm:info_gain_cnn}, that when $f$ satisfies either the GP or the RKHS assumption, \cntkalg  (resp.~its Sup variant) has a sublinear regret 
\[R(T) = \tilde{\mathcal{O}}\left(\frac{\tCNN}{d^{(d-1)/2d}} T^{\frac{2d-1}{2d}}\right)
\]
with high probability. Here $\tCNN$ is a coefficient depending on the eigen-decay of the NTK. 
\end{corollary}
Corollary \ref{cor:cntk_regret} implies that while regret for both algorithms grows at the same rate with $T$, \cntkalg  potentially outperforms \ntkalg. This claim is investigated on the online \mnist  classification task in Figure \ref{fig:cnnplot} in Appendix \ref{app:experiments}.
\paragraph{\cnnalg} Here we adopt the structure of \nnalg  and replace the previously used deep fully-connected network by a 2-layer convolutional network. 
%\nnalg  has a sublinear regret, when the width grows polynomially with $T$ and other parameters. 
We show that under the same setting as in \cref{thm:nucbregret}, for each $T$, there exists a 2-layer CNN with a sufficiently large number of channels, such that the Sup variant of \cnnalg  satisfies the same $\mathcal{O}(T^{(2d-1)/2d})$ regret rate. 
\begin{theorem} \label{thm:cnnucbregret}
Let $\delta \in (1, 0)$. Suppose $f$ lies in the RKHS of $\cntk$ with $\norm{f}_\cntk \leq B$. Set $J>1$ and $\beta_t = 2\log( 2T|\mathcal{A}|/\delta)$ constant. For any $T\geq 1$, there exists $m$ such that if $\eta = C(LmT+m\sigma^2)^{-1}$ with some universal constant $C$, then with probability greater than $1-\delta$, \supcnnalg  satisfies,
\begin{equation*}
    \begin{split}
        R(T)  = \mathcal{O}\Big( \sqrt{T}\Big(& \sqrt{\gamma_T\sigma^{-2}(\log T)^3\log(T\log T|\mathcal{A}|/\delta)}\\
        & + \sigma B\Big) \Big).
    \end{split}
\end{equation*}
\end{theorem}
The main ingredient in the proof of \cref{thm:cnnucbregret} is a convolutional variant of Lemma~\ref{lem:gd_gp} (Lemma~\ref{lem:neuralvalko1_conv}). We prove that a 2-layer CNN trained with gradient descent on the $\ell_2$ regularized loss can approximate the posterior mean of a GP with CNTK covariance, calculated from noisy rewards. To this end, we show that training $f_{\text{CNN}}(\vx; \vtheta)$ with gradient descent causes a small change in the network parameters $\vtheta$ and the gradient vector $\nabla_\vtheta f_{\text{CNN}}(\vx; \vtheta)$. Appendix \ref{app:thm6proof} presents the complete proof.
Comparing \cref{thm:nucbregret} and \cref{thm:cnnucbregret}, the assumption on $m$ is milder in the former, but the regrets for both algorithms grow at the same rate with $T$. 
We do not expect this rate to hold for convolutional networks whose depth is greater than two. In particular, deeper CNNs are no longer $\gC_d$-invariant, and our analysis for \cnnalg relies on this property for translating the results from the fully-connected setting to the convolutional case. Although the growth rate with $T$ has remained the same, the coefficients in the regret bounds are improved for the convolutional counterparts by a factor of $d^{(d-1)/2d}$. The next corollary presents this observation.
\begin{corollary} \label{cor:cnn_regret}
Under the RKHS assumption and provided that the CNN used in \supcnnalg  has enough channels, with high probability, this algorithm satisfies 
\[
R(T) = \tilde{\mathcal{O}}\left(\frac{\tCNN}{d^{(d-1)/2d}}T^{\frac{2d-1}{2d}} \right).
\]
\end{corollary}
Corollary~\ref{cor:cnn_regret} establishes the first sublinear regret bound for convolutional contextual bandits. Concurrent to our work, \citet{ban2021convolutional} give a $\gO(\gamma_T\sqrt{T})$ regret bound for a UCB-inspired algorithm that employs a CNN which has differentiable activation functions with Lipschitz derivatives. This bound suffers from an issue similar to the other works concerning neural contextual bandits, as the growth rate of $\gamma_T$ for the smooth CNN is not taken into account.
\section{Conclusion}

We proposed \nnalg, a UCB based method for contextual bandits when the context is rich or the reward function is complex. Under the RKHS assumption on the reward, and for any arbitrary sequence of contexts, we showed that the regret $R(T)$ grows sub-linearly as $\tilde{\mathcal{O}}(T^{(2d-1)/2d})$, implying convergence to the optimal policy. We extended this result to \cnnalg, a variant of \nnalg that uses a 2-layer CNN in place of the deep fully-connected network, yielding the first regret bound for convolutional neural contextual bandits.
Our approach analyzed regret for neural network based UCB algorithms through the lens of their respective kernelized methods. A key element in this approach is bounding the regret in terms of the maximum information gain $\gamma_T$. Importantly, we showed that $\gamma_T$ for both the NTK and the 2-layer CNTK is bounded by $\tilde{\mathcal{O}}(T^{(d-1)/d})$, a result that may be of independent interest. 
%Following \citet{scarlett2017lower}, we conjecture our bounds to be min-max optimal up to a $\log T$ factor. Experiments on the task of online \mnist \citep{lecun1998gradient} classification (Appendix \ref{app:experiments}) confirm our theoretical bounds and suggest that they match the practical behavior of our algorithms. 
We believe our work opens up further avenues towards extending kernelized methods for sequential decision making in a principled way to approaches harnessing neural networks.
%work motivates future neural network based methods for sequential decision making with complex side information. It suggests that classic kernel methods can be extended to more powerful ones with provable guarantees. 

\section*{Acknowledgments}% and Disclosure of Funding
This research was supported by the European Research Council (ERC) under the European Union’s Horizon 2020 research and innovation program grant agreement no. 815943. Moreover, we thank Guillaume Wang, Johannes Kirschner, Ya-Ping Hsieh, and Ilija Bogunovic for their valuable comments and feedback. 

\bibliographystyle{plainnat}
\bibliography{refs}

\begin{thebibliography}{54}
\providecommand{\natexlab}[1]{#1}
\providecommand{\url}[1]{\texttt{#1}}
\expandafter\ifx\csname urlstyle\endcsname\relax
  \providecommand{\doi}[1]{doi: #1}\else
  \providecommand{\doi}{doi: \begingroup \urlstyle{rm}\Url}\fi

\bibitem[Abbasi-Yadkori et~al.(2011)Abbasi-Yadkori, P{\'a}l, and
  Szepesv{\'a}ri]{abbasi2011improved}
Yasin Abbasi-Yadkori, D{\'a}vid P{\'a}l, and Csaba Szepesv{\'a}ri.
\newblock Improved algorithms for linear stochastic bandits.
\newblock In \emph{NIPS}, volume~11, pages 2312--2320, 2011.

\bibitem[Arora et~al.(2019)Arora, Du, Hu, Li, Salakhutdinov, and
  Wang]{arora2019exact}
Sanjeev Arora, Simon~S Du, Wei Hu, Zhiyuan Li, Russ~R Salakhutdinov, and
  Ruosong Wang.
\newblock On exact computation with an infinitely wide neural net.
\newblock In \emph{Advances in Neural Information Processing Systems}, pages
  8141--8150, 2019.

\bibitem[Auer(2002)]{auer2002using}
Peter Auer.
\newblock Using confidence bounds for exploitation-exploration trade-offs.
\newblock \emph{Journal of Machine Learning Research}, 3\penalty0
  (Nov):\penalty0 397--422, 2002.

\bibitem[Ban and He(2021)]{ban2021convolutional}
Yikun Ban and Jingrui He.
\newblock Convolutional neural bandit: Provable algorithm for visual-aware
  advertising.
\newblock \emph{arXiv preprint arXiv:2107.07438}, 2021.

\bibitem[Bastani and Bayati(2020)]{bastani2020online}
Hamsa Bastani and Mohsen Bayati.
\newblock Online decision making with high-dimensional covariates.
\newblock \emph{Operations Research}, 68\penalty0 (1):\penalty0 276--294, 2020.

\bibitem[Belkin(2018)]{belkin2018approximation}
Mikhail Belkin.
\newblock Approximation beats concentration? an approximation view on inference
  with smooth radial kernels.
\newblock In \emph{Conference On Learning Theory}, pages 1348--1361. PMLR,
  2018.

\bibitem[Berkenkamp et~al.(2021)Berkenkamp, Krause, and
  Schoellig]{berkenkamp2016bayesian}
Felix Berkenkamp, Andreas Krause, and Angela~P Schoellig.
\newblock Bayesian optimization with safety constraints: safe and automatic
  parameter tuning in robotics.
\newblock \emph{Machine Learning}, pages 1--35, 2021.

\bibitem[Bietti(2022)]{bietti2021approximation}
Alberto Bietti.
\newblock Approximation and learning with deep convolutional models: a kernel
  perspective.
\newblock In \emph{International Conference on Learning Representations}, 2022.

\bibitem[Bietti and Bach(2021)]{bietti2020deep}
Alberto Bietti and Francis Bach.
\newblock {Deep Equals Shallow for ReLU Networks in Kernel Regimes}.
\newblock In \emph{{ICLR 2021 - International Conference on Learning
  Representations}}, pages 1--22, Virtual, Austria, 2021.
\newblock URL \url{https://hal.inria.fr/hal-02963250}.

\bibitem[Bogunovic et~al.(2020)Bogunovic, Krause, and
  Scarlett]{bogunovic2020corruption}
Ilija Bogunovic, Andreas Krause, and Jonathan Scarlett.
\newblock Corruption-tolerant gaussian process bandit optimization.
\newblock In \emph{International Conference on Artificial Intelligence and
  Statistics}, pages 1071--1081. PMLR, 2020.

\bibitem[Calandriello et~al.(2019)Calandriello, Carratino, Lazaric, Valko, and
  Rosasco]{calandriello2019gaussian}
Daniele Calandriello, Luigi Carratino, Alessandro Lazaric, Michal Valko, and
  Lorenzo Rosasco.
\newblock Gaussian process optimization with adaptive sketching: Scalable and
  no regret.
\newblock In \emph{Conference on Learning Theory}, pages 533--557. PMLR, 2019.

\bibitem[Cao and Gu(2019)]{cao2019generalization}
Yuan Cao and Quanquan Gu.
\newblock Generalization bounds of stochastic gradient descent for wide and
  deep neural networks.
\newblock In \emph{Advances in Neural Information Processing Systems}, pages
  10836--10846, 2019.

\bibitem[Cao et~al.(2021)Cao, Fang, Wu, Zhou, and Gu]{cao2019towards}
Yuan Cao, Zhiying Fang, Yue Wu, Ding-Xuan Zhou, and Quanquan Gu.
\newblock Towards understanding the spectral bias of deep learning.
\newblock In Zhi-Hua Zhou, editor, \emph{Proceedings of the Thirtieth
  International Joint Conference on Artificial Intelligence, {IJCAI-21}}, pages
  2205--2211. International Joint Conferences on Artificial Intelligence
  Organization, 2021.

\bibitem[Chen and Xu(2021)]{chen2020deep}
Lin Chen and Sheng Xu.
\newblock Deep neural tangent kernel and laplace kernel have the same
  {\{}rkhs{\}}.
\newblock In \emph{International Conference on Learning Representations}, 2021.

\bibitem[Chizat et~al.(2019)Chizat, Oyallon, and Bach]{chizatlazy}
L\'{e}na\"{\i}c Chizat, Edouard Oyallon, and Francis Bach.
\newblock On lazy training in differentiable programming.
\newblock In H.~Wallach, H.~Larochelle, A.~Beygelzimer, F.~d\textquotesingle
  Alch\'{e}-Buc, E.~Fox, and R.~Garnett, editors, \emph{Advances in Neural
  Information Processing Systems}, volume~32. Curran Associates, Inc., 2019.

\bibitem[Chowdhury and Gopalan(2017)]{chowdhury2017kernelized}
Sayak~Ray Chowdhury and Aditya Gopalan.
\newblock On kernelized multi-armed bandits.
\newblock In \emph{International Conference on Machine Learning}, pages
  844--853. PMLR, 2017.

\bibitem[Chu et~al.(2011)Chu, Li, Reyzin, and Schapire]{chu2011contextual}
Wei Chu, Lihong Li, Lev Reyzin, and Robert Schapire.
\newblock Contextual bandits with linear payoff functions.
\newblock In \emph{Proceedings of the Fourteenth International Conference on
  Artificial Intelligence and Statistics}, pages 208--214. JMLR Workshop and
  Conference Proceedings, 2011.

\bibitem[Djolonga et~al.(2013)Djolonga, Krause, and Cevher]{djolonga2013high}
Josip Djolonga, Andreas Krause, and Volkan Cevher.
\newblock High-dimensional gaussian process bandits.
\newblock In \emph{Neural Information Processing Systems}, 2013.

\bibitem[Du et~al.(2019)Du, Lee, Li, Wang, and Zhai]{du2019gradient}
Simon Du, Jason Lee, Haochuan Li, Liwei Wang, and Xiyu Zhai.
\newblock Gradient descent finds global minima of deep neural networks.
\newblock In \emph{International Conference on Machine Learning}, pages
  1675--1685. PMLR, 2019.

\bibitem[Gu et~al.(2021)Gu, Karbasi, Khosravi, Mirrokni, and
  Zhou]{gu2021batched}
Quanquan Gu, Amin Karbasi, Khashayar Khosravi, Vahab Mirrokni, and Dongruo
  Zhou.
\newblock Batched neural bandits.
\newblock \emph{arXiv preprint arXiv:2102.13028}, 2021.

\bibitem[Hao et~al.(2020)Hao, Lattimore, and Wang]{hao2020high}
Botao Hao, Tor Lattimore, and Mengdi Wang.
\newblock High-dimensional sparse linear bandits.
\newblock In H.~Larochelle, M.~Ranzato, R.~Hadsell, M.~F. Balcan, and H.~Lin,
  editors, \emph{Advances in Neural Information Processing Systems}, volume~33,
  pages 10753--10763. Curran Associates, Inc., 2020.

\bibitem[Jacot et~al.(2018)Jacot, Gabriel, and Hongler]{jacot2018neural}
Arthur Jacot, Franck Gabriel, and Cl{\'e}ment Hongler.
\newblock Neural tangent kernel: Convergence and generalization in neural
  networks.
\newblock In \emph{Advances in neural information processing systems}, pages
  8571--8580, 2018.

\bibitem[Janz et~al.(2020)Janz, Burt, and Gonz{\'a}lez]{janz2020bandit}
David Janz, David Burt, and Javier Gonz{\'a}lez.
\newblock Bandit optimisation of functions in the mat{\'e}rn kernel rkhs.
\newblock In \emph{International Conference on Artificial Intelligence and
  Statistics}, pages 2486--2495. PMLR, 2020.

\bibitem[Kandasamy et~al.(2016)Kandasamy, Dasarathy, Oliva, Schneider, and
  P{\'o}czos]{kandasamy2016gaussian}
Kirthevasan Kandasamy, Gautam Dasarathy, Junier Oliva, Jeff Schneider, and
  Barnab{\'a}s P{\'o}czos.
\newblock Gaussian process optimisation with multi-fidelity evaluations.
\newblock In \emph{Proceedings of the 30th/International Conference on Advances
  in Neural Information Processing Systems (NIPS’30)}, 2016.

\bibitem[Kandasamy et~al.(2019)Kandasamy, Dasarathy, Oliva, Schneider, and
  Poczos]{kandasamy2019multi}
Kirthevasan Kandasamy, Gautam Dasarathy, Junier Oliva, Jeff Schneider, and
  Barnabas Poczos.
\newblock Multi-fidelity gaussian process bandit optimisation.
\newblock \emph{Journal of Artificial Intelligence Research}, 66:\penalty0
  151--196, 2019.

\bibitem[Kim and Paik(2019)]{kim2019doubly}
Gi-Soo Kim and Myunghee~Cho Paik.
\newblock Doubly-robust lasso bandit.
\newblock In H.~Wallach, H.~Larochelle, A.~Beygelzimer, F.~d\textquotesingle
  Alch\'{e}-Buc, E.~Fox, and R.~Garnett, editors, \emph{Advances in Neural
  Information Processing Systems}, volume~32. Curran Associates, Inc., 2019.

\bibitem[Kirschner and Krause(2019)]{kirschner2019stochastic}
Johannes Kirschner and Andreas Krause.
\newblock Stochastic bandits with context distributions.
\newblock \emph{Advances in Neural Information Processing Systems},
  32:\penalty0 14113--14122, 2019.

\bibitem[Kirschner et~al.(2020)Kirschner, Bogunovic, Jegelka, and
  Krause]{kirschner2020distributionally}
Johannes Kirschner, Ilija Bogunovic, Stefanie Jegelka, and Andreas Krause.
\newblock Distributionally robust bayesian optimization.
\newblock In \emph{International Conference on Artificial Intelligence and
  Statistics}, pages 2174--2184. PMLR, 2020.

\bibitem[Krause and Ong(2011)]{krause2011contextual}
Andreas Krause and Cheng~S Ong.
\newblock Contextual gaussian process bandit optimization.
\newblock In \emph{Advances in neural information processing systems}, pages
  2447--2455, 2011.

\bibitem[LeCun et~al.(1998)LeCun, Bottou, Bengio, and
  Haffner]{lecun1998gradient}
Yann LeCun, L{\'e}on Bottou, Yoshua Bengio, and Patrick Haffner.
\newblock Gradient-based learning applied to document recognition.
\newblock \emph{Proceedings of the IEEE}, 86\penalty0 (11):\penalty0
  2278--2324, 1998.

\bibitem[Li et~al.(2010)Li, Chu, Langford, and Schapire]{li2010contextual}
Lihong Li, Wei Chu, John Langford, and Robert~E Schapire.
\newblock A contextual-bandit approach to personalized news article
  recommendation.
\newblock In \emph{Proceedings of the 19th international conference on World
  wide web}, pages 661--670, 2010.

\bibitem[Mei et~al.(2021)Mei, Misiakiewicz, and Montanari]{mei2021learning}
Song Mei, Theodor Misiakiewicz, and Andrea Montanari.
\newblock Learning with invariances in random features and kernel models.
\newblock In Mikhail Belkin and Samory Kpotufe, editors, \emph{Proceedings of
  Thirty Fourth Conference on Learning Theory}, volume 134 of \emph{Proceedings
  of Machine Learning Research}, pages 3351--3418. PMLR, 15--19 Aug 2021.

\bibitem[Mutn{\`y} and Krause(2019)]{mutny2019efficient}
Mojm{\'\i}r Mutn{\`y} and Andreas Krause.
\newblock Efficient high dimensional bayesian optimization with additivity and
  quadrature fourier features.
\newblock \emph{Advances in Neural Information Processing Systems 31}, pages
  9005--9016, 2019.

\bibitem[Nabati et~al.(2021)Nabati, Zahavy, and Mannor]{zahavy2019deep}
Ofir Nabati, Tom Zahavy, and Shie Mannor.
\newblock Online limited memory neural-linear bandits with likelihood matching.
\newblock In Marina Meila and Tong Zhang, editors, \emph{Proceedings of the
  38th International Conference on Machine Learning}, volume 139 of
  \emph{Proceedings of Machine Learning Research}, pages 7905--7915. PMLR,
  18--24 Jul 2021.

\bibitem[Novak et~al.(2020)Novak, Xiao, Hron, Lee, Alemi, Sohl-Dickstein, and
  Schoenholz]{neuraltangents2020}
Roman Novak, Lechao Xiao, Jiri Hron, Jaehoon Lee, Alexander~A. Alemi, Jascha
  Sohl-Dickstein, and Samuel~S. Schoenholz.
\newblock Neural tangents: Fast and easy infinite neural networks in python.
\newblock In \emph{International Conference on Learning Representations}, 2020.

\bibitem[Paszke et~al.(2019)Paszke, Gross, Massa, Lerer, Bradbury, Chanan,
  Killeen, Lin, Gimelshein, Antiga, Desmaison, Kopf, Yang, DeVito, Raison,
  Tejani, Chilamkurthy, Steiner, Fang, Bai, and Chintala]{paszke19pytorch}
Adam Paszke, Sam Gross, Francisco Massa, Adam Lerer, James Bradbury, Gregory
  Chanan, Trevor Killeen, Zeming Lin, Natalia Gimelshein, Luca Antiga, Alban
  Desmaison, Andreas Kopf, Edward Yang, Zachary DeVito, Martin Raison, Alykhan
  Tejani, Sasank Chilamkurthy, Benoit Steiner, Lu~Fang, Junjie Bai, and Soumith
  Chintala.
\newblock Pytorch: An imperative style, high-performance deep learning library.
\newblock In H.~Wallach, H.~Larochelle, A.~Beygelzimer, F.~d\textquotesingle
  Alch\'{e}-Buc, E.~Fox, and R.~Garnett, editors, \emph{Advances in Neural
  Information Processing Systems 32}, pages 8024--8035. Curran Associates,
  Inc., 2019.

\bibitem[Rahimi and Recht(2008)]{rahimi2008random}
Ali Rahimi and Benjamin Recht.
\newblock Random features for large-scale kernel machines.
\newblock In \emph{Advances in neural information processing systems}, pages
  1177--1184, 2008.

\bibitem[Riquelme et~al.(2018)Riquelme, Tucker, and Snoek]{riquelme2018deep}
Carlos Riquelme, George Tucker, and Jasper Snoek.
\newblock Deep bayesian bandits showdown: An empirical comparison of bayesian
  deep networks for thompson sampling.
\newblock In \emph{International Conference on Learning Representations}, 2018.

\bibitem[Russo and Van~Roy(2016)]{russo2016information}
Daniel Russo and Benjamin Van~Roy.
\newblock An information-theoretic analysis of thompson sampling.
\newblock \emph{The Journal of Machine Learning Research}, 17\penalty0
  (1):\penalty0 2442--2471, 2016.

\bibitem[Santin and Schaback(2016)]{santin2016approximation}
Gabriele Santin and Robert Schaback.
\newblock Approximation of eigenfunctions in kernel-based spaces.
\newblock \emph{Advances in Computational Mathematics}, 42\penalty0
  (4):\penalty0 973--993, 2016.

\bibitem[Scarlett(2018)]{scarlett2018tight}
Jonathan Scarlett.
\newblock Tight regret bounds for bayesian optimization in one dimension.
\newblock In \emph{International Conference on Machine Learning}, pages
  4500--4508. PMLR, 2018.

\bibitem[Scarlett et~al.(2017)Scarlett, Bogunovic, and
  Cevher]{scarlett2017lower}
Jonathan Scarlett, Ilija Bogunovic, and Volkan Cevher.
\newblock Lower bounds on regret for noisy gaussian process bandit
  optimization.
\newblock In \emph{Conference on Learning Theory}, pages 1723--1742. PMLR,
  2017.

\bibitem[Sch{\"o}lkopf et~al.(2001)Sch{\"o}lkopf, Herbrich, and
  Smola]{scholkopf2001generalized}
Bernhard Sch{\"o}lkopf, Ralf Herbrich, and Alex~J Smola.
\newblock A generalized representer theorem.
\newblock In \emph{International conference on computational learning theory},
  pages 416--426. Springer, 2001.

\bibitem[Sessa et~al.(2019)Sessa, Bogunovic, Kamgarpour, and
  Krause]{sessa2019no}
Pier~Giuseppe Sessa, Ilija Bogunovic, Maryam Kamgarpour, and Andreas Krause.
\newblock No-regret learning in unknown games with correlated payoffs.
\newblock In H.~Wallach, H.~Larochelle, A.~Beygelzimer, F.~d\textquotesingle
  Alch\'{e}-Buc, E.~Fox, and R.~Garnett, editors, \emph{Advances in Neural
  Information Processing Systems}, volume~32. Curran Associates, Inc., 2019.

\bibitem[Sessa et~al.(2020)Sessa, Bogunovic, Krause, and
  Kamgarpour]{sessa2020contextual}
Pier~Giuseppe Sessa, Ilija Bogunovic, Andreas Krause, and Maryam Kamgarpour.
\newblock Contextual games: Multi-agent learning with side information.
\newblock \emph{Advances in Neural Information Processing Systems}, 33, 2020.

\bibitem[Shekhar et~al.(2018)Shekhar, Javidi, et~al.]{shekhar2018gaussian}
Shubhanshu Shekhar, Tara Javidi, et~al.
\newblock Gaussian process bandits with adaptive discretization.
\newblock \emph{Electronic Journal of Statistics}, 12\penalty0 (2):\penalty0
  3829--3874, 2018.

\bibitem[Srinivas et~al.(2010)Srinivas, Krause, Kakade, and
  Seeger]{srinivas2009gaussian}
Niranjan Srinivas, Andreas Krause, Sham Kakade, and Matthias Seeger.
\newblock Gaussian process optimization in the bandit setting: No regret and
  experimental design.
\newblock In \emph{Proceedings of the 27th International Conference on
  International Conference on Machine Learning}, ICML'10, page 1015–1022,
  Madison, WI, USA, 2010. Omnipress.
\newblock ISBN 9781605589077.

\bibitem[Sui et~al.(2018)Sui, Zhuang, Burdick, and Yue]{sui18a}
Yanan Sui, Vincent Zhuang, Joel Burdick, and Yisong Yue.
\newblock Stagewise safe {B}ayesian optimization with {G}aussian processes.
\newblock In \emph{Proceedings of the 35th International Conference on Machine
  Learning}, volume~80 of \emph{Proceedings of Machine Learning Research},
  pages 4781--4789. PMLR, 2018.

\bibitem[Vakili et~al.(2021{\natexlab{a}})Vakili, Khezeli, and
  Picheny]{vakili2020information}
Sattar Vakili, Kia Khezeli, and Victor Picheny.
\newblock On information gain and regret bounds in gaussian process bandits.
\newblock In \emph{International Conference on Artificial Intelligence and
  Statistics}, pages 82--90. PMLR, 2021{\natexlab{a}}.

\bibitem[Vakili et~al.(2021{\natexlab{b}})Vakili, Scarlett, and
  Javidi]{vakili2021open}
Sattar Vakili, Jonathan Scarlett, and Tara Javidi.
\newblock Open problem: Tight online confidence intervals for rkhs elements.
\newblock In \emph{Conference on Learning Theory}, pages 4647--4652. PMLR,
  2021{\natexlab{b}}.

\bibitem[Valko et~al.(2013)Valko, Korda, Munos, Flaounas, and
  Cristianini]{valko2013finite}
Michal Valko, Nathan Korda, R\'{e}mi Munos, Ilias Flaounas, and Nello
  Cristianini.
\newblock Finite-time analysis of kernelised contextual bandits.
\newblock In \emph{Proceedings of the Twenty-Ninth Conference on Uncertainty in
  Artificial Intelligence}, UAI'13, page 654–663, Arlington, Virginia, USA,
  2013. AUAI Press.

\bibitem[Yang et~al.(2020)Yang, Jin, Wang, Wang, and Jordan]{yang2020optimism}
Zhuoran Yang, Chi Jin, Zhaoran Wang, Mengdi Wang, and Michael~I. Jordan.
\newblock Bridging exploration and general function approximation in
  reinforcement learning: Provably efficient kernel and neural value
  iterations.
\newblock \emph{CoRR}, abs/2011.04622, 2020.

\bibitem[ZHANG et~al.(2021)ZHANG, Zhou, Li, and Gu]{zhang2020neural}
Weitong ZHANG, Dongruo Zhou, Lihong Li, and Quanquan Gu.
\newblock Neural thompson sampling.
\newblock In \emph{International Conference on Learning Representations}, 2021.

\bibitem[Zhou et~al.(2020)Zhou, Li, and Gu]{zhou2019neural}
Dongruo Zhou, Lihong Li, and Quanquan Gu.
\newblock Neural contextual bandits with ucb-based exploration.
\newblock In \emph{International Conference on Machine Learning}, pages
  11492--11502. PMLR, 2020.

\end{thebibliography}

    % \item  If you get warning messages as described above, then
    % immediately after $\texttt{\textbackslash
    % begin\{document\}}$, write
    % \begin{flushleft}
    % \texttt{\textbackslash runningtitle\{Provide here an alternative
    % shorter version of the title of your paper\}}\\
    % \texttt{\textbackslash runningauthor\{Provide here the surnames of
    % the authors of your paper, all separated by commas\}}
    % \end{flushleft}
    % Note that the text that appears as argument in \texttt{\textbackslash
    %   runningtitle} will be printed as a heading in the \emph{even}
    % pages. The text that appears as argument in \texttt{\textbackslash
    %   runningauthor} will be printed as a heading in the \emph{odd}
    % pages.  If even the author surnames do not fit, it is acceptable
    % to give a subset of author names followed by ``et al.''

%%%%%%%%%%%%%%%%%%%%%%%%%%%%%%%%%%%
%%%%%% SUPPLEMENT (OPTIONAL) %%%%%%
%%%%%%%%%%%%%%%%%%%%%%%%%%%%%%%%%%%

\clearpage
\appendix
\numberwithin{equation}{section}
\thispagestyle{empty}

% For one-column format, uncomment the following:
\onecolumn \makesupplementtitle
% For two-column format, uncomment the following:
%\twocolumn[ \makesupplementtitle 

% {\centering
%   {\Large\bfseries Neural Contextual Bandits without Regret:\\ Supplementary Material\par }}
%   \vskip -0.05in
%   \bottomtitlebar
 \section{Experiments} \label{app:experiments}
We carry out experiments on the task of online \mnist \citep{lecun1998gradient} classification, to assess how well our analysis of information gain and regret matches the practical behavior of our algorithms. We find that the UCB algorithms exhibit a fairly consistent behavior while solving this classification task.
%, hence the plots are given without error bars. 
In Section \ref{app:interp} we design two experiments to demonstrate that $\hat\sigma_{t-1}$ and the approximate UCB defined for \nnalg{}, are a meaningful substitute for the posterior variance and upper confidence bound for \ntkalg. 
%We provide the code at \texttt{https://github.com/pkassraie/NNUCB}.
\subsection{Technical Setup} \label{app:exp_setup}
We formulate the \mnist classification problem such that it fits the bandit setting, following the setup of \citet{li2010contextual}. To create the context matrix $\vz$ from a flattened image $\tilde{\vz} \in \sR^{d}$, we construct 
\begin{equation*}
   \vz^T =  \begin{bmatrix}
\tilde{\vz}^T & \bm{0} & \cdots & \bm{0} \\
 \bm{0} & \tilde{\vz}^T & \cdots & \bm{0} \\
%\bm 0 & \cdots & \cdots & \bm 0 \\
\vdots &\vdots&\vdots& \vdots\\
 \bm{0} & \cdots & \bm{0} &  \tilde{\vz}^T \\
\end{bmatrix}\in \sR^{K\times Kd}
\end{equation*}
where $K= \vert \mathcal{A}\vert = 10$ in the case of \mnist dataset. The action taken $\va$ is a one-hot $K$-dimensional vector, indicating which class is selected by the algorithm. The context vector corresponding to an action $i$ is
\begin{equation*}
    \vx_i = \vz \va_i = \big[ 0 \qquad \cdots \tilde \vz \cdots \qquad 0\big] \in \sR^{Kd}
\end{equation*}
where $\tilde \vz$ occupies indices $(i-1)d$ to $id$. At every step, the context is picked at random and presented to the algorithm. If the correct action is picked, then a noiseless reward of $y_t = 1$ is given, and otherwise $y_t = 0$.
\paragraph{Training Details} For \ntkalg{} and \cntkalg{} we use the implementation of the NTK from the \textsc{Neural Tangents} package \citep{neuraltangents2020}. \textsc{PyTorch} \citep{paszke19pytorch} is used for defining and training the networks in \nnalg{} and \cnnalg{}. 
To calculate the approximate UCB, we require the computation of the empirical gram matrix $\mG^T\mG$ where $\mG^T = [\vg(\vx_t; \vtheta^0)]_{t\leq T}$.
To keep the computations light, we always use the diagonalized matrix as a proxy for $\mG^T\mG$. Algorithm \ref{alg:approx_ntkucb} implies that at every step $t$ the network should be trained from initialization. In practice, however, we use a Stochastic Gradient Descent (SGD) optimizer. To train the network at step $t$ of the algorithm, we consider the loss summed over the data points observed so far, and run the optimizer on $f(\vx; \vtheta_{t-1})$ rather than starting from $f(\vx; \vtheta^0)$. At every step $t$, we stop training when $J$ reaches $1000$, or when the average loss gets small, i.e.,  $\gL(\vtheta^J)/(t-1) \leq 10^{-3}$.
While $t \leq  1000$ we train at every step, for $t>1000$, however, we train the network once every $100$ steps. 
\paragraph{Hyper-parameter Tuning} For the simple task of \mnist classification, we observe that the width of the network does not significantly impact the results, and after searching the exponential space $m \in \{64, 128, 512, 1024\}$ we set $m = 128$ for all experiments.
The models are not extensively fine-tuned and hyper-parameters of the algorithms, $\beta, \sigma$, are selected after a light search over $\{10^{-2k}, 0 \leq k  \leq 5 \}$, and vary between plots. For all experiments, we set the learning rate of the SGD optimizer to $\eta = 0.01$. 
\subsection{Growth rates: Empirical vs Theoretical}
\begin{figure}
    \centering
    \begin{minipage}[t]{0.49\textwidth}
        \centering
\includegraphics[width=\textwidth]{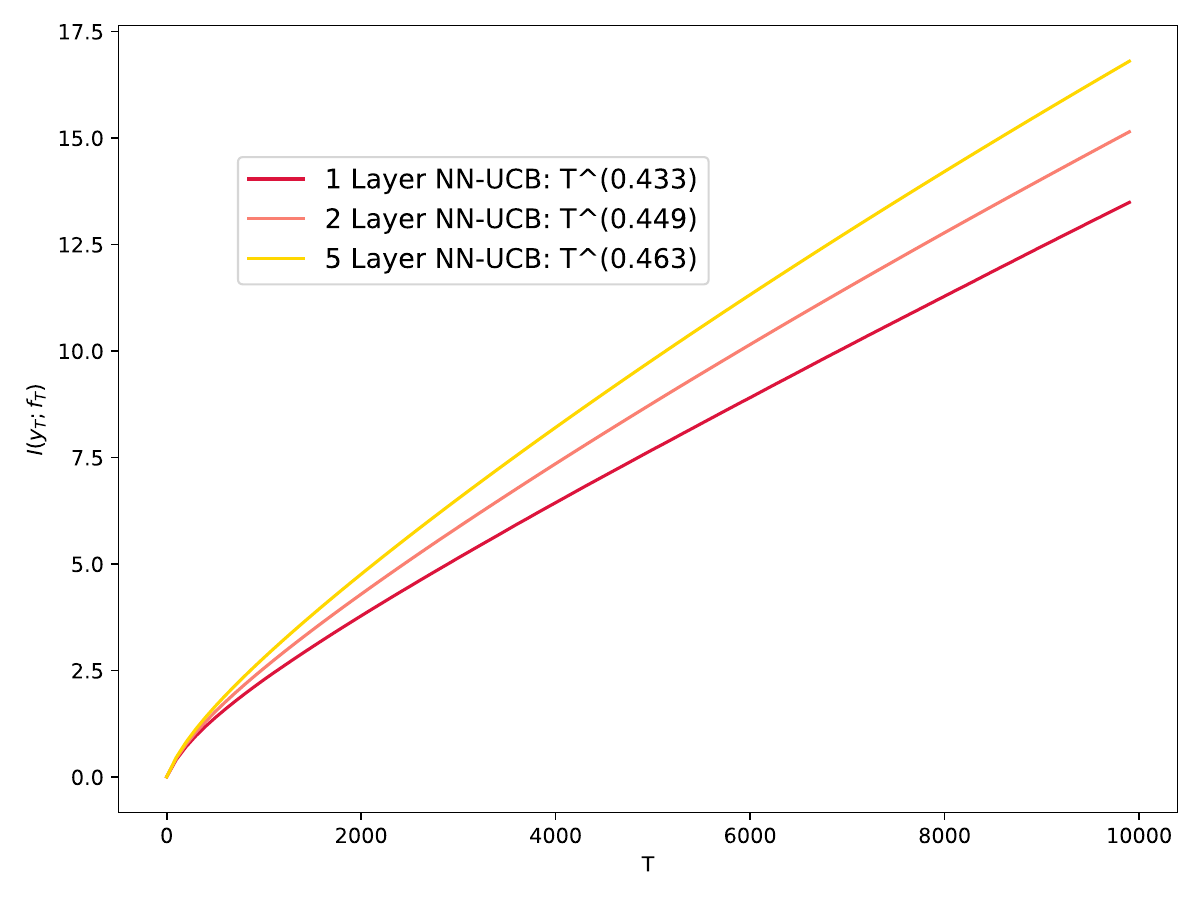}
    \caption{\label{fig:infogain_antkvnucb}
    Empirical information gain of \nnalg{} with networks of various depths, on the online \mnist classification problem, is of rate $O(T^{0.45})$. Our bound on the maximum information gain (Theorem \ref{thm:info_gain}) grows as $ \tilde{\mathcal{O}}(T^{(d-1)/d})$.}
    \end{minipage}\hfill
    \begin{minipage}[t]{0.49\textwidth}
\centering
\includegraphics[width=\textwidth]{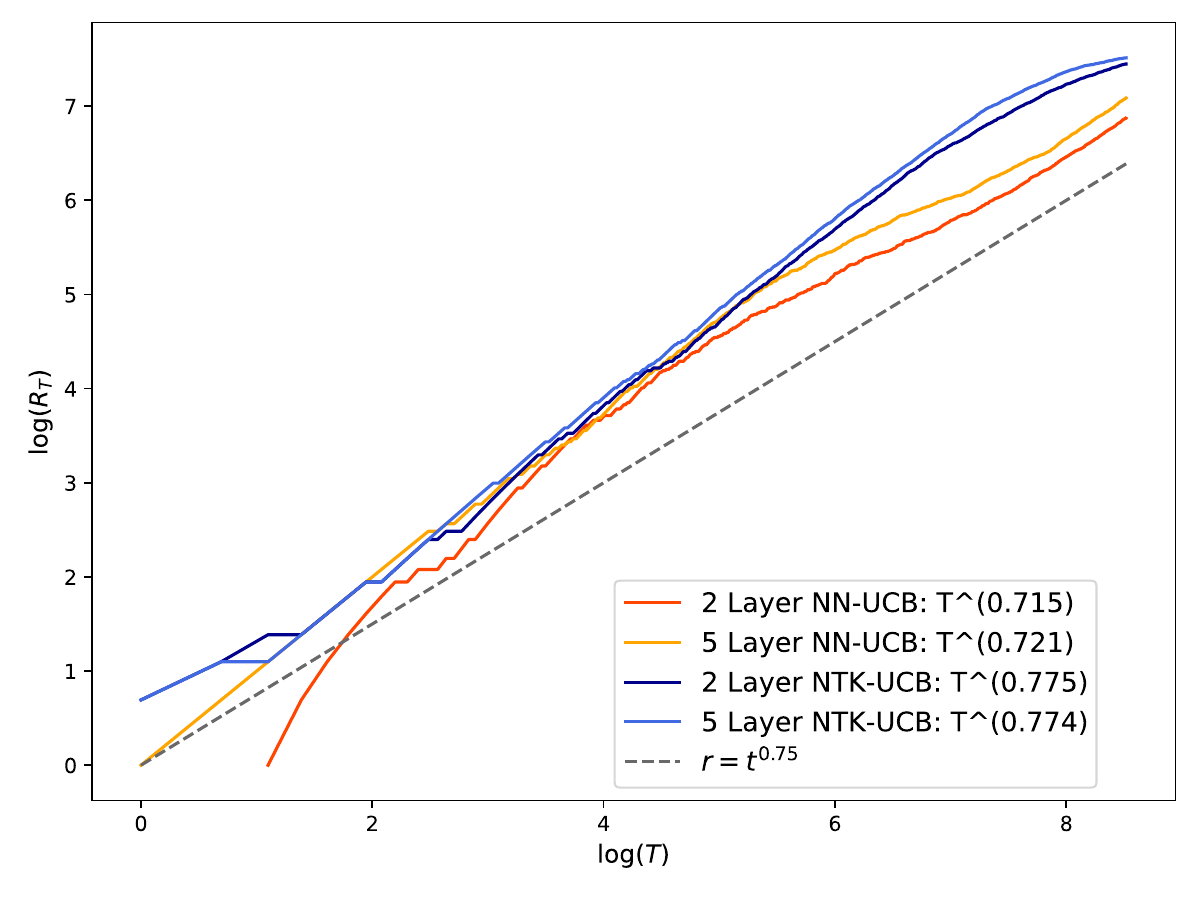}
        \caption{\label{fig:regret_allntks} 
        The regret of \ntkalg{} and \nnalg{} with networks of various depths, on the online \mnist classification problem, is of rate $O(T^{0.75})$. In the worst case it grows as $\tilde{\mathcal{O}}(T^{(2d-1)/2d})$ (Corollaries \ref{cor:NTK_regret}~\&~\ref{cor:nucbregret}).}
    \end{minipage}
\end{figure}

We test our algorithms on the online \mnist classification task. We plot the empirical information gain and regret to verify the tightness of our bounds. For the information gain of the \nnalg{}, we let the algorithm run for $T=10000$ steps. Then we take the sequence $(\vx_t)_{t\leq T}$ from this run and plot $\hat I(\vy_t; \vf_t)$ the empirical information gain against time $t$, where
$
\hat I(\vy_t; \vf_t) = \frac{1}{2}\log\det (\mI + \sigma^{-2}\mG_t^T\mG_t)
$, with $\mG_t^T = [\vg(\vx_\tau; \vtheta^0)]_{\tau\leq t}$. Figure \ref{fig:infogain_antkvnucb} shows the growth of $\hat I$ for \nnalg{} with networks of various depth. To calculate the empirical growth rate (given in the figure's labels), we fit a polynomial to the curve, and obtain a rate of roughly $\mathcal{O}(T^{0.45})$. Our theoretical rate is $\tilde{\mathcal{O}}(T^{(d-1)/d})$, where $d=10\times784$ is the dimension of $\vx_t$. Note that the theoretical rate bounds the information gain for arbitrary sequences of contexts including adversarial worst cases. Moreover, in the case of MNIST, the contexts reside on a low-dimensional manifold. As for the regret, we run \nnalg{} and \ntkalg{} for $T=5000$ steps and plot the regret, which in the case of online \mnist classification, shows the number of misclassified digits. Fitting a polynomial to the data from Figure \ref{fig:regret_allntks} gives an empirical rate of $O(T^{0.75})$. Our theoretical bound for these algorithms grows as $\mathcal{O}(T^{(2d-1)/2d})$.
%Figure \ref{fig:regret_allntks} suggest that the $2$-layer network (resp. the $2$-layer NTK) outperforms the deeper one. We conjecture that this behavior is an artifact of approximating the gram matrix $\mG^T\mG \in \sR^{p\times p}$ with its diagonal. For the $5$-layer network, the number of parameters $p$ is larger and the diagonal approximation is not as representative.  

\begin{figure}
\centering
\begin{minipage}[t]{0.49\textwidth}
\includegraphics[width=\textwidth]{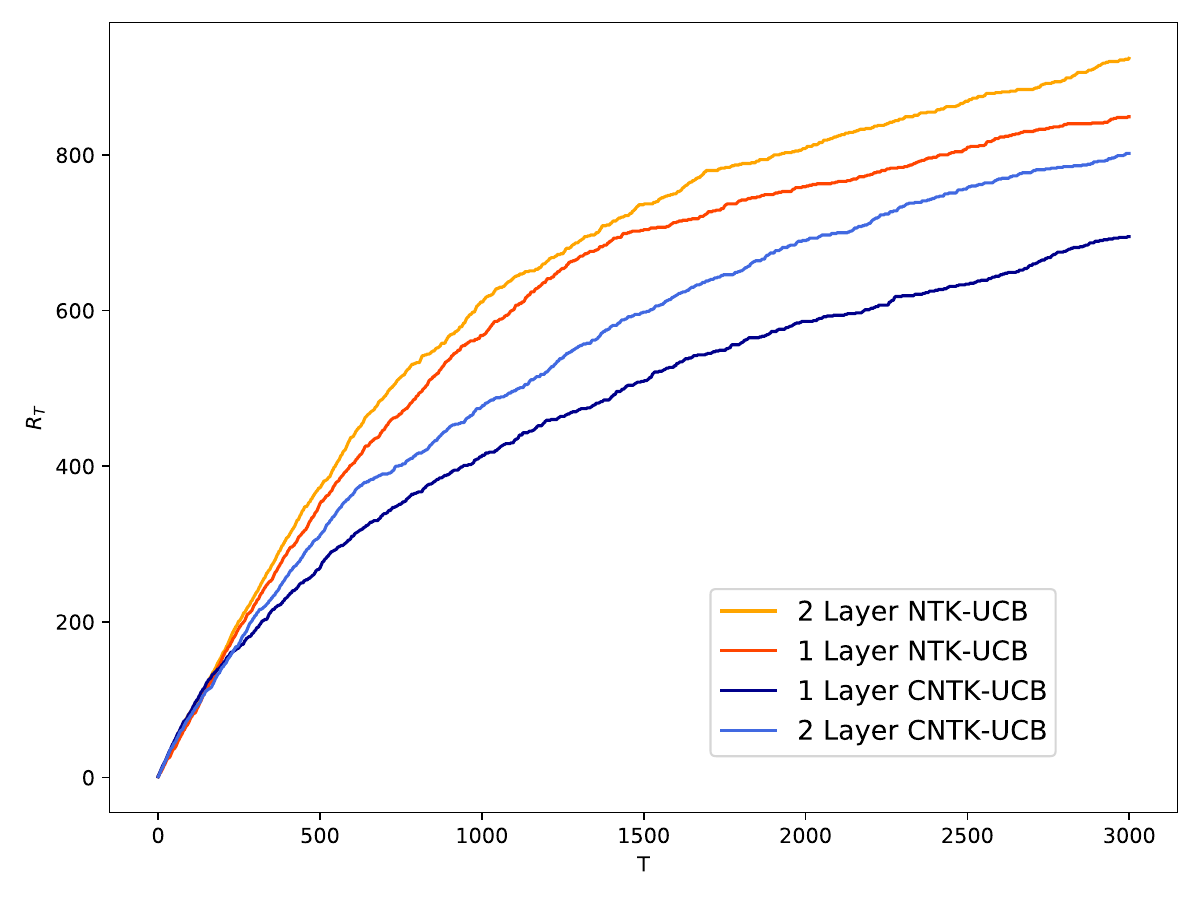}
\caption{\label{fig:cnnplot}\looseness -1  \nnalg{} vs. \cnnalg{} for online \mnist{} classification. Both algorithms exhibit a similar growth rate with $T$, while \cnnalg{} outperforms \nnalg{}, as described in Corollary~\ref{cor:cntk_regret}.}
\end{minipage}\hfill
\begin{minipage}[t]{0.49\textwidth}
\includegraphics[width=\textwidth]{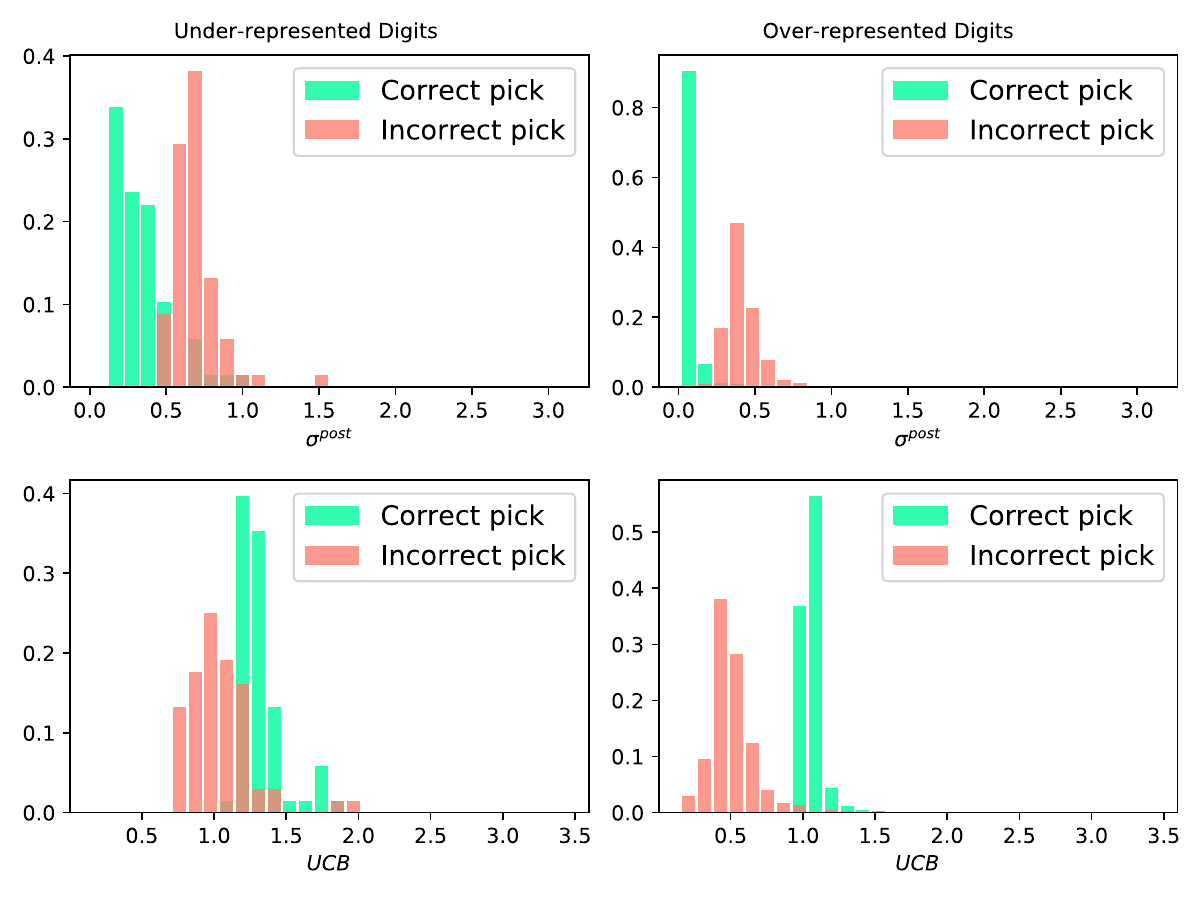}
\caption{\label{fig:underrep}Under-representation Test. Histograms of $\hat \sigma_{t}$ and approximate UCB are plotted for an imbalanced dataset of zeros and ones. In lack of data, \nnalg chooses the sub-optimal action with a high confidence.}
\end{minipage}
\end{figure}

In Section \ref{sect:cnn}, we conclude that the information gain and the regret grow with the same rate for both \ntkalg{} and its convolutional variant. \cntkalg{} however, tends to have a smaller regret for every $T$. This is due to the fact that $R(T) = \tilde{\mathcal{O}}(\tCNN T^{(2d-1)/2d})$ for the \ntkalg{}, while for the convolutional counterpart $R(T) = \tilde{\mathcal{O}}(\tCNN T^{(2d-1)/2d}/d^{(d-1)/2d})$ and $d\geq 1$. The upper bound on the regret being tighter for \cntkalg{} does not imply that the regret will be smaller as well. In Figure \ref{fig:cnnplot} we present both algorithms with the same set of contexts, and investigate whether in practice the convolutional variant can outperform \ntkalg, which seems to be the case for the online \mnist{} classification task. 
\subsection{\nnalg{} in the face of Uncertainty}\label{app:interp}
\begin{figure}
\centering
\begin{minipage}[t]{0.344\textwidth}
\includegraphics[width=\textwidth]{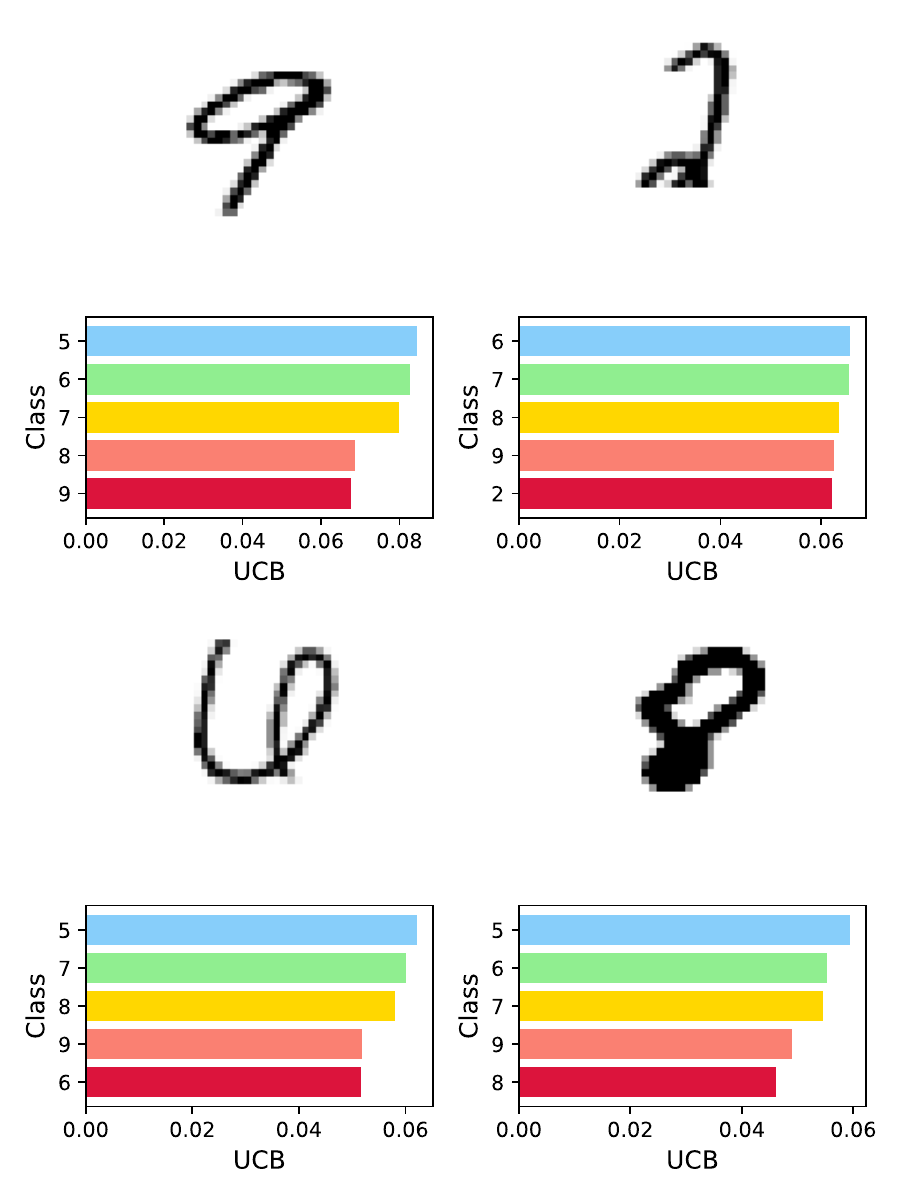}
\caption{\label{fig:ambig_digits} Examples of ambiguous Digits and the top 5 choices of \nnalg with the largest UCBs.}
\end{minipage}\hfill
\begin{minipage}[t]{0.60\textwidth}
\includegraphics[width=0.9\textwidth]{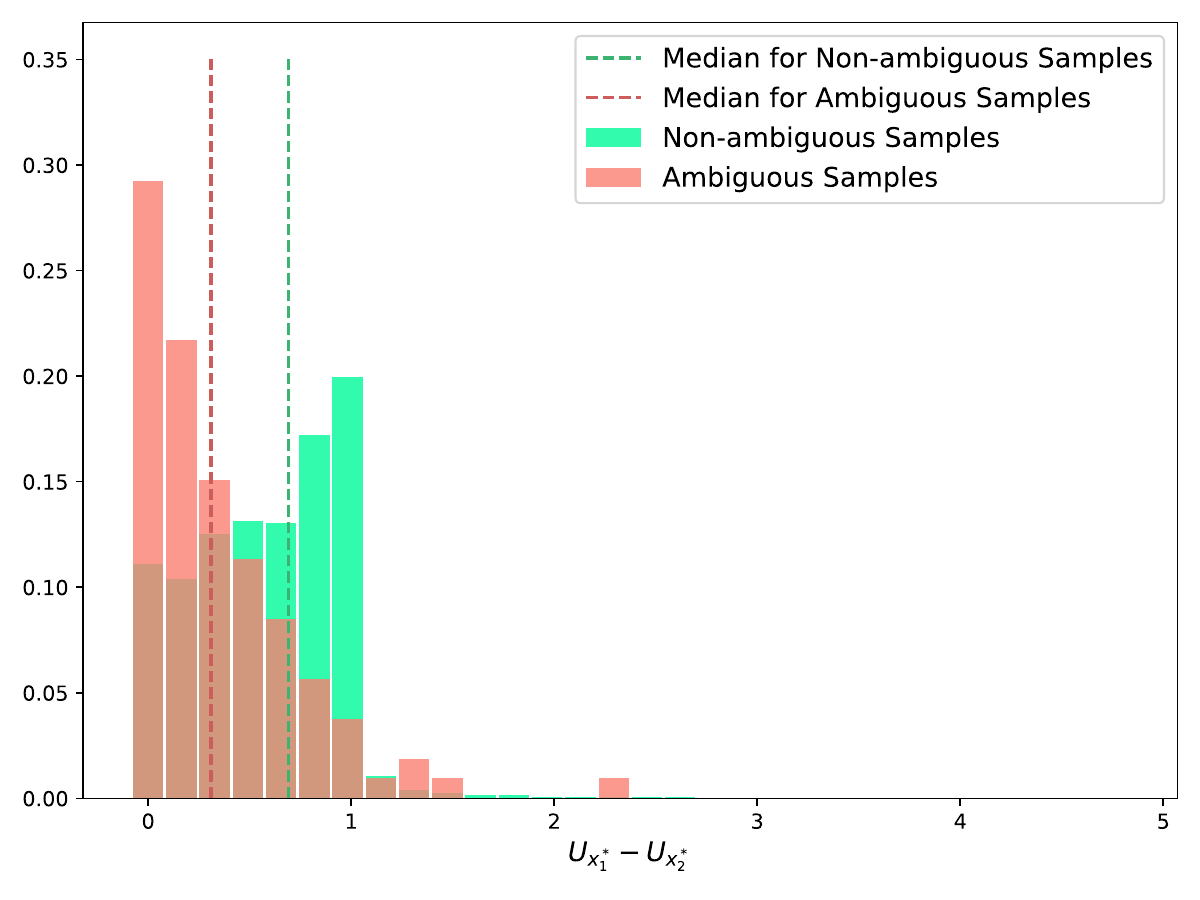}
\caption{\label{fig:ambig_ucb_dif}Ambiguity Test. Histogram of $U_{\vx^*_1} - U_{\vx^*_2}$ is plotted. For non-ambiguous digits, \nnalg{} is significantly more confident about the class it picks.}
\end{minipage}\hfill
\end{figure}

\looseness -1 The posterior mean and variance have a transparent mathematical interpretation. For designing \nnalg{}, however, we approximate $\mu_{t-1}$ and $\sigma_{t-1}$ with $f^{(J)}_{t-1}$ and $\hat \sigma_{t-1}$ which are not as easy to interpret. We design two experiments on \mnist{} to assess how well this approximation reflects the properties of the posterior mean and variance. 
\paragraph{Effect of Imbalanced Classes} For this experiment, we limit \mnist{} to only zeros and ones. We create a dataset with underrepresented zeros, such that the ratio of class populations is $1:20$. This experiment shows that $\hat\sigma_{t-1}$ the approximate posterior variance of \nnalg{} behaves as expected, similar to $\sigma_{t-1}$. The setup is as follows. Using $80\%$ of this dataset, we first run \nnalg{} and train the network. On the remaining $20\%$, we continue to run the algorithm; no longer training the network, but still updating the posterior variance at every step.
Presented in Figure \ref{fig:underrep}, we plot the histogram of the posterior variance and upper confidence bound during the \emph{test} phase.
The first row presents histograms for $\sigma^{\mathrm{(post)}}(\vx_t^*)$, the posterior variance calculated for the true digit $\vx_t^*$ at step t.
The lower plots show histograms of the corresponding $\mathrm{UCB}(\vx_t^*)$ values.
 We give two separate plots for the times when the true digit $\vx^*$ comes from the under-represented class, and when it comes from the over-represented class, respectively in the left and right columns.
 In every plot, to distinguish between the steps at which the digit picked by \nnalg{} matches the true digit or not, we make a separate histogram for each case. 
Figure \ref{fig:underrep} demonstrates that when the ground truth is over-represented and action is picked correctly, the algorithm always has a high confidence (large UCB) and small uncertainty (posterior variance) for the true class. When the true class is over-represented, for both $\sigma^{\mathrm{post}}$ and UCB, the histogram of the steps with correct picks (green) is concentrated and well separated from the histogram of the steps with incorrect picks (red). This implies that the reason behind misclassifying an over-represented digit is having a large variance, and the algorithm is effectively performing exploration because the estimated reward is small for every action. For the under-represented class, however, the red and green histograms are not well separated. Figure \ref{fig:underrep} shows that sometimes an under-represended digit has been misclassified with a small posterior variance, or a large UCB.
In lack of data, the learner is not able to refine its estimation of the posterior mean or variance. It is forced to do explorations more often which results in incorrect classifications. 
\paragraph{Effect of Ambiguous Digits} We also assess the ability of the UCB to quantify uncertainty in light of ambiguous \mnist samples. To this end, we define an ambiguous digit to be a data point that is classified incorrectly by a well-trained classifier. We first train a $2$-layer CNN with $80\%$ of the \mnist dataset as the standard \mnist classifier. The rest of the data we use for testing. We save the misclassified digits from the test set to study \nnalg. Figure \ref{fig:ambig_digits} shows a few examples of such digits and the UCB for the top 5 choices of \nnalg after observing the digit. It can be seen that for these digits there is no action which the algorithm can pick with high confidence.
We first run \nnalg on the training set. On the test set, we continue to run \nnalg, however we do not train network any further and only update the posterior variance. For any digit, let $\vx^*_1$ denote the maximizer of the UCB, and $\vx^*_2$ be the class with the second largest UCB value. Figure \ref{fig:ambig_ucb_dif} shows the histogram of $U_{\vx^*_1} - U_{\vx^*_2}$. The red histogram is for ambiguous samples and the green one for the non-ambiguous ones. Looking at the medians, we see that for clear samples, the algorithm often has a high confidence on its choice, while this is not the case for the ambiguous digits. 

\section{Details of the Main Result} \label{app:insight}
Here we elaborate on a few matters from the main text. 
 \subsection{On Section \ref{sect:f_assumptions}: Connections Between GP and RKHS Assumptions} \label{app:gp_vs_rkhs}
We explain how the GP and RKHS assumption imposes smoothness on the reward.
By assuming that $f \sim \mathrm{GP}(0, k)$, we set $\Cov(f(\vx),f(\vx'))$ to $k(\vx,\vx')$. In doing so, we enforce smoothness properties of $k$ onto $f$. As an example, suppose some normalized kernel $k$ satisfies boundedness or Lipschitz-continuity, then for $\norm{\vx-\vx'}\leq \delta$, $k(\vx,\vx')$ is close to $k(\vx,\vx) = 1$. The GP assumption then ensures high correlation for value of $f$ at these points, making a smooth $f$ more likely to be sampled. The NTK is rotationally invariant and can be written as $\kappa(\vx^T\vx')$, where $\kappa$ is continuous and $C^\infty$ over $(-1,1)$, but is not differentiable at $\pm 1$ \citep{bietti2020deep}. Therefore, our GP assumption only implies that it is more likely for $f$ to be continuous. Regarding the RKHS assumption, the Stone-Weierstrass theorem shows that any continuous function can be uniformly approximated by members of $\mathcal{H}_\ntk$. We proceed by laying out the connection between the two assumptions in more detail.

Equipped with the Mercer's theorem, we can investigate properties of $f \sim \text{GP}(0,k)$, in the general case where $k$ is a Mercer kernel and $\mathcal{X}$ is compact. The following proposition shows that sampling $f$ from a GP is equivalent to assuming $f = \sum_i \beta_i \phi_i$, and sampling the coefficients $\beta_i$ from $\mathcal{N}(0,\lambda_i)$, where $\phi_i$ and $\lambda_i$ are the eigenfunctions and eigenvalues of the GP's kernel function.
\begin{prop}\label{prop:f_gp}
Let $k$ to be a Hilbert-Schmidt continuous positive semi-definite kernel function, with $(\lambda_i)_{i=1}^\infty$, and $(\phi_i)_{i=1}^\infty$ indicating its eigenvalues and eigenfunctions. Assume $\mathcal{X}$ is compact. If $f \sim \text{GP}(0,k)$, then $f = \sum_i \beta_i \phi_i$, where $\beta_i \stackrel{\text{i.i.d}}{\sim} \mathcal{N}(0, \lambda_i)$.
\end{prop}
\begin{proof}
It suffices to show that if $f = \sum_i \beta_i \phi_i$, then for any $\vx,\, \vx' \in \mathcal{X}$, we have:
\[\mathbb{E}f(\vx) = 0 , \quad \mathbb{E}f(\vx)f(\vx') = k(\vx, \vx'). \]
Since $\beta_i$ are Gaussian and independent,
\begin{align*}
 &\mathbb{E} f(\vx) = \sum_i \phi_i(\vx) \,\mathbb{E}\beta_i = 0,\\
&\mathbb{E} f(\vx)f(\vx') = \sum_i \phi_i(\vx)\phi_i(\vx') \,\mathbb{E}\beta_i^2 = \sum_i \lambda_i \phi_i(\vx)\phi_i(\vx') = k(\vx, \vx').
\end{align*}
Here we have used orthonormality of $\phi_i$s.
\end{proof}
Proposition \ref{prop:f_gp} suggests that if $f \sim \mathrm{GP}(0,k)$ then $\norm{f}_k$ is almost surely unbounded, since by the definition of the inner product on $\gH_k$ we have
\[
\mathbb{E} \vert\vert f\vert\vert_k^2 = \sum_i \mathbb{E} \frac{\beta_i^2}{\lambda_i} = \sum_i \frac{\lambda_i}{\lambda_i}.
\]
This expectation is unbounded for any kernel with an infinite number of nonzero eigenvalues. Therefore, with probability one, $\vert\vert f\vert\vert_k$ is unbounded and not a member of $\gH_k$. However, the posterior mean of $f$ after observing $T$ samples lies in $\gH_k$ (Proposition \ref{lem:posterior_mean}). This connection implies that our estimate of $f$, under both RKHS and GP assumption will be a $k$-norm bounded function, similarly reflecting the smoothness properties of the kernel. 
\begin{prop} \label{lem:posterior_mean}
Assume $f \sim \text{GP}(0, k)$, with $k$ Mercer and $\mathcal{X}$ compact. Then $\mu_{T}$ the posterior mean of $f$ given $(\vx_i, y_i)_{i=1}^T$ has bounded $k$-norm,  and $\mu_{T} \in \mathcal{H}_k$.
\end{prop}
\begin{proof}
We first recall the Representer Theorem \citep{scholkopf2001generalized}. Consider the loss function $J(f) = Q(\vf_T; \vy_T) + \sigma^2 \vert\vert f \vert \vert_k$, where $\vf_T = [f(\vx_1), \cdots, f(\vx_T)]^T$. $Q$ is a $\ell_2$-loss assessing the fit of $\vf_T$ to $\vy_T$. Then $J(f)$ has a unique minimizer, which takes the form:
\[
\hat{f} = \sum_{i=1}^T \alpha_i k(\vx_i, \cdot)
\]
Note that this sum is finite, hence $\hat f$ is $k$-norm bounded and in $\mathcal{H}_k$. Minimizing $J$ over $\bm{\alpha}$, we get
\[
\hat{\bm{\alpha}} = (\bm{K}_T+ \sigma^2 \bm{I})^{-1}\vy_T
\]
Indicating that $\hat{f}(\vx) = \mu_T(\vx)$.
\end{proof}
% \subsection{On Section \ref{sect:info_gain}: Information Gain as a Measure of Complexity}\label{app:infogain_complexity}
% Let $\lambda_i(\mNTK) \geq 0$ denote the eigenvalues of $\mNTK$. Then,
% \[
% I(\vy_T; \vf_T) = \frac{1}{2}\log\det (\mI + \sigma^{-2}\mNTK) = \frac{1}{2} \sum_{i=1}^T \log (1 + \sigma^{-2}\lambda_i) \leq  \frac{1}{2} \sigma^{-2}\sum_{i=1}^T \lambda_i.
% \]
% There are two ways to interpret the equation above. 
% Here we see that the information gain is controlled by this eigendecay; a rapid decay of eigenvalues of $k$ would imply faster decay of eigenvalues of $\mK$, and a smaller $I(\vy_T; \vf_T)$. The eigendecay of a kernel also controls the complexity of its RKHS (Equation \ref{eq:ntk_rkhs}). Therefore, the information gain can be viewed as a proxy for the complexity of the RKHS of its kernel. If $\gamma_T$ grows rapidly with $T$, then the corresponding function class is more complex. Another interpretation is that the information gain is a proxy for $\mathrm{Rank}(\mK)$. If $\mK$ is normalized w.r.t its spectral norm, i.e. $\lambda_i(\mK) \leq 1$, then the information gain is a convex surrogate for $\text{Rank}(\mK)$.  Let $\bm{\phi}(\vx) \in \sR^p$ be the feature map corresponding to $k$. For simple under-parametrized models with $p \leq T$, the rank itself is a notion of complexity for simpler models. 
\subsection{On Assumptions of Theorem \ref{thm:nucbregret}}
\label{app:nn_additional_assump}
For technical simplicity, in Theorem \ref{thm:nucbregret} and its convolutional extension, Theorem \ref{thm:cnnucbregret}, we require the network at initialization to satisfy $f(\vx; \vtheta^0)=0$ for all $\vx \in \mathcal{X}$.
We explain how this assumption can be fulfilled without limiting the problem setting to a specific network or input domain. 
 Without loss of generality, we can initialize the network as follows. For $l \leq L$ let the weights at initialization be,
\[ W^{(l)} = 
\begin{pmatrix}
W & 0\\
0 & W
\end{pmatrix}, \quad W^{(L+1)} = \big( \vw^T, -\vw^T\big)
\]
where entries of $W$ and $\vw$ are i.i.d and sampled from the normal distribution. Moreover, we assume that $[\vx]_j = [\vx]_{j + d/2}$ for any $\vx = \vz_t\va$ where $1 \leq t \leq T$ and any $\va \in \mathcal{A}$. Any input $\vx$ can be converted to satisfy this assumption by defining an auxiliary input  $\tilde\vx = [\vx, \vx]/\sqrt{2}$ in a higher dimension. This mapping together with the initialization method, guarantee that output of the network at initialization is zero for every possible input that the learner may observe during a $T$-step run of the algorithm.  Essentially, this property comes into effect when using result from \citet{arora2019exact} or working with the Taylor expansion of the network around initialization. It allows us to write $f(\vx_i; \vtheta) \approx \langle\vg^T(\vx_i; \vtheta^0),\vtheta-\vtheta^0\rangle$.
\subsection{On Section \ref{sect:cnn}: CNTK as an averaged NTK} \label{app:cntk_vs_ntk}
This section gives an intuition on, and serves as a sketch for proving Equation \ref{eq:cntk_def}. Consider a 2-layer ReLU network with width $m$ defined as,
\[
f_{\text{NN}}(\vx; \vtheta^{(0)}) = \frac{2}{\sqrt{m}} \sum_{i=1}^m v_i^{(0)} \sigma(\langle \vw_i^{(0)}, \vx\rangle)
\]
where each weight parameter is an i.i.d sample from $\mathcal{N}(0,1)$. We denote the complete weight vector by $\vtheta$ and write the first order Taylor approximation of this function with respect to $\vtheta$ around the initialization.
\begin{align*}
    f_{\text{NN}}(\vx; \vtheta) \simeq f_{\text{NN}}(\vx; \vtheta^{(0)}) & + \overbrace{\frac{2}{\sqrt{m}}\sum_{i=1}^m (v_i - v_i^{(0)}) \sigma(\langle \vw_i^{(0)}, \vx\rangle)}^{f_{\text{NN},1}(\vx; \vv)}\\
    & + \underbrace{\frac{2}{\sqrt{m}}\sum_{i=1}^m v_i^{(0)} \dot\sigma(\langle \vw_i^{(0)}, \vx\rangle) \langle \vw_i-\vw_i^{(0)}, \vx \rangle}_{f_{\text{NN},2}(\vx; \mW)}
\end{align*}
We limit the input domain to $\sS^{d-1}$, equipped with the uniform measure. Consider the function class $\mathcal{F}_{\text{NTK}} =\{ f(\vx)=f_{\text{NN},1}(\vx; \vv)+f_{\text{NN},2}(\vx; \mW)\, \text{ s.t. } \,  \vv \in \sR^{m},\, \mW \in \sR^{m \times d} \}$. It is straightforward to show that $\mathcal{F}_{\text{NTK}}$ is an RKHS and the following kernel function satisfies the reproducing property for $\gF_{\text{NTK}}$ \citep{chizatlazy}. 
\[
h(\vx, \vx') = \frac{1}{m} \sum_{i=1}^m \sigma(\langle \vw_i^{(0)}, \vx\rangle)\sigma(\langle \vw_i^{(0)}, \vx' \rangle) +  \frac{1}{m} \sum_{i=1}^m \vx^T\vx' (v_i^{(0)})^2\dot\sigma(\langle \vw_i^{(0)}, \vx\rangle)\dot\sigma(\langle \vw_i^{(0)}, \vx' \rangle)
\]
\paragraph{Random Feature Derivation of 2-layer CNTK} We repeat the first order Taylor approximation for $f_{\text{CNN}}$. 
\begin{align*}
    f_{\text{CNN}}(\vx; \vtheta) \simeq f_{\text{CNN}}(\vx; \vtheta^{(0)}) & + \overbrace{\frac{2}{\sqrt{m}}\sum_{i=1}^m (v_i - v_i^{(0)}) \big[ \frac{1}{d} \sum_{l = 1}^{d} \sigma(\langle \vw_i^{(0)}, c_l \cdot \vx\rangle)\big]}^{f_{\text{CNN,}1}(\vx; \vv)}\\
    & + \underbrace{\frac{2}{\sqrt{m}}\sum_{i=1}^m v_i^{(0)} \big[ \frac{1}{d} \sum_{l = 1}^{d} \dot\sigma(\langle \vw_i^{(0)},  c_l \cdot\vx\rangle) \langle \vw_i-\vw_i^{(0)}, c_l \cdot\vx \rangle\big]}_{f_{\text{CNN,}2}(\vx; \mW)}
\end{align*}
And define $\mathcal{F}_{\text{CNTK}}$ to be the convolutional counterpart of $\mathcal{F}_{\text{NTK}}$
\[
\mathcal{F}_{\text{CNTK}} =\{ f(\vx)=f_{\text{CNN,}1}(\vx; \vv)+f_{\text{CNN,}2}(\vx; \mW)\, \text{ s.t. } \,  \vv \in \sR^{m},\, \mW \in \sR^{m \times d} \}.
\]
Then $\mathcal{F}_{\text{CNTK}}$ is reproducing for the following kernel function,
\begin{align*}
    \bar h(\vx, \vx') & = \frac{1}{m} \sum_{i=1}^m \frac{1}{d^2} \sum_{l = 1}^{d}  \sum_{l' = 1}^{d}   \sigma(\langle \vw_i^{(0)}, c_l \cdot \vx\rangle)\sigma(\langle  \vw_i^{(0)},c_{l'}\cdot \vx' \rangle)\\
    & \quad +  \frac{1}{m} \sum_{i=1}^m \frac{1}{d^2} \sum_{l = 1}^{d}  \sum_{l' = 1}^{d}  \langle c_l \cdot \vx, c_{l'}\cdot \vx'\rangle (v_i^{(0)})^2 \dot\sigma(\langle \vw_i^{(0)}, c_l \cdot \vx\rangle)\dot\sigma(\langle \vw_i^{(0)},c_{l'}\cdot  \vx' \rangle)
\end{align*}
As $m$ the number of channels grows, the average over the parameters converges to the expectation over $\vw$ and $\vv$, and the kernel becomes rotation invariant, i.e. only depends on the angle between $\vx$ and $\vx'$. 
\begin{align*}
    \lim_{m \rightarrow \infty}\bar h(\vx, \vx') & = \lim_{m \rightarrow \infty} \frac{1}{d} \frac{1}{m} \sum_{i=1}^m  \sum_{l = 1}^{d} \sigma(\langle \vw_i^{(0)}, c_l \cdot \vx\rangle)\sigma(\langle  \vw_i^{(0)}, \vx' \rangle) \\
    & \qquad \quad+  \langle c_l \cdot \vx, \vx'\rangle (v_i^{(0)})^2 \dot\sigma(\langle \vw_i^{(0)}, c_l \cdot \vx\rangle)\dot\sigma(\langle \vw_i^{(0)},\vx' \rangle)\\
    & = \frac{1}{d} \sum_{l = 1}^{d}  h(c_l\cdot \vx, \vx')
\end{align*}
Equation \ref{eq:cntk_def} follows by noting that in the infinite $m$ limit, $h$ and $\bar h$ converge to $\ntk$ and $\cntk$ respectively. A rigorous proof is given in Propostition 4 from \citet{mei2021learning}.

\subsection{Proof of Lemma \ref{lem:cnn_ext}} \label{app:cntk_mercer}
Let $V_{d,k}$ be the space of degree-$k$ polynomials that are orthogonal to the space of polynomials of degree less than $k$, defined on $\sS^{d-1}$. Then $V_{d,k}(\gC_d)$ denotes the subspace of $V_{d,k}$ that is also $\gC_d$-invariant. 
\begin{proof}[Proof of Lemma \ref{lem:cnn_ext}]
The NTK kernel satisfies the Mercer condition and has the following Mercer decomposition \citep{bietti2020deep},
\begin{equation*}
    k_{\text{NTK}}(\vx, \vx') = \sum_{k=0}^\infty \mu_k \sum_{j=1}^{N(d,k)} Y_{j,k}(\vx)Y_{j,k}(\vx')
\end{equation*}
where $\{Y_{j,k}\}_{j \leq N(d,k)}$ form an orthonormal basis for $V_{d,k}$. \citet{bietti2020deep} further show that $Y_{j,k}$ is the $j$-th spherical harmonic polynomial of degree $k$, and $N(d,k) = \text{dim}(V_{d,k})$ gives  the total count of such polynomials, where
\[N(d, k) = \frac{2k+d-2}{k} \binom{k+d-3}{d-2}.\]
For any integer $k$, it also holds that \citep{bietti2020deep}
\begin{equation}\label{eq:gegenbauer}
\sum_{j=1}^{N(d,k)} Y_{j,k}(\vx)Y_{j,k}(\vx') = N(d,k) Q^{(d)}_k(\vx^T\vx')
\end{equation}
where $Q_k^{(d)}$ is the $k$-th Gegenbauer polynomial in dimension $d$. It follows from Equation \ref{eq:cntk_def} that $k_{\text{CNTK}}$ is also Mercer. Assume that it has a Mercer decomposition of the form
\[
k_{\text{CNTK}}(\vx, \vx') =  \sum_{k=0}^\infty \bar \mu_k \sum_{j=1}^{\bar N(d,k)} \bar Y_{j,k}(\vx)\bar Y_{j,k}(\vx').
\]
 We now proceed to identify $\bar Y_{j,k}$ and calculate $\bar N(d,k)$. From Equations \ref{eq:cntk_def} and \ref{eq:gegenbauer} we conclude that
\begin{equation*}
    \bar k(\vx, \vx') =\frac{1}{d}\sum_{l=1}^{d} k(\vx, c_l\cdot \vx') = \frac{1}{d}\sum_{l=1}^{d} \sum_{k=0}^\infty \mu_k N(d,k) Q^{(d)}_k (\langle\vx, c_l\cdot\vx'\rangle). 
\end{equation*}
Lemma 1 in \citet{mei2021learning} states that for any integer $k$, 
\begin{equation}\label{eq:mei_replemma}
     \frac{1}{d}N(d,k)\sum_{l=1}^{d} Q_k^{(d)}(\langle\vx, c_l\cdot\vx'\rangle) = \sum_{j=1}^{M(d,k)} Z_{j,k}(\vx)Z_{j,k}(\vx') 
\end{equation}

where $Z_{j,k}$ form an orthonormal basis for $V_{d,k}(\gC_d)$. Moreover, for the orthonormal basis $(Z_{j,k})_j$ over $\sS^{d-1}$,
\begin{equation}\label{eq:mei_replemma_2}
\sum_{j=1}^{M(d,k)} Z_{j,k}(\vx)Z_{j,k}(\vx) =  M(d,k).
\end{equation}

Therefore, $(\mu_k,Z_{j,k})_{j,k}$ is a sequence eigenvalue eigenfunction pairs for $\bar k$ and the following assignments satisfy the Mercer decomposition for $k_{\text{CNTK}}$:
\begin{align*}
    \bar \mu_k &= \mu_k,\\
    \bar Y_{j,k} &= Z_{j,k},\\
    \bar N(d,k) & = M(d,k).
\end{align*}
It remains to show that $M(d,k)\simeq N(d,k)/d$ when $C_d$ is the group of cyclic shifts acting on $\sS^{d-1}$. This follows from Equations \ref{eq:mei_replemma} and \ref{eq:mei_replemma_2},
\begin{align*}
  \frac{M(d,k)}{N(d,k)} & = \frac{1}{d} \sum_{l=1}^{d} Q_k^{(d)}(\langle\vx, c_l\cdot\vx \rangle)\\
  & = \Theta(d^{-1})
\end{align*}
where the last equation holds directly due to Lemma 4 in \citet{mei2021learning}.
\end{proof}

\section{Details of \ntkalg{} and \cntkalg} \label{app:ntk}
Section \ref{app:infogainbound} gives the proof to our statement on the information gain (Theorem~\ref{thm:info_gain}, Proposition~\ref{thm:info_gain_cnn}).
The regret bounds of \ntkalg under the GP (Theorem~\ref{thm:reg1}) and the RKHS assumptions (Theorem~\ref{thm:rkhsregret}) are proven in Section~\ref{app:gpregret} and Section~\ref{app:ntkrkhsregret}, respectively. In Section~\ref{app:supvar}, we present \supntkalg (Algorithm~\ref{alg:supcgp}), and discuss the exploration policy of this algorithm and its properties.

\subsection{Proof of Theorem \ref{thm:info_gain} and Proposition \ref{thm:info_gain_cnn}} \label{app:infogainbound}
We begin by giving an overview of the proof. Under the conditions of Theorem \ref{thm:info_gain}, the NTK is Mercer \citep{cao2019towards} and can be written as $k(\vx,\vx') = \sum_{i\geq 0}\lambda_i \phi_i(\vx)\phi_i(\vx')$, with $(\phi_i)_{i\geq 0}$ denoting the orthonoromal eigenfunctions. The main idea is to break $k$ into $k_p + k_o$ where $k_p(\vx,\vx') = \sum_{i\leq \tilde D} \lambda_i \phi_i(\vx) \phi_i(\vx') $ has a finite dimensional feature map $(\phi_i)_{i\leq \tilde D}$, corresponding to the sequence of $\tilde D$ largest eigenvalues of $k$. For any arbitrary sequence $X_T$, we are then able to decompose $I(\vy_T; \vf_T)$ in two terms, one corresponding to information gain of $k_p$, and the other, the tail sum of the eigenvalue series $\sum_{i\geq \tilde D}\lambda_i$. We proceed by bounding each term separately and picking $\tilde D$ such that the second term becomes negligible. 

\begin{proof}[Proof of Theorem \ref{thm:info_gain}] This proof adapts the finite-dimensional projection idea used in \citet{vakili2020information}.
\citet{cao2019towards} show that the NTK has a Mercer decomposition on the $d$-dimensional unit hyper-sphere. \citet{bietti2020deep} establish that
\begin{equation*}
    \ntk(\vx, \vx') = \sum_{k\geq 0}\mu_k \sum_{j=1}^{N(d,k)} Y_{j,k}(\vx)Y_{j,k}(\vx'),
\end{equation*}
where $Y_{j,k}$ is the $j$-th spherical harmonic polynomial of degree $k$, and for $k \geq 1$ there are \[N(d, k) = \frac{2k+d-2}{k} \binom{k+d-3}{d-2}\] of such $k$-degree polynomials. Using Stirling's approximation we can show that $N(d,k) = \Theta(k^{d-2})$. The functions $\{Y_{j,k}\}$ are an algebraic basis for $\mathcal{H}_\ntk$ the RKHS that is reproducing for $\ntk$. Consider a finite dimensional subspace of $\mathcal{H}_\ntk$ that is spanned by the eigenfunctions corresponding to the first $D$ \textit{distinct} eigenvalues of $\ntk$, $\vphi_D=\left((Y_{j,0})_{j\leq N(d,0)}, \cdots, (Y_{j, D})_{j\leq N(d,D)}\right)$. We decompose the NTK as $\ntk = k_P + k_O$, where $k_P$ is the kernel for the finite dimensional RKHS, and $k_O$ represents the kernel for the Hilbert space orthogonal to it. Let $\tilde D$ denote the length of $\vphi_D$, the feature map corresponding to $k_P$,
\begin{equation}\label{eq:tilde_D}
D \leq \tilde D = \sum_{k=0}^{D} N(d,k) \simeq C \sum_{k=0}^{D} k^{d-2} \leq C\frac{(D+1)^{d-1}}{d-1}.
\end{equation}
Note that the first $\tilde D$ eigenvalues of $\ntk$ make up its first $ D$ distinct eigenvalues. We write the information gain in terms of eigenvalues of $k_p$ and $k_O$, and find $D$ such that the finite-dimensional term dominates the infinite-dimensional tail. Assume the arbitrary sequence $X_T = (\vx_1, \cdots, \vx_T)$ is observed. The information gain is $I(\vy_T; \vf_T) = \frac{1}{2}\log\det(\mI + \sigma^{-2}K_{X_T})$, with $K_{X_T}$ being the kernel matrix, $(K_{X_T})_{i,j} = \ntk(\vx_i,\vx_j)$. Using a similar notation for the kernel matrices of $k_P$ and $k_O$, we may write
\begin{equation} \label{eq:infogain_decomp}
    \begin{split}
        I(\vy_T; \vf_T) & = \frac{1}{2}\log\det(\mI + \sigma^{-2}(K_{P,X_T}+K_{O,X_T}))\\
        & =  \frac{1}{2}\log\det(\mI + \sigma^{-2}K_{P,X_T}) + \frac{1}{2}\log\det(\mI + (\mI + \sigma^{-2}K_{P,X_T})^{-1}K_{O,X_T}) \\
    \end{split}
\end{equation}
We now separately bound the two terms. Let $\bm{\Phi}_{D,T} = [\vphi_D(\vx_1), \cdots, \vphi_D(\vx_T)]^T$, then by the mercer decomposition,
\[
K_{P, X_T} = \bm{\Phi}_{D,T} \Lambda_D \bm{\Phi}_{D,T}^T
\]
Where $\Lambda_D \in \sR^{\tilde D \times \tilde D}$ is a diagonal matrix with the first $\tilde D$ eigenvalues. Let $\mH_T = \Lambda^{1/2}_D \bm{\Phi}^T_{D,T} \bm{\Phi}_{D,T} \Lambda_D^{1/2}$, by Weinstein-Aronszajn identity,
\begin{equation*}
\begin{split}
    \frac{1}{2}\log\det(\mI_T + \sigma^{-2}K_{P,X_T}) & = \frac{1}{2}\log\det(\mI + \sigma^{-2}\mH_T)\\
    & \leq \frac{1}{2} \tilde D \log\left( \text{tr}(\mI + \sigma^{-2}\mH_T)/\tilde D\right)
\end{split}    
\end{equation*}    
For positive definite matrices $\mP \in \mathbb{R}^{n\times n}$, we have $\log\det \mP \leq n \log \text{tr} (\mP/n)$. The inequality follows from  $\mI_D + \sigma^{-2}\mH_T$ being positive definite. Plugging in the definition of $\mH_T$
\begin{equation*}
\begin{split}
    \frac{1}{2}\log\det(\mI_T + \sigma^{-2}K_{P,X_T}) & \leq \frac{1}{2} \tilde D \log\left(  1 +  \frac{\sigma^{-2}}{\tilde D}\text{tr}\left(\Lambda^{1/2}_D \bm{\Phi}^T_{D,T} \bm{\Phi}_{D,T} \Lambda_D^{1/2}\right) \right) \\
    & \leq \frac{1}{2} \tilde D \log\left(  1 +  \frac{\sigma^{-2}}{\tilde D}\sum_{t=1}^T\Lambda^{1/2}_D \Phi^T_D(\vx_t) \vphi_D(\vx_t)  \Lambda_D^{1/2} \right) \\
    & \leq \frac{1}{2} \tilde D \log\left(  1 +  \frac{\sigma^{-2}}{\tilde D}\sum_{t=1}^T \vert\vert\vphi_D({\vx_t})\Lambda^{1/2}\vert\vert_2^2 \right) \\ 
    &  \leq \frac{1}{2} \tilde D \log\left(  1 +  \frac{\sigma^{-2}}{D}\sum_{t=1}^T \sum_{k=0}^{D} \mu_k\sum_{j=1}^{N(d,k)}Y_{j,k}^2(\vx_t) \right) \\ 
    &  \leq \frac{1}{2} \tilde D \log\left(  1 +  \frac{\sigma^{-2}}{\tilde D}\sum_{t=1}^T k_P(\vx_t, \vx_t) \right) \\ 
    & \leq \frac{1}{2} \tilde D \log\left(  1 +  \frac{\sigma^{-2}T}{\tilde D}\right)
\end{split}    
\end{equation*}

The last inequality holds since the NTK is uniformly bounded by $1$ on the unit sphere. We now bound the second term in Equation \ref{eq:infogain_decomp}, which corresponds to the infinite-dimensional orthogonal space. Similar to the first term, we bound $\log\det \mP$ with $n \log\text{tr}(\mP/n)$,
\begin{equation} \label{eq:gamma_secondterm}
\begin{split}
        \frac{1}{2}\log\det(\mI_T + (\mI_T + \sigma^{-2}K_{P,X_T})^{-1}K_{O,X_T}) & \leq \frac{T}{2}\log\left( 1 + \frac{\text{tr}\left((\mI_T + \sigma^{-2}K_{P,X_T})^{-1}K_{O,X_T}\right)}{T}\right) \\
        &\leq
        \frac{T}{2}\log\left( 1 + \text{tr}\left(K_{O,X_T}\right)/T\right)
\end{split}
\end{equation}
The second inequality holds due to $(\mI_T + \sigma^{-2}K_{P,X_T})^{-1}$ being positive definite, with eigenvalues smaller than $1$. We can bound the trace using the Mercer decomposition
\begin{equation*}
    \begin{split}
        \text{tr}\left(K_{O,X_T}\right) & = \sum_{t=1}^T k_O(\vx_t, \vx_t) \\
        & = \sum_{t=1}^T \sum_{k = D+1}^\infty \mu_k \sum_{j=1}^{N(d, k)} Y_{j,k}^2(\vx_t)\\
        & = \sum_{t=1}^T \sum_{k = D+1}^\infty \mu_k N(d, k)Q^{(d)}_k(\vx_t^T\vx_t)
    \end{split}
\end{equation*}
where $Q^{(d)}_k$ is the Gegenbauer polynomial of degree $k$ and $\vx_t$ are on $\sS^{d-1}$. We have $Q^{(d)}_K(1) = 1$ which gives,
\begin{equation} \label{eq:tr_KO}
     \text{tr}\left(K_{O,X_T}\right) = T  \sum_{k = D+1}^\infty \mu_k N(d, k)
\end{equation}
Plugging in Equation \ref{eq:tr_KO} in Equation \ref{eq:gamma_secondterm}, the information gain can be bounded as,
\begin{equation*}
    I(\vy_T, \vf_T) \leq \frac{\tilde D}{2} \log\left(  1 +  \frac{T}{\sigma^{2}\tilde D}\right) + \frac{T}{2} \log\left(  1 + \sum_{k = D+1}^\infty \mu_k N(d, k) \right)
\end{equation*}
We now bound the second term using \citeauthor{bietti2020deep}'s result on decay rate of $\mu_k$. They show that there exists a constant $C_1(d,L)$ such that $\mu_k \leq C_1 k^{-d}$. Using Stirling approximation, there exists $C_2$ such that $N(d,k) \leq C_2 k^{d-2}$. Then,
\[
\sum_{k = D+1}^\infty \mu_k N(d, k) \leq C(d,L) \sum_{k=D+1}^\infty k^{-2}
\]
We can simply bound the series
\[
\sum_{k=D+1}^\infty k^{-2} \leq \int_D^\infty z^{-2}dz = \frac{1}{D}.
\]
Therefore, there exists there exists $\CNN$ such that,
\begin{equation} \label{eq:infogain_last}
\begin{split}
       I(\vy_T, \vf_T) & \leq \frac{\tilde D}{2} \log\left(  1 +  \frac{T}{\sigma^{2}\tilde D}\right) + \frac{T}{2} \log\left(  1 + \frac{\CNN}{D} \right) \\
       & \leq \frac{\tilde D}{2} \log\left(  1 +  \frac{T}{\sigma^{2}\tilde D}\right) + \frac{T\CNN}{2D}
\end{split}
\end{equation}
Note that the first term is increasing with $\tilde D$. We pick $\tilde D$ such that the first term in Equation \ref{eq:infogain_last} is dominant. Via Equation \ref{eq:tilde_D} we set, 
\begin{equation*}
    \tilde D = \Big\lceil \left( \frac{\CNN T}{\log (1+ \sigma^{-2}T)} \right)^{\frac{d-1}{d}} \Big\rceil
\end{equation*}
The treatment above holds for any arbitrary sequence $X_T$. Plugging in $\tilde D$ with Equation \ref{eq:infogain_last}, we then may write,
\[
\gamma_T \leq \left( \frac{\CNN T}{\log (1+ \sigma^{-2}T)} \right)^{\frac{d-1}{d}} \log \left( 1 + \sigma^{-2}T \left( \frac{\CNN T}{\log (1+ \sigma^{-2}T)} \right)^{\frac{d}{d-1}}\right)
\]
which concludes the proof.
\end{proof}
\begin{proof}[Proof of Proposition \ref{thm:info_gain_cnn}]
The CNTK is Mercer by Lemma \ref{lem:cnn_ext}. Which implies that we may repeat the steps taken in the proof of Theorem \ref{thm:info_gain}. To avoid confusion, we use the \emph{``bar''} notation to indicate the convolutional equivalent of the parameters from that proof. The 2-layer CNTK and the NTK share the same eigenvalues. The eigenfunctions of the CNTK also bounded by one over the hyper sphere. The only difference is that $\bar N(d,k) = N(d,k)/d$, which comes into effect for calculating $\text{tr}(\bar K_{O,X_T})$. For the CNTK we would have,
\[ \text{tr}\left(\bar K_{O,X_T}\right)= T  \sum_{k = \bar D+1}^\infty \mu_k \bar N(d, k) = T  \sum_{k = \bar D+1}^\infty \mu_k N(d, k)/d\]
Equivalent to Equation \ref{eq:infogain_last} we may write,
\[
       \bar I(\vy_T, \vf_T) \leq \frac{\tilde{ D}}{2} \log\left(  1 +  \frac{T}{\sigma^{2}\tilde{ D}}\right) + \frac{T\CNN}{2\bar Dd}
\]
where similar to the proof of Theorem \ref{thm:info_gain}, we have $\tilde D = \Theta(\bar D^{d-1})$. For the first term to be larger than the second, $\tilde D$ has to be set to
\[
\tilde D = \Big\lceil \left(\frac{\CNN T}{d\log (1+ \sigma^{-2}T)}\right)^{\frac{d-1}{d}}\Big\rceil
\]
which then concludes the proof. 
\end{proof}

\subsection{Proof of Theorem \ref{thm:reg1}} \label{app:gpregret}  
The proof closely follows the method in \citet{srinivas2009gaussian} and \citet{krause2011contextual}, with modifications on the assumptions on context domain and actions. The following lemmas will be used. 

\begin{lemma} \label{lem:subg1}
Let $\delta \in (0,1)$, and set $\beta_t = 2\log (|\mathcal{A}|\pi_t/\delta)$, where $\sum_{t>1} \pi_t^{-1} = 1$ and $\pi_t>0$. Then with probability of at least $1-\delta$
\begin{equation*}
    \vert f(\vz_t \va) - \mu_{t-1}(\vz_t \va) \vert \leq \sqrt{\beta_t} \sigma_{t-1}(\vz_t \va), \quad \forall \va \in \mathcal{A}, \, \forall t>1.
\end{equation*}
\end{lemma}
\begin{lemma} \label{lem:mutual_info}
Let $\sigma_{t-1}^2(\vx_t)$ be the posterior variance, computed at $\vx_t = \vz_t \va_t$, where $\va_t$ is the action picked by the UCB policy. Then,
\begin{equation*}
    \frac{1}{2} \sum_{t=1}^T \log( 1 + \sigma^{-2}\sigma_{t-1}^2(\vx_t)) \leq \gamma_T
\end{equation*}
\end{lemma}

\begin{proof}[Proof of Theorem \ref{thm:reg1}]
We use Lemma \ref{lem:subg1}, to bound the regret at step $t$. Let $\vx_t^* = \vz_t \va^*$ denote the optimal point, and $\vx_t = \vz_t \va_t$ be the maximizer of the UCB. Then by Lemma \ref{lem:subg1}, with probability of at least $1-\delta$ 
\begin{equation*}
    \begin{split}
        & \vert f(\vx_t) - \mu_{t-1}(\vx_t) \vert \leq \sqrt{\beta_t} \sigma_{t-1}(\vx_t)\\
        & \vert f(\vx_t^*) - \mu_{t-1}(\vx_t^*) \vert \leq \sqrt{\beta_t} \sigma_{t-1}(\vx_t^*)
    \end{split}
\end{equation*}
Therefore, by definition of $\vx_t$ (Equation \ref{eq:UCB_policy}) we can write:
\begin{equation*} 
\begin{split}
    r_t = f(\vx_t^*) - f(\vx_t) & \leq \mu_{t-1}(\vx_t^*) + \sqrt{\beta_t} \sigma_{t-1}(\vx_t^*) - f(\vx_t) \\
    & \leq \mu_{t-1}(\vx_t) + \sqrt{\beta_t} \sigma_{t-1}(\vx_t) - f(\vx_t) \\
    & \leq 2\sqrt{\beta_t} \sigma_{t-1}(\vx_t).
\end{split}
\end{equation*}
For the regret over $T$ steps, by Cauchy-Schwartz we have,
\begin{equation} \label{eq:R_t1}
\begin{split}
    R_T = \sqrt{\sum_{t=1}^T r_t} & \leq \sqrt{T} \sqrt{\sum_{t=1}^T r_t^2} \\
    & \leq \sqrt{T} \sqrt{\sum_{t=1}^T 4\beta_t \sigma_{t-1}^2(\vx_t)}. 
\end{split}
\end{equation}
Recall that Lemma \ref{lem:subg1}, holds for any  $\beta_t = 2\log (|\mathcal{A}|\pi_t/\delta)$, where $\sum_{t>1} \pi_t^{-1} = 1$. We pick $\pi_t = \frac{\pi^2t^2}{6}$, so that $\beta_t$ is non-increasing and can be upper bounded by $\beta_T$. This allows to reduce the problem of bounding regret to the bounding the sum of posterior variances. By Equation \ref{eq:GPposteriors}, and since $\ntk(\vx,\vx)\leq 1$,
\[
\sigma^{-2}\sigma_{t-1}^2(\vx_t) \leq \sigma^{-2}\ntk(\vx_t, \vx_t) \leq \sigma^{-2}
\]
For any $r \in [0,a]$, it holds that $r \leq \frac{a \log (1+ r)}{ \log(1+a)}$. Therefore,
\[
\sigma^{-2}\sigma_{t-1}^2 \leq \frac{\sigma^{-2}}{\log (1+\sigma^{-2})} \log (1 + \sigma^{-2}\sigma_{t-1}^2)
\]
Putting together the sum in Equation \ref{eq:R_t1} we get,
\begin{equation*}
\begin{split}
    R_t & \leq \sqrt{4T\beta_T\sigma^2 {\sum_{t=1}^T}\sigma^{-2}\sigma_{t-1}^2 }\\
    & \leq \sqrt{\frac{4T\beta_T\sigma^2}{\log (1+\sigma^{-2})} \sum_{t=1}^T\log (1 + \sigma^{-2}\sigma_{t-1}^2)}\\
    & \leq \sqrt{\frac{8\sigma^2}{\log (1+\sigma^{-2})}} \sqrt{T\beta_T\gamma_T}
\end{split}
\end{equation*}
where the last inequality holds by plugging in Lemma \ref{lem:mutual_info}.
\end{proof}
%------------------------------------------------------------------------------------
%------------------------------------------------------------------------------------
%------------------------------------------------------------------------------------
%------------------------------------------------------------------------------------

\subsubsection{Proof of Lemma \ref{lem:subg1}}
\begin{proof}
Fix $t\geq 1$. Conditioned on $\vy_{t-1} = (y_1, \cdots, y_{t-1})$, $\vx_1, \cdots, \vx_{t-1}$ are deterministic. The posterior distribution is $f(\vx) \sim N (\mu_{t-1}(\vx), \sigma^2_{t-1}(\vx))$. Applying the sub-Gaussian inequality, and conditioned on the history, 
\[
\text{Pr}[\vert f(\vx) - \mu_{t-1}(\vx) \vert > \sqrt{\beta_t} \sigma_{t-1}(\vx)] \leq e^{-\beta_t/2}
\]
$\mathcal{A}$ is finite and $\vx = \vz_t \va$, then by union bound over $S$, the following holds with probability of at least $1-|\mathcal{A}|e^{-\beta_t/2}$,
\[
\vert f(\vz_t \va) - \mu_{t-1}(\vz_t \va) \vert \leq \sqrt{\beta_t} \sigma_{t-1}(\vz_t \va), \quad \forall \va \in \mathcal{A}
\]
It is only left to further apply a union bound over all $t\geq 1$. For the statement in lemma to hold, $\beta_t$ has to be set such that, $\sum_{t\geq1}\vert\mathcal{A}\vert e^{\beta_t/2} \leq \delta$. Setting $\beta_t = 2\log (|\mathcal{A}|\pi_t/\delta)$ with $\sum_{t>1}\pi_t =1$ satisfies the condition.
\end{proof}
%------------------------------------------------------------------------------------
%------------------------------------------------------------------------------------
\subsubsection{Proof of Lemma \ref{lem:mutual_info}}
\begin{proof} Recall that for a Gaussian random variable entropy is, $H(N(\bm{\mu, \Sigma})) = \frac{1}{2}\log\det(2\pi e\bm{\Sigma)}$. Let $\vy_T$ be the vector of observed rewards and $\vf_T = [f(\vx_t)]_{t\leq T} \in \mathbb{R}^T$ the true rewards. We have $\vy_T = \vf_T + \bm{\epsilon}_T$, therefore, $\vy_T\vert \vf_T \sim N(0,\bm{I}\sigma^2)$, and $y_{T}\vert \vy_{T-1} \sim N(\mu_{T-1},\sigma^2+\sigma_{T-1}^2(\vx_T)))$. By the definition of mutual information,
\begin{equation*}
\begin{split}
    I(\vy_T; \vf_T) & = H(\vy_T) - H(\vy_T\vert \vf_T)\\
    & = H(\vy_{T-1}) + H(y_{T}\vert \vy_{T-1})- \frac{T}{2}\log (2\pi e \sigma^2)\\
    & =  H(\vy_{T-1}) + \frac{1}{2}\log (2\pi e (\sigma^2+\sigma_{T-1}^2(\vx_T)))- \frac{T}{2}\log (2\pi e \sigma^2).\\
\end{split}
\end{equation*}
The first equality holds by the chain rule for entropy. By recursion, 
\[
\gamma_T\geq I(\vy_T; \vf_T) = \frac{1}{2} \sum_{t=1}^T \log( 1 + \sigma^{-2}\sigma_{t-1}^2(\vx_t))
\]
\end{proof}

\subsection{The Sup Variant and its Properties} \label{app:supvar}
\begin{algorithm}[ht]
\DontPrintSemicolon
\KwInput{$\Psi_t \subset \{ 1, \cdots, t-1\}$, $\vz_t$}
\init{
    $\bm{K} \leftarrow [\ntk(\vx_i,\vx_j)]_{i,j \in \Psi_t}$, $\mZ^{-1} \leftarrow (\bm{K}+ \sigma^2\bm{I})^{-1}$, $\bm{y} \leftarrow [y_i]^T_{i \in \Psi_t}$}{}
 \For{$\va \in \mathcal{A}$}
{
Define $\vx := \vz_t\va$,\\
$\bm{k} \leftarrow [\ntk(\vx_i, \vx)]^T_{i \in \Psi_t}$\\
$\mu_{t-1}^{(s)}(\vx) \leftarrow \bm{k}^T \mZ^{-1}\bm{y}$\\
$\sigma_{t-1}^{(s)}(\vx) \leftarrow \sqrt{
\ntk(\vx,\vx) - \bm{k}^T\mZ^{-1}\bm{k}
}$
}
\caption{ \label{alg:getpost}GetPosterior}
\end{algorithm}
\begin{algorithm}[ht]
\DontPrintSemicolon
\init{ 
    $\Psi^{(s)}_1 \leftarrow \emptyset \, \forall s \leq S$
    }{}
 \For{$t=1$ to $T$}
 {
 $m\leftarrow 1$, $A_1 \leftarrow \mathcal{A} $\\
 \While{action $\va_t$ is not chosen}
 {
 $\Big(\mu_{t-1}^{(s)}(\vz_t\va),\sigma_{t-1}^{(s)}(\vz_t\va)\Big) \leftarrow$ GetPosteriors$\big(\Psi_t^{(s)}, \vz_t\big)$ for all $\va \in A_s$\\
 \If{$\forall \va \in A_s$, $\sqrt{\beta_t} \sigma_{t-1}^{(s)}(\vz_t\va) \leq \frac{\sigma}{\sqrt{T}}$}
        {
        Choose $\va_t = \arg\max_{\va\in A_s}  \mu_{t-1}^{(s)}(\vz_t\va)+\sqrt{\beta_t}\sigma_{t-1}^{(s)}(\vz_t\va) $.\\
        Keep the index sets $\Psi_{t+1}^{(s')} = \Psi_{t}^{(s')}$ for all $s'\leq S$
        }
\ElseIf{$\forall \va \in A_s$, $\sqrt{\beta_t} \sigma_{t-1}^{(s)}(\vz_t\va) \leq \sigma2^{-s}$}{
$A_{s+1} \leftarrow \Big\{ \va \in A_s \Big\vert  \mu_{t-1}^{(s)}(\vz_t\va)+\sqrt{\beta_t}\sigma_{t-1}^{(s)}(\vz_t\va) \geq \max_{\va\in A_s}  \mu_{t-1}^{(s)}(\vz_t\va)+\sqrt{\beta_t}\sigma_{t-1}^{(s)}(\vz_t\va)  - \sigma 2^{1-s} \Big\}$\\
$s \leftarrow s + 1$}
\Else{
Choose $\va_t \in A_s$ such that $\sqrt{\beta_t} \sigma_{t-1}^{(s)}(\vz_t\va_t) > \sigma2^{-s}$.\\
Update the index sets at all levels $s'\leq S$,\\
$
\Psi_{t+1}^{(s')} = \begin{cases}
      \Psi_{t+1}^{(s)}\cup \{t\} & \text{if $s'=s$}\\
      \Psi_{t+1}^{(s)} & \text{otherwise}\\
    \end{cases} 
$}
 }
 }
\caption{\label{alg:supcgp} \supntkalg{} Algorithm\\
$T$ number of total steps, $S = \log T$ number of discretization levels}
\end{algorithm}
\supntkalg combines \ntkalg{} policy and Random Exploration, and at every step $t$, only uses a subset of the previously observed context-reward pairs. These subsets are constructed such that the rewards in each are statistically independent, conditioned on the contexts. Informally put, then the learner chooses an action either if its posterior variance is very high or if the reward is close to the optimal reward. As more steps are played, the criteria for \emph{closeness} to optimal reward and \emph{high variance} is refined. The method is given in Algorithm \ref{alg:supcgp}. We give some intuition on the key elements to which the algorithm's desirable properties can be credited.
\begin{itemize}
    \item  The set of indices of the context-reward pairs used for calculating $\mu_{t-1}^{(s)}$ and $\sigma_{t-1}^{(s)}$, is denoted by $\Psi_t^{(s)}$.
    Once an action is chosen, $\Psi_t^{(s)}$ is updated to $\Psi_{t+1}^{(s)}$ for all $s$. Each set either grows by one or remains the same.
    \item For every level $s$, the set $A_s$ includes $\va_t^*$ the true maximizer of the reward with high probability. At every step $t$ we start with $A_1$ which includes all the actions, and start removing actions which have a small UCB and are unlikely to be $\va_t^*$ . 
    \item The UCB strategy is only used if the learner is certain about the outcome of all actions within $A_s$, i.e. $\sigma_{t-1}^{(s)}(\vz_t\va) \leq \sigma/\sqrt{T\beta_t}$, for all $\va \in A_s$. The context-reward pairs of these UCB steps are not saved for future estimation of posteriors, i.e. $\Psi_{t+1}^{(s)} = \Psi_t^{(s)}$.
    \item At step $t$, if there are actions $\va \in A_s$ for which $\sqrt{\beta_t}\sigma_{t-1}^{(s)}(\vz_t\va) > \sigma2^{-s}$, then one is chosen at random, and the set  $\Psi_t^{(s)}$ is updated with the index $t$, while all other sets remain the same.
    \item \looseness -1 The last case of the \emph{if} statement in Algorithm \ref{alg:supcgp} considers a middle ground, when the learner is not certain \textit{enough} to pick an action by maximizing the UCB, but for all $\va \in A_s$ posterior variance is smaller than $\sigma2^{-s}/\sqrt{\beta_t}$. In this case, the level s is updated as $s\leftarrow s+1$. In doing so, the learner considers a finer uncertainty level, and updates its criterion for \emph{closeness} to the optimal action.
    \item The parameter $s$ discretizes the levels of uncertainty. For instance, in the construction of $\Psi_t^{(s)}$, the observed context-reward pairs at steps $t$ are essentially partitioned based on which $[2^{-(s+1)}, 2^{-s}]$ interval $\sigma_{t-1}^{(s)}$ belongs to. If for all $s \leq S$, $\sqrt{\beta_t}\sigma_{t-1}^{(s)} \leq \sigma 2^{-s}$, then that pair is disregarded. Otherwise, it is added to $\Psi_{t+1}^{(s)}$ with the smallest $s$, for which $\sqrt{\beta_t}\sigma_{t-1}^{(s)} \leq \sigma 2^{-s}$. We set $S = \log T$, ensuring that $\frac{\sigma}{\sqrt{T}} \leq \sigma 2^{-S}$. 
    \end{itemize}
\paragraph{Properties of the \supntkalg} The construction of this algorithm guarantees properties that will later facilitate the proof of a $\title{\gO}(\sqrt{T\gamma_T})$ regret bound. These properties are given formally in Lemma \ref{lem:supcgp_properties} and Proposition \ref{prop:supcgp_props}, here we give an overview. \supntkalg satisfies that for every $t \leq T$ and $s \leq S$:
\begin{enumerate}
    \item The true maximizer of reward remains within the set of plausible actions, i.e.,  $\va^*_t \in A_s$.
    \item \label{item:prop2sup} Given the context $\vz_t$, regret of the actions $\va \in A_s$, is bounded by $2^{3-s}$, 
\end{enumerate}
 with high probability over the observation noise.  Let $\mX^{(s)}$ denote sequence of $\vx_\tau$ with $\tau \in \Psi_t^{(s)}$. Conveniently, the construction of $\Psi_t^{(s)}$ guarantees that,
 \begin{enumerate}
   \setcounter{enumi}{2}
     \item \label{item:prop4sup}  \looseness -1 Given $\mX^{(s)}$, the corresponding rewards $\vy^{(s)}$ are independent random variables and $ \mathbb{E} y_\tau = f(\vx_\tau)$.
          \item \label{item:prop3sup} Cardinality of each uncertainty set $\vert \Psi^{(s)}_t\vert$, is bounded by $\mathcal{O}(\gamma_T\log T)$.
 \end{enumerate}
\subsection{Proof of Theorem \ref{thm:rkhsregret}} \label{app:ntkrkhsregret}
Our proof adapts the technique in \citet{valko2013finite}. Consider the average cumulative regret given the inputs, by property \ref{item:prop4sup} of the algorithm we may write it as,
\[
\mathbb{E}[R_T \vert X_T] = \sum_{t \in \bar\Psi} f(\vx_t^*)-f(\vx_t) + \sum_{t \in [T]/\bar\Psi} f(\vx_t^*)-f(\vx_t) 
\]
where $\bar{\Psi} := \{ t\leq T |\, \forall s, \, t \notin \Psi_{T}^{(s)} \}$ includes the indices of steps with small posterior variance, i.e. $\sigma_{t}(\vx)\leq \sigma/\sqrt{T\beta_t}$. For bounding the first term, we use Azuma-Hoeffding to control $f(\vx_t^*)-f(\vx_t) $, with $\sigma_t(\vx_t)C(B, \sqrt{\gamma_T}, \beta_T)$. Since $\sigma_{t-1}$ is small for $t \in \bar \Psi$, this term grows slower than $\mathcal{O}(\sqrt{\gamma_TT})$. We then use properties \ref{item:prop2sup} and \ref{item:prop3sup}, to bound the second term. Having bounded $\mathbb{E}[R_T \vert X_T]$, we again use Azuma-Hoeffding, to give a bound on the regret $R_T$. 

For this proof, we use both feature map and kernel function notation. Let $\mathcal{H}_k$ be the RKHS corresponding to $k$ and the sequence $\bm{\phi} = (\sqrt{\lambda_i}\phi_i)_{i=1}^\infty$, be an algebraic orthogonal basis for $\mathcal{H}_k$, such that $k(\vx, \vx') = \bm{\phi}^T(\vx) \bm{\phi}(\vx') $. For $f \in \mathcal{H}_k$ and there exists a unique sequence $\bm{\theta}$, such that $f = \bm{\phi}^T\bm{\theta}$. If $\norm{f}_k \leq B$ then $\vert \vert \bm{\theta}\vert \vert \leq B$, since
\[
\vert \vert f \vert \vert^2_k = \sum_i \frac{(\langle f, \phi_i \rangle)^2}{\lambda_i} = \sum_i \frac{(\langle \sum_j \sqrt{\lambda_j \theta_j \phi_j}, \phi_i \rangle)^2}{\lambda_i} = \sum_i \theta_i^2 = \vert \vert \bm{\theta} \vert \vert^2
\]
The following lemmas will be used to prove the theorem. 
\begin{lemma}\label{lem:valko1}
Fix $s\leq S$, for any action $\va \in \mathcal{A}$, let $\vx = \vz_t\va$. Then with probability of at least $1-2|\mathcal{A}| e^{-\beta_T/2}$,
\[
\vert \mu^{(s)}_t(\vx) - f(\vx) \vert \leq \sigma^{(s)}_t(\vx) \big[2\sqrt{\beta_t}+ \sigma  \big]B.
\]
\end{lemma}
\begin{lemma} \label{lem:supcgp_properties}
For any $t \leq T$, and $s \leq S$, with probability greater than $1-2|\mathcal{A}| \exp^{-\beta_T/2}$, 
    \begin{enumerate}
     \item $\va^*_t \in A_s$.
         \item For all $\va \in A_s$, given $\vz_t$,
    $
    f(\vx_t^*) - f(\vx) \leq 2^{3-s}
    $
    \end{enumerate}
\end{lemma}
\begin{lemma} \label{lem:cardofpsi}
For any $s\leq S$, the cardinality of $\Psi_{T}^{(s)}$ is grows with $T$ as follows,
    \[
    \vert \Psi_{T}^{(s)} \vert \leq \mathcal{O}\Big( 4^s \gamma_T \log T\Big)  
    \]
\end{lemma}
\begin{lemma} \label{lem:bounded_f}
Consider $\ntk$ defined on $\mathcal{X} \subset \mathbb{S}^{d-1}$, and its corresponding RKHS, $\gH_\ntk$. Any $f \in \mathcal{H}_\ntk$ where $\norm{f}_\ntk \leq B$, is uniformly bounded by $B$ over $\mathcal{X}$.
\end{lemma}
\begin{lemma}[Azuma-Hoeffding Inequality] \label{lem:Azuma}
Let $X_1, \cdots, X_T$ be random variables with $X_t \leq a_t$ for some $a_t > 0$. Then with probability greater than $1- 2\exp\Big(\frac{-B^2}{2\sum_t a_t^2}\Big)$,
\[
\big\vert \sum_t X_t - \sum_t \mathbb{E} [X_t\vert X_1, \cdots X_{t-1}] \big\vert \leq B
\]
\end{lemma}
\begin{proof}[Proof of Theorem \ref{thm:rkhsregret}]
Denote by $H_{t-1}$ the history of the algorithm at time $t$,
\[
H_{t-1} := \Big\{(\vz_i,\va_i, y_i)\Big\}_{i<t} \cup {\vz_t}
\]
Define $X_t = f(\vx_t^*) - f(\vx_t)$. By Lemma \ref{lem:bounded_f}, $f$ is bounded and we have $|X_t|\leq B$. Then by applying Azuma-Hoeffding (Lemma \ref{lem:Azuma}) on the random variables $X_1, \cdots, X_T$, with probability greater than $1-2T|\mathcal{A}|e^{-\beta_T/2}$,
\begin{equation}
    \label{eq:azumaonregret}
    \big \vert R(T) - \mathbb{E} \big[R(T) | H_{T-1}\big] \big \vert \leq \sqrt{2TB^2 \log (\frac{1}{T|\mathcal{A}|e^{-\beta_T/2}})}
\end{equation}
We now use lemmas \ref{lem:valko1}, \ref{lem:cardofpsi}, and \ref{lem:supcgp_properties} to bound the growth rate of $ \mathbb{E} \big[ R(T) | H_{t-1}\big]$. Recall $\bar{\Psi} := \{ t\leq T |\, \forall s, \, t \notin \Psi_{T}^{(s)} \}$.
\begin{equation} \label{eq:decompreg}
    R(T) = \sum_{t \in [T]/\bar{\Psi}} f(\vx_t^*) - f(\vx_t) + \sum_{t \in \bar{\Psi}} f(\vx_t^*) - f(\vx_t)
\end{equation}
Bounding the expectation of the first sum gives,
\begin{equation} \label{eq:term1regrkhs}
\begin{split}
     \sum_{t \in [T]/\bar{\Psi}} \mathbb{E}\big[ f(\vx_t^*) - f(\vx_t) | H_{t-1} \big] & = \sum_{s\leq S} \sum_{t \in \Psi_T^{(s)}} f(\vx_t^*) - f(\vx_t)\\
    & \leq \sum_{s\leq S} 2^{3-s}|\Psi_T^{(s)}|\\
    & \leq 8 S \sqrt{\beta_T  (10+\sigma^{-2}15)\gamma_T T\log T}
\end{split}
\end{equation}
\looseness -1 with a probability of at least $1-2ST|\mathcal{A}|e^{-\beta_T/2}$. The First equation holds by Prop. \ref{prop:supcgp_props}, and the second holds due to Lemma \ref{lem:supcgp_properties}. For the third, we have used the inequality in Lemma \ref{lem:valko5} and that $|\Psi_T^{(s)}|\leq T$. We now bound expectation of the second term in Equation \ref{eq:decompreg}. By Lemma \ref{lem:valko1}, with probability of greater than $1-2T|\mathcal{A}|e^{-\beta_T/2}$,
\begin{equation}
    \label{eq:term2regrkhs}
    \begin{split}
        \sum_{t \in \bar{\Psi}} \mathbb{E} \big[f(\vx_t^*) - f(\vx_t)  | H_{t-1} \big] & = \sum_{t \in \bar{\Psi}} f(\vx_t^*) - f(\vx_t)\\
        & \stackrel{\text{(a)}}{\leq} \sum_{t \in \bar{\Psi}} \mu_t(\vx_t^*) + \sqrt{\beta_t}\sigma_t(\vx_t^*) + B(2\sqrt{\beta_t}+ \sigma)\sigma_t(\vx_t^*)- f(\vx_t) \\
        & \stackrel{\text{(b)}}{\leq}\sum_{t \in \bar{\Psi}} \mu_t(\vx_t) + \sqrt{\beta_t}\sigma_t(\vx_t) + B(2\sqrt{\beta_t}+ \sigma)\sigma_t(\vx_t^*)- f(\vx_t)\\
        & \stackrel{\text{(c)}}{\leq} \sum_{t \in \bar{\Psi}} \sqrt{\beta_t}\sigma_t(\vx_t) + B(2\sqrt{\beta_t}+ \sigma)(\sigma_t(\vx_t^*)+\sigma_t(\vx_t))\\
        & \stackrel{\text{(d)}}{\leq} \sum_{t \in \bar{\Psi}} \frac{\sigma}{\sqrt{T}}+ 2B(2+\frac{\sigma}{\sqrt{\beta_T}})\frac{\sigma}{\sqrt{T}}\\
        & \leq  \sigma \sqrt{T}\big(1 + 2B(2+\frac{\sigma}{\sqrt{\beta_T}})\big)
    \end{split}
\end{equation}
For inequalities (a) and (c), we have used Lemma \ref{lem:valko1}, for $\vx_t^*$ and $\vx_t$ respectively. By Algorithm \ref{alg:supcgp}, if $t \in \bar{\Psi}$, then $\vx_t$ is the maximizer of the upper confidence bound, resulting in inequality (b). Lastly for inequality (d), by construction of $\bar{\Psi}$, we have that $\sqrt{\beta_t} \sigma_{t-1}^{(s)}(\vx) \leq \frac{\sigma}{\sqrt{T}}$. We plug in $S = \log T$, $\beta_t = \beta_T = 2 \log (2T|\mathcal{A}|/\delta) $ and substitute $\delta$ with $\delta/(1+\log T)$. Putting together equations \ref{eq:azumaonregret}, \ref{eq:term1regrkhs}, and \ref{eq:term2regrkhs} gives the result.
\end{proof}

%------------------------------------------------------------------------------------
%------------------------------------------------------------------------------------
%------------------------------------------------------------------------------------
\subsubsection{Proof of Lemma \ref{lem:valko1}}
\looseness -1 \supntkalg is constructed such that the proposition below holds following the results in \citet{valko2013finite}. 
\begin{prop}[Lemma 6 \citet{valko2013finite}]\label{prop:supcgp_props}
Consider the \supntkalg{} algorithm. For all $t \leq T$, $s \leq S$, and for a fixed sequence of $\vx_t$ where $t \in \Psi_t^{(s)}$. The corresponding rewards $y_t$ are independent random variables, and we have $\mathbb{E} [y_t | \vx_t ] = f(\vx_t)$. 
\end{prop}
\begin{proof}[Proof of Lemma \ref{lem:valko1}]
Let $\bm{k},\,\bm{K}$, and $\vy$ be defined as they are in Algorithm \ref{alg:getpost}, then by definition of $\bm{\phi}$,
\begin{equation} \label{eq:mu_rkhs}
\begin{split}
    \mu^{(s)}_t(\vx) & = \bm{k}^T(\bm{K}+\sigma^2\bm{I})^{-1}\vy \\
    & = \bm{\phi}^T(\vx) (\bm{K}+\sigma^2\bm{I})^{-1}\bm{\Phi}^T \vy \\
    & = \bm{\phi}^T(\vx) (\bm{\Phi}^T\bm{\Phi}+\sigma^2\bm{I})^{-1}\bm{\Phi}^T\vy
 \end{split}
\end{equation}
where $\bm{\Phi} = [\bm{\phi}^T(\vx_i)]^T_{i \in \Psi^{(s)}_t}$. For simplicity let $C = \bm{\Phi}^T\bm{\Phi}+\sigma^2\bm{I}$. Similarly for $\sigma^{(s)}_t(\vx)$ we can write:
\begin{equation}
    \label{eq:sigma_rkhs}
    \begin{split}
         \sigma ^{(s)}_t(\vx) & = \sqrt{\bm{\phi}^T(\vx) C^{-1}  \bm{\phi}(\vx)} \\
         & =\sqrt{ \bm{\phi}^T(\vx) C^{-1} (\bm{\Phi}^T\bm{\Phi}+\sigma^2\bm{I}) C^{-1} \bm{\phi}(\vx)} \\
         & \geq \vert \vert \bm{\Phi} C^{-1} \bm{\phi}(\vx) \vert \vert
    \end{split}
\end{equation}
Using Equation \ref{eq:mu_rkhs}, we get
\begin{equation} \label{eq:valko1_main}
    \begin{split}
       \mu^{(s)}_t(\vx) - f(\vx) & =  \bm{\phi}^T(\vx) (\bm{\Phi}^T\bm{\Phi}+\sigma^2\bm{I})^{-1}\bm{\Phi}^T\vy - \bm{\phi}^T\bm{\theta} \\
       & = \bm{\phi}^T(\vx) C^{-1}\bm{\Phi}^T \vy - \bm{\phi}^T C^{-1} C\bm{\theta} \\
       & = \bm{\phi}^T(\vx) C^{-1} \bm{\Phi}^T(\vy - \bm{\Phi}\bm{\theta}) - \sigma^2 \bm{\phi}^T(\vx)C^{-1}\bm{\theta}
    \end{split}
\end{equation}
We now bound the first term in Equation \ref{eq:valko1_main}. By Proposition \ref{prop:supcgp_props}, conditioned on $\vx$ and $[\vx_i]_{i \in \Psi_t}$, $\vy - \bm{\Phi}\bm{\theta}$ is a vector of zero-mean independent random variables. By Lemma \ref{lem:bounded_f}, $f$ is bounded. Similar to \citet{valko2013finite}, and without loss of generality, we may normalize the vector $\vy$ over $\Psi_t^{(s)}$, and assume that $y_i\leq B$. Then, each $\vert y_i - \bm{\phi}^T(\vx_i)\bm{\theta}\vert  \leq 2B$. By Equation \ref{eq:sigma_rkhs}, and Lemma \ref{lem:Azuma}, with probability greater than $1-2\exp(-\beta_T/2)$,
\begin{equation} \label{eq:valkoazuma1}
    \vert \bm{\phi}^T(\vx) (\bm{\Phi}^T\bm{\Phi}+\sigma^2\bm{I})^{-1} \bm{\Phi}^T(\vy - \bm{\Phi}\bm{\theta}) \vert \leq 2B\sqrt{\beta_t}\sigma ^{(s)}_t(\vx)
\end{equation}
For the second term, by Cauchy-Schwartz we have,
\begin{equation} \label{eq:valkosigma}
\begin{split}
   \sigma^2 \bm{\phi}^T(\vx)(\bm{\Phi}^T\bm{\Phi}+\sigma^2\bm{I})^{-1}\bm{\theta} & = \sigma^2 \bm{\phi}^T(\vx)C^{-1}\bm{\theta}\\
   & \leq B\sigma^2  \sqrt{\bm{\phi}^T(\vx)C^{-1} \sigma^{-2} \sigma^2 \bm{I}  C^{-1}\bm{\phi}(\vx)} \\
   & \leq B \sigma   \sqrt{\bm{\phi}^T(\vx)C^{-1}C  C^{-1}\bm{\phi}(\vx)} \\
   & \leq B \sigma \sigma^{(s)}_t(\vx)
\end{split}
\end{equation}
The proof is concluded by plugging in Equations \ref{eq:valkoazuma1} and \ref{eq:valkosigma} into Equation \ref{eq:valko1_main}, and taking a union bound over all $|\mathcal{A}|$ actions.
\end{proof}
%------------------------------------------------------------------------------------
%------------------------------------------------------------------------------------

\subsubsection{Proof of Other Lemmas}
\begin{proof}[\textbf{Proof of lemma \ref{lem:supcgp_properties}}]
We prove this lemma by showing the equivalent of \ntkalg and Kernelized UCB and refer to Lemma 7 in \citet{valko2013finite}. Consider the GP regression problem with gaussian observation noise, i.e., $y_i = f(\vx_i) + \varepsilon_i$, with $\varepsilon \sim \mathcal{N}(0, \sigma^2)$, and $f \sim \text{GP}(0, k)$. Let $\bm{\phi}(\cdot)$ be the feature map of the GP's covariance kernel function $k$, i.e. $k(\vx, \vx') = \bm{\phi}^T(\vx) \bm{\phi}(\vx')$. Denote $\mu^{(\text{GP})}_t(\cdot)$ as the posterior mean function, after observing $t$ samples of $(\vx_i, y_i)$ pairs. Then, it is straightforward to show that, for all $\vx \in \mathcal{X}$, 
\[
\mu^{(\text{GP})}_t(\vx) = \mu^{(\text{Ridge})}_t(\vx)
\]
Where $\mu^{(\text{Ridge})}_t(\vx) = \bm{\phi}^T(\vx)\bm{\theta}^*$, with $\bm{\theta}^*$ being the minimizer of the kernel ridge loss,
\[
\mathcal{L}(\vx) = \sum_{i=1}^t (y_i - \bm{\phi}^T(\vx_i)\bm{\theta})^2 + \sigma^2 \vert\vert \bm{\theta} \vert\vert^2.
\]
Similarly, let $\sigma^{(\text{GP})}_t(\vx)$ be the posterior variance function of the GP regression. Using matrix identities we have,
\[
\sigma^{(\text{GP})}_t(\vx) = \sqrt{\bm{\phi}^T(\vx)(K_T + \sigma^2\bm{I})^{-1}\bm{\phi}(\vx)} = \sigma^{(\text{Ridge})}_t(\vx)
\]
which is the width of confidence interval used in \textsc{Kernelized UCB} \citep{valko2013finite}. Their exploration policy is then defined as,
\[
x^{(\text{Ridge})}_t = \arg\max_{x \in \mathcal{X}} \mu^{(\text{Ridge})}_t(\vx) + \frac{\beta}{\sigma} \sigma^{(\text{GP})}_t(\vx) .
\]
We conclude that \ntkalg{} and \textsc{Kernelized UCB} \citep{valko2013finite} are equivalent, up to a constant factor in exploration coeficient $\beta$. We modify \textsc{SupKernelUCB} to \supntkalg{} such that the key lemmas still hold, and Lemma \ref{lem:supcgp_properties} immediately follows from \citet{valko2013finite}'s Lemma 7.
\end{proof}

\begin{proof}[\textbf{Proof of Lemma \ref{lem:cardofpsi}}]
Let $\lambda_i$ denote the eigenvalues of $\bm{K}_T+\sigma^2\bm{I}$, in decreasing order. \citeauthor{valko2013finite} define 
\[
\tilde{d} := min \big \{ j: \, j\sigma^2\log T\geq \sum_{i\geq j} \lambda_i - \sigma^2 \big\},
\]
and show that $\gamma_T$ gives a data independent upper bound on $\tilde{d}$,
\begin{equation*}
    \gamma_T\geq I(\vy_T; \vf) \geq \Omega (\tilde{d}\log\log T)
\end{equation*}
Due to the equivalence of \textsc{SupKernelUCB} and \supntkalg, as shown in the proof of lemma \ref{lem:supcgp_properties}, the following lemma holds for \supntkalg.
\begin{lemma}[Lemma 5, \citet{valko2013finite}] \label{lem:valko5}
Let $l_T = \max\{\log T, \log(T/(\sigma\tilde{d}))\}$. For all $s \leq S$,
\[
|\Psi_T^{(s)}| \leq 2^s \sqrt{\beta_T (10+15\sigma^{-2}) \tilde{d}|\Psi_T^{(s)}| l_T}
\]
\end{lemma}
By Lemma \ref{lem:valko5}, there exists $T_0$ such that for all $T>T_0$, 
\[
|\Psi_T^{(s)}| \leq 2^s \sqrt{\beta_T (10+15\sigma^{-2}) \gamma_T |\Psi_T^{(s)}|\log T}.
\]
\end{proof}

\begin{proof}[\textbf{Proof of Lemma \ref{lem:bounded_f}}]
By the Reproducing property of $\ntk$, for any $f \in \mathcal{H}_\ntk$ we have,
\begin{equation*}
    f(\vx) = \langle f, \ntk(\vx, \cdot)\rangle_{\mathcal{H}_\ntk} \leq \vert\vert f\vert\vert_\ntk \vert\vert \ntk(\vx, \cdot)\vert \vert_\ntk
\end{equation*}
where the inequality holds due to Cauchy-Schwartz. The NTK is Mercer over $\mathcal{X}$, with the mercer decomposition $\ntk = \sum_i \lambda_i \phi_i$. Then by definition of inner product in $\mathcal{H}_\ntk$, 
\begin{equation*}
\begin{split}
    \vert\vert \ntk(\vx, \cdot)\vert \vert_k^2 & = \sum_i \frac{\langle \ntk(\vx, \cdot), \phi_i \rangle^2}{\lambda_i}\\
    & = \sum_i \frac{\langle \sum_j \lambda_j \phi_j(\vx) \phi_j, \phi_i \rangle^2}{\lambda_i}\\
    & = \sum_i \lambda_i \phi_i(\vx)^2\\
    & = \ntk(\vx, \vx)\\
    & = 1
\end{split}
\end{equation*}
The second to last equality uses the orthonormality of $\phi_i$s, and the last equation follows from the definition of the NTK. We have $\norm{f}_\ntk \leq B$ which concludes the proof.
\end{proof}
\section{Details of \nnalg{} and \cnnalg{} }  \label{app:nn_supvar}
Algorithm \ref{alg:approx_ntkucb} and \ref{alg:supnucb} present \nnalg, and its Sup variant, respectively. The construction of the Sup variant is the same as for \ntkalg, with minor changes to the conditions of the \emph{if} statements.  

\begin{algorithm}[ht] 
\DontPrintSemicolon
\KwInput{$m,\,L,\,J,\,\eta,\,\sigma,\,\beta_t,\, T$}
\init{ network parameters to a random $\vtheta^0$, $\hat\mZ_0 = \sigma^2 \mI$}{}
\For{$t = 1 \cdots T$}{
Observe the context $\vz_t$.
 \For{$\va \in \mathcal{A}$}
{
Define $\vx := \vz_t\va$,\\
$\hat{\sigma}^2_{t-1}(\vx) \leftarrow \vg^T(\vx; \vtheta^0) \hat{\bm{Z}}^{-1}_{t-1} \vg(\vx; \vtheta^0)/m$\\
$U_{\va,t} \leftarrow f(\vx; \vtheta_{t-1}) + \sqrt{\beta_t}\hat\sigma_{t-1}(\vx)$
}
$\va_t = \argmax_{\va\in \mathcal{A}} U_{\va,t}$\\
Pick $\va_t$ and append the rewards vector $\vy_t$ by the observed reward. \\
$\hat{\bm{Z}}_{t} \leftarrow \sigma^2 \bm{I} + \sum_{i \leq t} \vg(\vx_i; \vtheta^0)\vg^T(\vx_i; \vtheta^0)/m$\\
$\vtheta_{t} \leftarrow \text{TrainNN}(m , L, J, \eta, \sigma^2, \vtheta^0, [\vx_i]_{i \leq t}, \vy_t)$\\
}
\caption{\label{alg:approx_ntkucb} \nnalg\\
$\sigma^2$ observation noise, $\beta_t$ exploration parameter, $J$ number of GD Steps, $\eta$ GD's learning rate, $m$ width of the network, $L$ depth of the network, $T$ total steps of the bandit}
\end{algorithm}

\begin{algorithm}[ht] 
\DontPrintSemicolon
\KwInput{$[\vx_i]_{i \leq t}$, $\vy_t$}
Define 
    $\Ls(\vtheta) = \sum_{i \in \Psi_t} (f(\vx_i; \vtheta) - y_i)^2 + m \sigma^2 \norm{\vtheta - \vtheta^0}_2^2$\\
 \For{$j = 0, \cdots, J-1$}
{
$\vtheta^{j+1} = \vtheta^j - \eta \nabla\Ls(\vtheta^j)$
}
\KwOutput{$\vtheta^J$}
\caption{\label{alg:trainnn} TrainNN$(m , L, J, \eta, \sigma^2, \vtheta^0, [\vx_i]_{i \leq t}, \vy_t)$}
\end{algorithm}

\begin{algorithm}[ht] 
\DontPrintSemicolon
\KwInput{$\Psi_t \subset \{ 1, \cdots, t-1\}$, $\vz_t$}
\init{
    $\hat{\bm{Z}} \leftarrow \sigma^2 \bm{I} + \sum_{i \in \Psi_t} \vg(\vx_i, \vtheta^0)\vg^T(\vx_i, \vtheta^0)/m$\\
    $\vtheta^{\text{(J)}} \leftarrow \text{TrainNN}(m , L, J, \eta, \sigma^2, \vtheta^0, [\vx_i]_{i \in  \Psi_t}, [y_i]_{i \in \Psi_t})
    $}{}
 \For{$\va \in \mathcal{A}$}
{
Define $\vx := \vz_t\va$,\\
$\hat{\sigma}^2_{t-1}(\vx) \leftarrow \frac{\vg^T(\vx, \vtheta^0)}{\sqrt{m}} \hat{\bm{Z}}^{-1} \frac{\vg(\vx, \vtheta^0)}{\sqrt{m}}$\\
\KwOutput{$f(\vx, \vtheta^{(J)})$ and $\hat{\sigma}^2_{t-1}(\vx)$ }
}
\caption{\label{alg:getpostnn}GetApproxPosterior\\
$J$ number of GD Steps, $\eta$ GD's learning rate, $m$ width of the network, $L$ depth of the network}
\end{algorithm}

\begin{algorithm}[ht]
\DontPrintSemicolon
\init{ 
    $\Psi^{(s)}_1 \leftarrow \emptyset$,  $\forall s \leq S$
    }{}
 \For{$t=1$ to $T$}
 {
 $s\leftarrow 1$, $A_1 \leftarrow \mathcal{A} $\\
 \While{action $\va_t$ not chosen}
 {
 $\Big(f(\vz_t\va, \vtheta^{\text{(J)}}),\hat{\sigma}_{t-1}^{(s)}(\vz_t\va)\Big) \leftarrow$ GetApproxPosterior$\big(\Psi_t^{(s)}, \vz_t\big)$ for all $\va \in A_{s}$\\
 \If{$\forall \va \in A_s$, $\sqrt{\beta_t} \hat{\sigma}_{t-1}^{(s)}(\vz_t\va) \leq \frac{\sigma}{T^2}$}
        {
        Choose $\va_t = \arg\max_{\va\in A_{s}}  f(\vz_t\va, \vtheta^{\text{(J)}})(\vz_t\va)+\sqrt{\beta_t}\hat\sigma_{t-1}^{(s)}(\vz_t\va) $,\\
        Keep the index sets $\Psi_{t+1}^{(s')} = \Psi_{t}^{(s')}$ for all $s' \leq S$.
        }
\ElseIf{$\forall \va \in A_s$, $\sqrt{\beta_t} \hat\sigma_{t-1}^{(s)}(\vz_t\va) \leq \sigma2^{-s}$}{
$A_{s+1} \leftarrow \Big\{ \va \in A_{s} \Big\vert  f(\vz_t\va, \vtheta^{\text{(J)}})+\sqrt{\beta_t}\sigma_{t-1}^{(s)}(\vz_t\va) \geq \max_{\va\in A_{s}}  f(\vz_t\va, \vtheta^{\text{(J)}})+\sqrt{\beta_t}\hat\sigma_{t-1}^{(s)}(\vz_t\va)  - \sigma 2^{1-s} \Big\}$\\
$s \leftarrow s + 1$}
\Else{
Choose $\va_t \in A_s$ such that $\sqrt{\beta_t} \hat\sigma_{t-1}^{(s)}(\vz_t\va_t) > \sigma2^{-s}$.\\
Update the index sets at all levels $s'\leq S$,\\
$
\Psi_{t+1}^{(s')} = \begin{cases}
      \Psi_{t+1}^{(s)}\cup \{t\} & \text{if $s'=s$}\\
      \Psi_{t+1}^{(s)} & \text{otherwise}\\
    \end{cases} 
$}
 }
 }
\caption{\label{alg:supnucb}Sup variant for \nnalg\\
$T$ number of total steps, $S = 2\log T$ number of discretization levels}
\end{algorithm}
\subsection{Proof of Theorem \ref{thm:nucbregret} }\label{app:thm4proof}
To bound the regret for \nnalg, we define an auxiliary algorithm that allows us to use the result from \ntkalg. Consider a kernelized UCB algorithm, which uses $\hat k$ an approximation of the NTK function, where $\hat k(\cdot,\cdot) = \vg^T(\cdot; \vtheta^0)\vg(\cdot; \vtheta^0)/m$. We argued in the main text that this kernel can well approximate $\ntk$, and its feature map $\hat \vphi(\vx) = \vg(\vx; \vtheta^0)/\sqrt{m}$ can be viewed as a finite approximation of $\vphi$, the feature map of the NTK. Throughout the proof, we denote the posterior mean and variance calculated via $\mathrm{GP}(0, \hat k)$ by $\hat \mu_{t-1}$ and $\hat \sigma_{t-1}$, respectively. In comparison to \nnalg, we use $\hat \mu_{t-1}$ instead of $f^{(J)}$ to approximate the true posterior mean, however $\hat \sigma_{t-1}$ has the same expression as in \nnalg. Using lemma \ref{lem:mu_to_f}, we reduce the problem of bounding the regret for \nnalg{} to the regret of this auxiliary method. We then repeat the technique used for Theorem \ref{thm:rkhsregret} on the auxiliary Sup variant which yields a regret bound depending on $\hat\gamma_T$, the information gain of the approximate kernel matrix. Finally, for width $m$ large enough, we bound $\hat\gamma_T$ with $\gamma_T$, information gain of the exact NTK matrix. 

%%%%%%%%%%%%%%%%%%%%%%%%%%%%%%%%%%%%%%%%%%
The following lemma provides the grounds for approximating the NTK with the empirical gram matrix of the neural network at initialization. Let $\mG = [\vg^T(\vx_t; \vtheta^0)]^T_{t\leq T}$.
\begin{lemma}[\citet{arora2019exact} Theorem 3.1]\label{lem:NTK_gram}
Set $0<\delta<1$. If $m = \Omega(L^6\log(TL/\delta)/\epsilon^4)$, then with probability greater than $1-\delta$,
\[
\norm{\mG^T\mG/m-\mNTK}_F \leq T\epsilon
\]
\end{lemma}
%%%%%%%%%%%%%%%%%%%%%%%%%%%%%%%%%%%%%%%%%%%%
The next lemma allows us to write the unknown reward as a linear function of the feature map over the finite set of points that are observed while running the algorithm.
 \begin{lemma}[\citet{zhou2019neural} Lemma 5.1] \label{lem:gu51} Let $f^*$ be a member of $\gH_\ntk$ with bounded RKHS norm $\norm{f}_\ntk \leq B$. If for some constant $C$, 
 \[m \geq \frac{C T^4|\mathcal{A}|^4 L^6}{\lambda_0^4} \log\big(T^2 |\mathcal{A}|^2L/\delta\big),\]
then for any $\delta \in (0,1)$, there exists $\vtheta^* \in \R^p$ such that
\[
f^*(\vx^i) = \langle \vg(\vx^i; \vtheta^0), \vtheta^* \rangle, \quad \sqrt{m}\norm{\vtheta^*}_2 \leq \sqrt{2}B
\]
with probability greater than $1-\delta$, for all $i \leq T|\mathcal{A}|$.
\end{lemma}
%%%%%%%%%%%%%%%%%%%%%%%%%%%%%%%%%%%%%
The following lemma acts as the bridge between \nnalg{} and the auxiliary UCB algorithm.
\begin{lemma}\label{lem:mu_to_f}
Fix $s\leq S$. Consider a given context set, $\{\vx_\tau\}_{\tau \in \Psi_t^{(s)}}$. Assume construction of $\Psi_t^{(s)}$ is such that the corresponding rewards, $y_\tau$ are statistically independent. Then there exists $C_1$, such that for any $\delta>0$, if the learning rate is picked $\eta = C_1(LmT+m\sigma^2)^{-1}$, and
\[
m \geq \mathrm{Poly}\Big(T, L, |\mathcal{A}|,\sigma^{-2},  \log(1/\delta)\Big).
\]
Then with probability of at least $1-\delta$, for all $i \leq T|\mathcal{A}|$,
\[
\vert f(\vx^i; \vtheta^{(J)}) - \hat\mu^{(s)}(\vx^i) \vert \leq \hat\sigma^{(s)}(\vx^i) \sqrt{\frac{TB}{m\eta\sigma^2}}\big( 3 + \sqrt{2}(1-m\eta\sigma^2 )^{J/2}\big) + \bar C(\frac{TB}{m\sigma^2})^{2/3}L^3\sqrt{m\log m} 
\]
for some constant $\bar C$, where $\hat\mu^{(s)}$ and $\hat\sigma^{(s)}$ are the posterior mean and variance of the reward with $\mathrm{GP}(0, \hat k)$ prior, after observing $(\vx_\tau, y_\tau)_{\tau \in \Psi_t^{(s)}}$.
\end{lemma}
%%%%%%%%%%%%%%%%%%%%%%%%%%%%%%%%%%%%%
The following lemma provides the central concentration inequality and is the \emph{neural} equivalent of Lemma \ref{lem:valko1}.
\begin{lemma}[Concentration of $f$ and $f^{(J)}$, Formal] \label{lem:neuralvalko1}
Fix $s\leq S$. Consider a given context set, $\{\vx_\tau\}_{\tau \in \Psi_t^{(s)}}$. Assume construction of $\Psi_t^{(s)}$ is such that the corresponding rewards, $y_\tau$ are statistically independent. Let $\delta>0$, $\eta = C_1(LmT+m\sigma^2)^{-1}$, and
\[
m \geq \mathrm{Poly}\Big(T, L, |\mathcal{A}|,\lambda_0^{-1}, \sigma^{-2},  \log(1/\delta)\Big).
\]
Then for any action $\va \in \mathcal{A}$, and for some constant $\bar C$ with probability of at least $1-2|\mathcal{A}|e^{-\beta_T/2}-\delta$,
\begin{equation*}
    \begin{aligned}
\vert f(\vx; \vtheta^{(J)}) - f^*(\vx) \vert \leq \hat\sigma^{(s)}(\vx) \Big(2B\sqrt{\beta_T} &+ \sigma \sqrt{\frac{2}{m}}B  + \sqrt{\frac{TB}{m\eta \sigma^2}}\big( 3 + \sqrt{2}(1-m\eta\sigma^2 )^{J/2}\big) \Big) \\
& + \bar C(\frac{TB}{m\sigma^2})^{2/3}L^3\sqrt{m\log m} 
\end{aligned}
\end{equation*}
where $\vx = \vz_t \va$.
\end{lemma}
%%%%%%%%%%%%%%%%%%%%%%%%%%%%%%%%%%%%%
\begin{lemma}\label{lem:gammahat}
If for any $0<\delta<1$
\[
m = \Omega \Big(T^6L^6\log(TL/\delta) \Big),
\]
then with probability greater than $1-\delta$
\[
\hat\gamma_T \leq \gamma_T + \sigma^{-2}
\]
\end{lemma}
%------------------------------------------------------------------------------------------------------------------------------------------------------------------------------
We are now ready to give the proof. 
\begin{proof}[Proof of Theorem \ref{thm:nucbregret}]
Construction of $\Psi_t^{(s)}$ is the same in both Algorithm \ref{alg:supnucb} and Algorithm \ref{alg:supcgp}. Hence, Proposition \ref{prop:supcgp_props} immediately follows.
It is straightforward to show that lemmas \ref{lem:cardofpsi}, \ref{lem:supcgp_properties}, \ref{lem:valko5} all apply to \supnnalg{}. We consider them for the approximate feature map $\hat{\bm\phi}$ and use lemma \ref{lem:gu51} to write the unknown reward $f^* = \bm\phi^T(\vx)\vtheta^*$ with high probability. By the union bound, all lemmas hold for the \supnnalg{} setting, with a probability greater than $1-2T|\mathcal{A}|e^{-\beta/2}-\delta$ for any $0<\delta<1$.
\par Recall $H_{t-1}$ is the history of the algorithm at time $t$,
\[
H_{t-1} := \Big\{(\vz_i,\vs_i, y_i)\Big\}_{i<t} \cup {\vz_t}
\]
Similar to proof of Theorem \ref{thm:rkhsregret}, we apply the Azuma-Hoeffding bound (Lemma \ref{lem:Azuma}) to the random variables, $X_t=: f^*(\vx_t^*) - \vf^*(\vx_t)$. We get, with probability greater than $1-2T|\mathcal{A}|e^{-\beta_T/2}$,
\begin{equation}
    \label{eq:azumaonregretnucb}
    \big \vert R(T) - \mathbb{E} \big[R(T) | H_{T-1}\big] \big \vert \leq \sqrt{4T \log (\frac{1}{T|\mathcal{A}|e^{-\beta_T/2}})}
\end{equation}
We now bound $\mathbb{E} \big[ R(T) | H_{t-1}\big]$. Let $\bar{\Psi} := \{ t\leq T |\, \forall s, \, t \notin \Psi_{T}^{(s)} \}$. By lemmas \ref{lem:supcgp_properties} and \ref{lem:valko5} applied to $\hat k$ the approximate kernel, with probability of at least $1-ST\vert\mathcal{A}\vert E^{-\beta_T/2}$,
\begin{equation*}
\begin{split}
        \mathbb{E} \big[ R(T) | H_{t-1}\big] & = \sum_{\substack{\mathrm{all}\\ t \notin \bar{\Psi}}} f^*(\vx_t^*) - f^*(\vx_t) + \sum_{t \in \bar{\Psi}} f^*(\vx_t^*) - f^*(\vx_t)\\
        & \leq 8 S \sqrt{\beta_T  (10+\sigma^{-2}15)\hat\gamma_T T\log T} + \underbrace{\sum_{t \in \bar{\Psi}} f^*(\vx_t^*) - f^*(\vx_t)}_{I},
\end{split}
\end{equation*}
where $\hat\gamma_T$ is the information gain corresponding to the approximate kernel. Applying Lemma \ref{lem:neuralvalko1}, with probability of greater than $1-2T|\mathcal{A}|e^{-\beta_T/2}$,
\begin{align*}
    I & \leq \sum_{t \in \bar{\Psi}} \big( \hat\sigma^{(s)}(\vx_t) + \hat\sigma^{(s)}(\vx^*_t)\big)
        \Big(2B\sqrt{\beta_T}  + \sigma \sqrt{\frac{2}{m}}B  + \sqrt{\frac{TB}{m\eta \sigma^2}}\big( 3 + \sqrt{2}(1-m\eta\sigma^2 )^{J/2}\big) \Big) \\
        &\qquad \quad+ \sqrt{\beta_T}\hat\sigma^{(s)}(\vx_t)  +2\bar C(\frac{TB}{m\sigma^2})^{2/3}L^3\sqrt{m\log m}\\
    & \leq \sum_{t \in \bar{\Psi}} 4B \frac{\sigma}{T^2} + \frac{2\sigma^2B}{T^2} \sqrt{\frac{2}{m\beta_T}}  + 2\sqrt{\frac{B}{m\eta\beta_T T^2}}\big( 3 + \sqrt{2}(1-m\eta\sigma^2 )^{J/2}\big) \\
    & \qquad\quad + 2\bar C(\frac{TB}{m\sigma^2})^{2/3}L^3\sqrt{m\log m} \\
        & \leq 4B \frac{\sigma}{T} + 2\sigma^2B\sqrt{\frac{2}{m\beta_TT^2}} + 2\sqrt{\frac{B}{m\eta \beta_T}}\big( 3 +\sqrt{2} (1-m\eta\sigma^2 )^{J/2}\big) \\
        & \qquad\quad+ 2\bar CT(\frac{TB}{m\sigma^2})^{2/3}L^3\sqrt{m\log m} 
\end{align*}
\looseness -1 The next to last inequality holds by the construction of $\bar \Psi$, and the last one by $|\bar\Psi| \leq T$. Set $\delta \leq T|\mathcal{A}|e^{-\beta_T/2}$, $S = \log T$ and choose $m$ large enough according to the conditions of Theorem \ref{thm:nucbregret}. Putting together the two terms,
\begin{equation*}
\begin{split}
    R(T) & \leq 4B \frac{\sigma}{T} + 2 \sqrt{\frac{B(TL+\sigma^2)}{C_1\beta_T}}\big( 3 + \sqrt{2}(1-\frac{\sigma^2C_1}{TL+\sigma^2})^{J/2}\big)\\
    & \quad+ 8 \sqrt{\beta_T  (10+\sigma^{-2}15)\hat\gamma_T T(\log T)^3}+ \sqrt{4T \log (\frac{1}{T|\mathcal{A}|e^{-\beta_T/2}})}\\
\end{split}
\end{equation*}
Choosing $\beta_t = \beta_T = 2\log(2T|\mathcal{A}|/\tilde\delta)$, with probability greater than $1-(\log T + 1)\tilde\delta$,
\begin{equation*}
\begin{split}
    R(T) & \leq 2(1+B)\frac{\sigma}{T} + 8\sqrt{2\log(2T|\mathcal{A}|/\tilde\delta)}\sqrt{\hat\gamma_T T(\log T)^3(10+\sigma^{-2}15)}\\
    & \quad + 2 \sqrt{\frac{B(TL+\sigma^2)}{2C_1\log(2T|\mathcal{A}|/\tilde\delta)}}\big( 3 + \sqrt{2}(1-\frac{\sigma^2C_1}{TL+\sigma^2})^{J/2}\big) + 2 \sqrt{T \log(2/\tilde\delta)}.
\end{split}
\end{equation*}
We conclude the proof by using Lemma \ref{lem:gammahat}, to bound $\hat\gamma_T$ with $\gamma_T$, and substitute $\tilde\delta$ with $\tilde\delta/(1+ \log T)$.
\end{proof}
Before giving the proof of the Lemmas, we present Condition \ref{cond:m_eta}. Lemmas \ref{lem:gu51} through \ref{lem:gammahat}, often rely on the width $m$ being large enough, or the learning rate being small enough. For easier readability, we put those inequalities together and present it at $m$ satisfying a certain polynomial rate and $\eta$ being of order $1/m$.

\begin{condition} \label{cond:m_eta}
The following conditions on $m$ and $\eta$, ensure convergence in the lemmas used in this section. For every $t\leq T$,
\begin{align*}
    & 2 \sqrt{t/(m \sigma^2)} \geq C_1 \Big[\frac{mL}{\log(TL^2|\mathcal{A}|/\sigma^2)}\Big]^{-3/2}, \\
    & 2 \sqrt{t/(m \sigma^2)} \leq C_2 \min \big\{L^{-6} (\log m)^{-3/2}, \big(m(\eta\sigma^2)^2 L^{-6}t^{-1}(\log m)^{-1}\big)^{3/8}\big\}, \\
    & \eta \leq C_3(m \sigma^2+ tmL)^{-1},\\
    & m^{1/6} \geq C_4 \sqrt{\log m} L^{7/2}(t/\sigma^2)^{7/6}(1+\sqrt{t/\sigma^2}).
\end{align*}
Setting $m$ and $\eta$ as follows satisfies the inequalities above.
\begin{align*}
    &m \geq \mathrm{Poly}\Big(T, L, |\mathcal{A}|,\sigma^{-2},  \log(1/\delta)\Big),\\
    &\eta = C(mTL+m\sigma^2)^{-1}
\end{align*}
\end{condition}
%--------------------------------------------------------------------------------------------
%--------------------------------------------------------------------------------------------
\subsubsection{Proof of lemma \ref{lem:mu_to_f}} \label{app:proofof53}
We first give an overview of the proof. Recall that $\vtheta^{(J)}$ is the result of running gradient descent for $J$ steps on 
\[
\Ls(\vtheta) =  \sum_{i\leq t} (f(\vx_i;\vtheta)^2 - y_i)^2 + m\sigma^2\norm{\vtheta-\vtheta^0}_2^2.
\]
The network $f(\vx; \vtheta)$ is a complex nonlinear function, and it is not feasible to produce the expression for the updates $\vtheta^j$ at a step $j$ of the gradient descent algorithm. Working with the first order Taylor expansion of the network around initialization, $\langle \vg(\vx;  \vtheta^0), \vtheta - \vtheta^0\rangle$,  we instead consider running the gradient descent on
\[
\tilde\Ls(\vtheta) =  \sum_{i\leq t} \big(\langle\vg^T(\vx_i; \vtheta^0),\vtheta-\vtheta^0\rangle - y_i\big)_2^2     + m \sigma^2 \norm{\vtheta-\vtheta^0}_2^2.
\]
\looseness -1 Let $\tilde\vtheta^j$ denote the GD update at step $j$. It can be proven that gradient descent follows a similar path in both scenarios, i.e. the sequences $(\vtheta^j)_{j>1}$ and $(\tilde\vtheta^j)_{j>1}$ remain close for all $j$ with high probability. Now gradient descent on $\tilde\Ls$ converges to the global optima, $\tilde\vtheta^*$, for which we have $\langle\vg^T(\vx_i; \vtheta^0),\tilde\vtheta^*-\vtheta^0\rangle = \hat\mu(\vx)$, concluding the proof. 
\begin{proof}
Let $\vf(\vtheta) = [f(\vx_\tau; \vtheta)]_{\tau \in \Psi_t^{(s)}}$, and $\vy = [y_\tau]_{\tau \in \Psi_t^{(s)}}$. Consider the loss function used for training the network
\begin{equation}\label{eq:loss1}
    \Ls_1(\vtheta) = \norm{\vf(\vtheta) - \vy}_2^2 + m \sigma^2 \norm{\vtheta}_2^2
\end{equation}
And let the sequence of $(\vtheta^j)$ denote the gradient descent updates. The following lemma shows that at step $j$ of the GD, $f(\vx; \vtheta^j)$ can be approximated with its first order Taylor approximation at initialization, $\langle \vg(\vx; \vtheta^0), \vtheta^j-\vtheta^0\rangle$.
\begin{lemma} \label{lem:f_with_gtheta}
There exists constants $(C_i)_{i\leq4}$, such that for any $\delta>0$, if $\eta$ and $m$ satisfy Condition \ref{cond:m_eta}, then,
\[
\abs{f(\vx^i; \vtheta^J)-f(\vx^i; \vtheta^0)-\langle \vg(\vx^i; \vtheta^0), \vtheta^J - \vtheta^0 \rangle} \leq C (\frac{TB}{m\sigma^2})^{2/3} L^3 \sqrt{m\log m}
\]
for some constant $C$ with probability greater than $1-\delta$, for any $i \leq T |\mathcal{A}|$.
\end{lemma}
Lemma above holds, as $m$ and $\eta$ are chosen such that condition \ref{cond:m_eta} is met. We now show that $\langle \vg(\vx^i; \vtheta^0), \vtheta^J - \vtheta^0 \rangle$ approximates $\hat\mu^{(s)}$ well. Let
\begin{align*}
 \mG &= [\vg^T(\vx_\tau; \vtheta^0)]^T_{\tau \in \Psi_t^{(s)}},\\
 \hat\mZ &= \sigma^2\mI + \sum_{\tau \in \Psi_t^{(s)}}  \vg(\vx_\tau; \vtheta^0) \vg^T(\vx_\tau; \vtheta^0)/m= \sigma^2\mI + \mG^T\mG/m.
\end{align*}
Let $\tilde\vtheta^j$ to be the GD updates of $\Ls_2$, the loss function corresponding to the auxiliary UCB algorithm
\begin{equation} \label{eq:loss2}
    \Ls_2(\vtheta) = \norm{\mG^T(\vtheta-\vtheta^0) - \vy}_2^2 + m \sigma^2 \norm{\vtheta-\vtheta^0}_2^2.
\end{equation}
Lemma \ref{lem:gd_conv_ridge} states the nice convergence properties of this optimization problem. 
\begin{lemma}[\citet{zhou2019neural} Lemma C.4]\label{lem:gd_conv_ridge}
If $\eta$, $m$ satisfy conditions \ref{cond:m_eta}, then
\begin{align*}
& \norm{\tilde\vtheta^j - \vtheta^0 - \hat\mZ^{-1} \mG^T\vy/m}_2 \leq (1-\eta m \sigma^2)^{j/2} \sqrt{2TB/(m \sigma^2)}\\
& \norm{\tilde\vtheta^j - \vtheta^0}_2 \leq \sqrt{\frac{TB}{m\sigma^2}}
\end{align*}
with a probability of at least $1-\delta$, for any $j \leq J$.
\end{lemma}
Furthermore, we can show that in space of $\vtheta$, the path of gradient descent on $\Ls_2$ follows GD's path on $\Ls_1$. In other words, as the width $m$ grows the sequences $(\vtheta^j)_{j \leq J}$ and $(\tilde\vtheta^j)_{j \leq J}$ converge uniformly.
\begin{lemma} \label{lem:thetatilde_theta}
There exists constants $(C_i)_{i\leq 4}$, such that for any $\delta>0$, if $\eta$ and $m$ satisfy Condition \ref{cond:m_eta}, then,
\[\norm{\vtheta^J - \tilde{\vtheta}^J}_2 \leq 3\sqrt{\frac{TB}{m\sigma^2}}\]
with probability greater than $1-\delta$.
\end{lemma}
We now have all necessary ingredients to finish the proof. For simplicity, here we denote $\vg(\vx; \vtheta^0)$ by $\vg$. Applying Cauchy-Schwartz inequality we have,
\begin{equation} \label{eq:gtheta1}
\begin{split}
    \langle \vg, \vtheta^J - \vtheta^0 \rangle & = \langle \vg, \vtheta^J - \tilde\vtheta^J \rangle + \langle \vg, \tilde\vtheta^J - \vtheta^0 \rangle\\
    & \leq \norm{\vg}_{\hat\mZ^{-1}} \norm{\vtheta^J - \tilde\vtheta^J}_{\hat\mZ} + \langle \vg, \tilde\vtheta^J - \vtheta^0 \rangle\\
    & \leq \frac{1}{\sqrt{m\eta}}\norm{\vg}_{\hat\mZ^{-1}} \norm{\vtheta^J - \tilde\vtheta^J}_2 + \langle \vg, \tilde\vtheta^J - \vtheta^0 \rangle\\
    & \leq 3\norm{\frac{\vg}{\sqrt{m}}}_{\hat\mZ^{-1}} \sqrt{\frac{TB}{m\eta\sigma^2}} + \langle \vg, \tilde\vtheta^J - \vtheta^0 \rangle
\end{split}    
\end{equation} 
Recall that $\hat\mZ= \sigma^2\mI + \mG^T\mG$. By Lemma \ref{lem:boundG}, $\norm{\mG}_F \leq C\sqrt{TLm}$ and we have,
\begin{equation} \label{eq:Khat_upper}
    \hat \mZ \preccurlyeq (\sigma^2 + CTLm)\mI \preccurlyeq \frac{1}{m\eta}\mI,
\end{equation}
since $\eta$ is set such that $\eta \leq C(m\sigma^2+TLm)^{-1}$. Therefore, for any $\vx\in \R^p$, $\norm{\vx}_{\hat\mZ} \leq \frac{1}{\sqrt{m\eta}}\norm{\vx}_2$, and follows the second inequality. For the third inequality we have used Lemma \ref{lem:thetatilde_theta}. Define $\vb = \sum_{i\leq T}y_i\vg(\vx_i; \vtheta^0)/\sqrt{m}$. Decomposing the second term of the right hand side in equation \ref{eq:gtheta1} we get,
\begin{equation} \label{eq:gtheta2}
    \begin{split}
         \langle \vg, \tilde\vtheta^J - \vtheta^0 \rangle &= \langle \vg, \frac{\hat\mZ\vb}{\sqrt{m}}\rangle + \langle \vg, \tilde\vtheta^J - \vtheta^0 - \frac{\hat\mZ\vb}{\sqrt{m}} \rangle  \\
         & \leq \frac{\vg^T\hat\mZ\vb}{\sqrt{m}} + \frac{1}{\sqrt{\eta}}\norm{\frac{\vg}{\sqrt{m}}}_{\hat\mZ^{-1}}\norm{\tilde\vtheta^J - \vtheta^0 - \frac{\hat\mZ\vb}{\sqrt{m}}}_2 \\
         & \leq \frac{\vg^T\hat\mZ\vb}{\sqrt{m}} + \norm{\frac{\vg}{\sqrt{m}}}_{\hat\mZ^{-1}}\sqrt{\frac{2TB}{m\eta \sigma^2}} (1-\eta m \sigma^2)^{J/2} 
    \end{split}
\end{equation}
The second inequality follows from the convergence of GD to $\Ls_2$, given in Lemma \ref{lem:gd_conv_ridge}. By the definition of posterior mean and variance we have,
\begin{align*}
    &\hat\mu^{(s)}(\vx) = \frac{\vg(\vx; \vtheta^0)^T\hat\mZ\vb}{\sqrt{m}}\\
    &\hat\sigma^{(s)}(\vx) = \norm{\frac{\vg(\vx; \vtheta^0)}{\sqrt{m}}}_{\hat\mZ^{-1}}.
\end{align*}
The upper bound on $f(\vx; \vtheta^J) - \hat\mu^{(s)}(\vx)$ follows from plugging in Equation \ref{eq:gtheta2} into Equation \ref{eq:gtheta1}, and applying Lemma \ref{lem:f_with_gtheta}. Similarly, for the lower bound we have,
\begin{align} \label{eq:lowerbound1}
    - f(\vx^i; \vtheta^J) &\leq \langle \vg,  \vtheta^0 - \vtheta^J \rangle + C (\frac{TB}{m\sigma^2})^{2/3} L^3 \sqrt{m\log m}\\
    \label{eq:lowerbound2}
    \langle \vg,  \vtheta^0 - \tilde\vtheta^J \rangle &\leq -\hat\mu^{(s)}(\vx)  + \hat\sigma^{(s)}(\vx)\sqrt{\frac{2TB}{m\eta \sigma^2}} (1-\eta m \sigma^2)^{J/2} \\
    \label{eq:lowerbound3}
     \langle \vg,  \tilde\vtheta^J - \vtheta^J \rangle &\leq 3\hat\sigma^{(s)}(\vx) \sqrt{\frac{TB}{m\eta\sigma^2}}
\end{align}
Where inequality \ref{eq:lowerbound1} holds by Lemma \ref{lem:f_with_gtheta}, and the next two inequalities are driven similarly to equations \ref{eq:gtheta1} and \ref{eq:gtheta2}. The lower bound results by putting together equations \ref{eq:lowerbound1}-\ref{eq:lowerbound3}, and this concludes the proof.
\end{proof}
%--------------------------------------------------------------------------------------------
\subsubsection{Proof of Lemma \ref{lem:neuralvalko1}}
\begin{proof}
Consider Lemma \ref{lem:valko1}, and substitute the approximate feature map $\hat{\bm\phi}(x) = \vg(\vx; \vtheta^0)$ for the NTK feature map $\bm\phi(\vx)$. For simplicity we denote $\vg(\vx; \vtheta^0)$ by $\vg$. $m$ is chosen such that lemma \ref{lem:gu51} holds. Then, by lemma \ref{lem:valko1} applied to $\hat{\bm\phi}$, with probability greater than $1-\delta-2|\mathcal{A}|e^{-\beta_T/2}$, 
\begin{equation*}
    \begin{split}
       \vert \hat\mu^{(s)}_t(\vx) - f^*(\vx) \vert & =  \vert \vg^T(\vx) (\mG^T\mG+\sigma^2\bm{I})^{-1}\mG^T\vy - \vg^T\bm{\theta}^* \vert \\
       & \leq 2B\sqrt{\beta_t}\hat\sigma^{(s)}_t(\vx) - \sigma^2 \vg^T(\vx)(\mG^T\mG+\sigma^2\bm{I})^{-1}\bm{\theta}^*
    \end{split}
\end{equation*}
for any $\vx = \vz_t\va$, $\vs \in \mathcal{A}$. By Lemma \ref{lem:gu51}, $\sqrt{m}\norm{\vtheta^*}_2 \leq \sqrt{2}B$, and plugging in Lemma \ref{lem:mu_to_f} concludes the proof.
\end{proof}

%--------------------------------------------------------------------------------------------
%--------------------------------------------------------------------------------------------
\subsubsection{Proof of Lemma \ref{lem:gammahat}}
\begin{proof}
We use some of the inequalities given in the proof of Lemma 5.4 \citet{zhou2019neural}. Consider an arbitrary sequence $(\vx_t)_{t\leq T}$. For the approximate feature map $\vg(\vx; \vtheta^0)/\sqrt{m}$, recall the definition of information gain after observing $T$ samples,
\[
I_{g} = \frac{1}{2}\log\det \big(\mI + \sigma^{-2}\mG_T\mG_T^T/m\big)
\]
where $\mG_T = [\vg(\vx_t; \vtheta^0)]^T_{t\leq T} \in \R^{T\times p}$. Let $[\mNTK]_{i,j\leq T} = \ntk(\vx_i, \vx_j)$ with $k$ the NTK function of the fully-connected $L$-layer network.
\begin{equation} \label{eq:Ihat_gamma}
    \begin{split}
        I_g & = \frac{1}{2}\log\det \big(\mI + \sigma^{-2}\mNTK + \sigma^{-2}(\mG_T\mG_T^T/m-\mNTK)\big)\\
        & \stackrel{\mathrm{(a)}}{\leq} \frac{1}{2}\log\det \big(\mI + \sigma^{-2}\mNTK\big) + \langle (\mI + \sigma^{-2}\mNTK)^{-1}, \sigma^{-2}(\mG_T\mG_T^T/m-\mNTK) \rangle\\
        & \leq I_k + \sigma^{-2}\norm{(\mI + \sigma^{-2}\mNTK)^{-1}}_F \norm{\mG_T\mG_T^T/m-\mNTK}_F\\
        & \stackrel{\mathrm{(b)}}{\leq}  I_k + \sigma^{-2}\sqrt{T} \norm{\mG_T\mG_T^T/m-\mNTK}_F \\
        & \stackrel{\mathrm{(c)}}{\leq}  I_k + \sigma^{-2}T\sqrt{T}\epsilon \\
        & \stackrel{\mathrm{(d)}}{\leq}  \gamma_T + \sigma^{-2}.
    \end{split}
\end{equation}
Inequality (a) holds by concavity of $\log\det(\cdot)$. Inequality (b) holds since $\mI  \preccurlyeq \mI + \sigma^{-2}\mNTK$. Inequality (c) holds due to Lemma \ref{lem:NTK_gram}. Finally, inequality (d) arises from the choice of $m$. Equation \ref{eq:Ihat_gamma} holds for any arbitrary context set, thus it also holds for the sequence which maximizes the information gain.
\end{proof}
%--------------------------------------------------------------------------------------------
\subsubsection{Proof of Other Lemmas in Section \ref{app:thm4proof}}
%--------------------------------------------------------------------------------------------
\begin{proof}[\textbf{Proof of Lemma \ref{lem:f_with_gtheta}}] Let $m$, $\eta$ satisfy the first two conditions in the lemma. By \citet{cao2019generalization} Lemma 4.1, if $\norm{\vtheta^J - \vtheta^0}_2 \leq \tau$, then
\[\abs{f(\vx^i; \vtheta^J)-f(\vx^i; \vtheta^))-\langle \vg(\vx^i; \vtheta^0), \vtheta^J - \vtheta^0 \rangle} \leq C \tau^{4/3} L^3 \sqrt{m\log m}.\]
Under the given conditions however, Lemma B.2 from \citet{zhou2019neural} holds and we have,
\[\norm{\vtheta^J - \vtheta^0}_2 \leq 2\sqrt{\frac{TB}{m\sigma^2}}.\]
\end{proof}
%--------------------------------------------------------------------------------------------
\begin{proof}[\textbf{Proof of lemma \ref{lem:gd_conv_ridge}}] This proof repeats proof of lemma C.4 in \citet{zhou2019neural} and is given here for completeness. $\Ls_2$ is $m\sigma^2$-strongly convex, and $C(TmL+m\sigma^2)$-smooth, since
\[
\nabla^2\Ls_2 = G^TG + m\sigma^2 \mI \leq (\norm{G}_2^2 + m\sigma^2)\mI \leq C (tmL+m\sigma^2) \mI,
\]
where the inequality follows from lemma \ref{lem:boundG}. It is widely known that gradient descent on smooth strongly convex functions converges to optima given that the learning rate is smaller than the smoothness coefficient inversed. Moreover, the minima of $\Ls_2$ is unique and has the closed form
\[
\tilde\vtheta^* = \vtheta^0 +  \hat\mZ^{-1} \mG^T\vy/m
\]
Having set $\eta \leq C(tmL+m\sigma^2)^{-1}$, we get that $\tilde\vtheta^j$ converges to $\tilde\vtheta^*$ with exponential rate,
\begin{align*}
    \norm{\tilde\vtheta^j - \vtheta^0 - \hat\mZ^{-1} \mG^T\vy/m}_2^2 & \leq (1-\eta m \sigma^2)^j \frac{2}{m\sigma^2}(\Ls_2(\vtheta^0)-\Ls_2(\tilde\vtheta^*))\\
    & \leq \frac{2(1-\eta m \sigma^2)^j}{m \sigma^2} \norm{y}_2^2\\
    & \leq \frac{2TB(1-\eta m \sigma^2)^j}{m \sigma^2}.
\end{align*}
The second inequality holds due to $\Ls_2 \geq 0$. From the RKHS assumption, the true reward is bounded by $B$ and hence the last inequality follows from $|\Psi_t^{(s)}| \leq T$. Strong Convexity of $\Ls_2$, guarantees a monotonic decrease of the loss and we have,
\begin{align*}
    m\sigma^2 \norm{\tilde\vtheta^J - \tilde\vtheta^0}_2^2 & \leq m\sigma^2 \norm{\tilde\vtheta^J - \tilde\vtheta^0}_2^2 + \norm{\mG^T(\tilde\vtheta^J - \tilde\vtheta^0)-\vy}_2^2\\
    & \leq m\sigma^2 \norm{\tilde\vtheta^0 - \tilde\vtheta^0}_2^2 + \norm{\mG^T(\tilde\vtheta^0 - \tilde\vtheta^0)-\vy}_2^2\\
    & \leq \norm{\vy}^2_2\\
    & \leq TB
\end{align*}
\end{proof}
%--------------------------------------------------------------------------------------------
\begin{proof}[\textbf{Proof of lemma \ref{lem:thetatilde_theta}}]
Choose $m$, $\eta$ such that they satisfy condition \ref{cond:m_eta}. By lemma B.2 \citet{zhou2019neural}, $\norm{\vtheta^J - \vtheta^0}_2 \leq 2\sqrt{TB/(m\sigma^2)}$. 
\begin{align*}
    \norm{\vtheta^J - \tilde\vtheta^J}_2 & \leq \norm{\vtheta^J - \vtheta^0}_2 + \norm{\tilde\vtheta^J - \tilde\vtheta^0}_2\\
    & \leq 3 \sqrt{\frac{TB}{m\sigma^2}.}
\end{align*}
Where the second inequality holds by Lemma \ref{lem:gd_conv_ridge}.
\end{proof}
%--------------------------------------------------------------------------------------------
\begin{lemma}\label{lem:boundG}
Consider the fixed set $\{\vx_i\}_{i\leq t}$ of inputs. Let $\mG = [\vg^T(x_i; \vtheta^0)]_{i\leq t}^T$, where $\vg$ shows the gradients of a $L$-layer feedforward network of width $m$ at initialization. There exists constants $(C_i)_{i\leq 4}$ such that if for any $\delta >0$, $m$ satisfies condition \ref{cond:m_eta}, then, with probability greater than $1-\delta$.
\[
\norm{\mG}_F \leq C \sqrt{tmL}
\]
for some constant $C$.
\end{lemma}
\begin{proof}[\textbf{Proof of Lemma \ref{lem:boundG}}]
From Lemma B.3 \citet{cao2019generalization}, we have $\norm{\vg(\vx_i; \vtheta^0)}_2 \leq \bar C \sqrt{mL}$ with high probability. By the definition of Frobenius norm, it follows,
\begin{equation*}
    \norm{\mG}_F \leq \sqrt{t} \max_{i \leq t} C\norm{\vg(\vx_i; \vtheta^0)}_2 \leq C \sqrt{tmL}
\end{equation*}
\end{proof}
\subsection{Proof of Theorem \ref{thm:cnnucbregret}} \label{app:thm6proof}
This proof will closely follow the proof of Theorem \ref{thm:nucbregret}, with small adjustments to the condition on $m$. We begin by giving intuition on why this is the case. In this section, $m$ refers to the number of channels of the convolutional network. Recall that,
\[
\bar f(\vx;\vtheta) := f_{\mathrm{CNN}}(\vx; \vtheta) = \frac{1}{d} \sum_{l=1}^{d}f_{\mathrm{NN}}(c_l\cdot \vx; \vtheta).
\]
For simplicity in notation, from now on we will refer to $f_{\mathrm{NN}}(\vx; \vtheta)$ just as $f(\vx; \vtheta)$ and use a ``bar'' notation to refer to the convolutional counterpart of a variable, to emphasize that the 2-layer convolutional network is the average of the 2-layer fully-connected networks taken over all circular shifts of the input. 
\[\bar k(\vx, \vx') = \cntk(\vx, \vx') := \frac{1}{d} \sum_{l=1}^{d}\ntk(c_l\cdot \vx,\vx'),\]
Then it is straight-forward to show that 
\[\bar g (\vx; \vtheta^0):= \nabla_\vtheta f_{\mathrm{CNN}}(\vx; \vtheta^0) = \frac{1}{d} \sum_{l=1}^{d} g (c_l\cdot\vx; \vtheta^0)\]
where the vector $\vtheta^0$ is referring to the same set of parameter in both case. Starting with an identical initialization, and running gradient descent on the $\ell_2$ regularized LSE loss will cause $\bar\vtheta^{(J)} := \vtheta^{(J)}_{\mathrm{CNN}}$ and $\vtheta^{(J)}$ not to be equal anymore, but we can still show that $\bar\vtheta^{(J)}$ and $\vtheta^0$ are close as it was in the fully-connected case. Similarly, we can show that $\bar \vg$ faces a small change during training with gradient descent. We will now present the lemmas needed for proving Theorem \ref{thm:cnnucbregret}. These are equivalent to the lemmas in Section \ref{app:thm4proof} repeated for the convolutional network. Some statements, however, are weaker with respect to the condition on $m$, since convergence of the gram matrix to the CNTK (Lemma \ref{lem:CNTK_gram}) is only proven in the $m\rightarrow \infty$ limit.
%%%%%%%%%%%%%%%%%%%%%%%%%%%%%%%%%%%%%%%%%%
\begin{lemma}[Convolutional variant of Lemma \ref{lem:NTK_gram}]\label{lem:CNTK_gram} Let $\bar \mG = [\bar \vg^T(\vx_t; \vtheta^0)]^T_{t\leq T}$ and $\bar\mK = [\bar k(\vx_i, \vx_j)]_{i,j \leq T}$. For any $\epsilon>0$, there exists $M$ such that for every $m \geq M$,
\[
\norm{\bar\mG^T\bar\mG/m- \bar\mK}_F \leq T\epsilon
\]
\end{lemma}
%%%%%%%%%%%%%%%%%%%%%%%%%%%%%%%%%%%%%
 \begin{lemma}[Convolutional variant of Lemma \ref{lem:gu51}] \label{lem:gu51_conv} Let $f^*$ be a member of $\gH_\cntk$ with bounded RKHS norm $\norm{f}_\cntk \leq B$. Then there exists $M$ such that for every $m \geq M$, there is a $\vtheta^* \in \R^p$ that satisfies
\[
f^*(\vx^i) = \langle \bar\vg(\vx^i; \vtheta^0), \vtheta^* \rangle, \quad \sqrt{m}\norm{\vtheta^*}_2 \leq \sqrt{2}B
\]
for all $i \leq T|\mathcal{A}|$.
\end{lemma}
%%%%%%%%%%%%%%%%%%%%%%%%%%%%%%%%%%%%%
\begin{lemma}[Convolutional variant of Lemma \ref{lem:mu_to_f}]\label{lem:mu_to_f_conv}
Fix $s\leq S$. Consider a given context set, $\{\vx_\tau\}_{\tau \in \Psi_t^{(s)}}$. Assume construction of $\Psi_t^{(s)}$ is such that the corresponding rewards, $y_\tau$ are statistically independent. Then there exists $C_1$, such that for any $\delta>0$, if the learning rate is picked $\eta = C_1(LmT+m\sigma^2)^{-1}$, and
\[
m \geq \mathrm{Poly}\Big(T, L, |\mathcal{A}|,\sigma^{-2},  \log(1/\delta)\Big).
\]
Then with probability of at least $1-\delta$, for all $i \leq T|\mathcal{A}|$,
\[
\vert \bar f(\vx^i; \bar \vtheta^{(J)}) - \hat{\bar \mu}^{(s)}(\vx^i) \vert \leq \hat{\bar\sigma}^{(s)}(\vx^i) \sqrt{\frac{TB}{m\eta\sigma^2}}\big( 3 + (1-m\eta\sigma^2 )^{J/2}\big) + \bar C(\frac{TB}{m\sigma^2})^{2/3}L^3\sqrt{m\log m} 
\]
for some constant $\bar C$, where $\hat{\bar \mu}^{(s)}$ and $\hat{\bar\sigma}^{(s)}$ are the posterior mean and variance of the reward with $\mathrm{GP}(0, \hat{K}_{\mathrm{CNN}})$ prior, after observing $(\vx_\tau, y_\tau)_{\tau \in \Psi_t^{(s)}}$.
\end{lemma}
%%%%%%%%%%%%%%%%%%%%%%%%%%%%%%%%%%%%%
\begin{lemma}[Convolutional variant of \ref{lem:neuralvalko1}] \label{lem:neuralvalko1_conv}
Fix $s\leq S$. Consider a given context set, $\{\vx_\tau\}_{\tau \in \Psi_t^{(s)}}$. Assume construction of $\Psi_t^{(s)}$ is such that the corresponding rewards, $y_\tau$ are statistically independent. Let $\delta>0$ and $\eta = C_1(LmT+m\sigma^2)^{-1}$. Then, there exists $M$ such that for all $m\geq M$, for any action $\va \in \mathcal{A}$, and for some constant $\bar C$ with probability of at least $1-2|\mathcal{A}|e^{-\beta_T/2}-\delta$,
\begin{equation*}
    \begin{aligned}
\vert \bar f(\vx; \bar \vtheta^{(J)}) - f^*(\vx) \vert \leq \hat{\bar \sigma}^{(s)}(\vx) \Big(2B \sqrt{\beta_T} &+ \sigma \sqrt{\frac{2}{m}}B  + \sqrt{\frac{TB}{m\eta \sigma^2}}\big( 3 + (1-m\eta\sigma^2 )^{J/2}\big) \Big) \\
& + \bar C(\frac{TB}{m\sigma^2})^{2/3}L^3\sqrt{m\log m} 
\end{aligned}
\end{equation*}
where $\vx = \vz_t \va$.
\end{lemma}
%%%%%%%%%%%%%%%%%%%%%%%%%%%%%%%%%%%%%
\begin{lemma}[Convolutional variant of Lemma \ref{lem:gammahat}]\label{lem:gammahat_conv}
There exists $M$ such that for all $m\geq M$,
\[
\hat{\bar\gamma}_T \leq \bar \gamma_T + \sigma^{-2}
\]
\end{lemma}
\begin{proof}[Proof of Theorem \ref{thm:cnnucbregret}]
Repeating the proof of Theorem \ref{thm:nucbregret}, and plugging in Lemmas \ref{lem:CNTK_gram} through \ref{lem:gammahat_conv} instead of Lemmas \ref{lem:NTK_gram}-\ref{lem:gammahat} concludes the result. 
\end{proof}
%%%%%%%%%%%%%%%%%%%%%%%%%%%%%%%%%%%%%
%%%%%%%%%%%%%%%%%%%%%%%%%%%%%%%%%%%%%
\subsubsection{Proof of Lemmas in Section \ref{app:thm6proof}}
\begin{proof}[Proof of Lemma \ref{lem:CNTK_gram}]
From \citet{arora2019exact}, for any $\vx$, $\vx'$ on the hyper-sphere, with probability one,
\[
\lim_{m \rightarrow \infty} \langle\bar \vg(\vx; \vtheta),\bar \vg(\vx'; \vtheta)\rangle = \cntk(\vx,\vx').
\]
In other words, for every $\epsilon>0$, there exists $M$ such that for all $m \geq M$, with probability one,
\[
\vert \langle\bar \vg(\vx; \vtheta),\bar \vg(\vx'; \vtheta)\rangle - \cntk(\vx,\vx')\vert \leq \epsilon
\]
Recall that $\bar \mG = [\bar \vg^T(\vx_t; \vtheta^0)]^T_{t\leq T}$. Let $M_l$ denote the number of channels that satisfies the equation above for the $l$-th pair of $(\vx_i, \vx_j)$, where $i, \,j\leq T$. Setting 
\[
M = \max_{l \leq \binom{2}{T}} M_l
\]
will ensure that each two elements of $\bar \mG^T \bar \mG/m$ and $\bar \mK$ are closer than $\epsilon$. The proof is concluded via the definition of Frobenius norm.
\end{proof}
%%%%%%%%%%%%%%%%%%%%%%%%%%%%%%%%%%%%%
\begin{proof}[Proof of Lemma \ref{lem:gu51_conv}]
This proof closely tracks the proof of Lemma 5.1 in \citet{zhou2019neural}. Consider Lemma \ref{lem:CNTK_gram} and let $\epsilon = \lambda_0/(2TK)$, where $\lambda_0>0$ is the smallest eigenvalue of $\bar K$. Then with probability one, we have $\norm{\bar\mG^T\bar\mG-\bar \mK}_F \leq \lambda_0/2$. Therefore,
\begin{equation}\label{eq:G_posdef}
    \bar \mG^T\bar \mG \succcurlyeq \bar \mK - \norm{\bar\mG^T\bar\mG-\bar \mK}_F\mI \succcurlyeq \bar \mK - \lambda_0\mI/2 \succcurlyeq \bar\mK/2  \succ0
\end{equation}
where the first inequality is due to the triangle inequality. This implies that $\bar G$ is also positive definite. Suppose $\bar \mG = \mP\mA\mQ^T$, with $\mA \succ 0$. Setting $\vtheta^* = \mP \mA^{-1}\mQ^T\vf^*$ satisfies the equation in the lemma, where $\vf^* = [f^*(\vx^i)]_{i\leq T|\gA|}$. Moreover, due to Equation \ref{eq:G_posdef}, 
\[
m \norm{\vtheta^*}_2^2 = (\vf^*)^T\mQ\mA^{-2}\mQ \vf^* =  (\vf^*)^T(\bar\mG^T\bar\mG)^{-1} \vf^* \leq 2 \vf^* \bar\mK^{-1} \vf^* = 2 \norm{f^*}_\cntk^2.
\]
By the assumption on reward, $\norm{f^*}_\cntk \leq B$, completing the proof.
\end{proof}
%%%%%%%%%%%%%%%%%%%%%%%%%%%%%%%%%%%%%
\begin{proof}[Proof of Lemma \ref{lem:mu_to_f_conv}] It suffices to show that Lemmas \ref{lem:f_with_gtheta}, \ref{lem:gd_conv_ridge}, and \ref{lem:thetatilde_theta} hold for the convolutional variant, and the proof follows by repeating the steps taken in proof of Lemma \ref{lem:mu_to_f}. 
We start by showing that lemma \ref{lem:gd_conv_ridge} applies to a CNN. Consider the convolutional variant of the auxiliary loss,
\begin{equation} 
    \bar \Ls_2(\vtheta) = \norm{\bar\mG^T(\bar\vtheta-\vtheta^0) - \vy}_2^2 + m \sigma^2 \norm{\bar\vtheta-\vtheta^0}_2^2.
\end{equation}
Let $(\tilde{\bar \vtheta}^j)$ to denote the gradient descent updates. The loss $\bar \Ls_2$ is strongly convex, which allows us to repeat the proof for the convolutional equivalent of the parameters. We conclude that Lemma \ref{lem:gd_conv_ridge} still holds.

By Lemma 4.1 \citet{cao2019generalization}, if $\norm{\bar\vtheta^{(J)}-\vtheta^0} \leq \tau$, then,
\[
\abs{\bar f(\vx^i; \bar\vtheta^J)-\bar f(\vx^i; \bar\vtheta^J)-\langle \bar\vg(\vx^i; \vtheta^0), \bar\vtheta^J - \vtheta^0 \rangle} \leq C \tau^{4/3} L^3 \sqrt{m\log m}.
\]
To show that Lemma \ref{lem:f_with_gtheta} can be applied to \cnnalg, it remains to show that $\norm{\bar\vtheta^{(J)}-\vtheta^0} \leq 2\sqrt{TB/(m\sigma^2)}$. Recall that $\bar \vg(\vx; \vtheta^0)$ is equal to $\vg(c_l \cdot \vx; \vtheta^0)$ averaged over all $c_l$. Therefore all inequalities that bound a norm of $\vg(\vx; \vtheta^0)$ uniformly for all $\vx \in \mathcal{X}$, carry over to $\bar \vg(\vx; \vtheta^0)$. It is straightforward to show that Lemma B.2 from \citet{zhou2019neural} also holds in the convolutional case, which completes the proof for convolutional variant of Lemma \ref{lem:f_with_gtheta}. From the triangle inequality, and by Lemma \ref{lem:gd_conv_ridge} we also get
\[
\norm{\bar \vtheta^J - \tilde{\bar \vtheta}^J}_2 \leq 3 \sqrt{\frac{TB}{m\sigma^2}}
\]
which proves Lemma \ref{lem:thetatilde_theta} for the convolutional setting. 
\end{proof}
%%%%%%%%%%%%%%%%%%%%%%%%%%%%%%%%%%%%%
%%%%%%%%%%%%%%%%%%%%%%%%%%%%%%%%%%%%%
\begin{proof}[Proof of Lemma \ref{lem:neuralvalko1_conv}]
We may repeat the proof for Lemma \ref{lem:neuralvalko1} and use Lemma \ref{lem:gu51_conv} and \ref{lem:mu_to_f_conv} when needed, in place of Lemmas \ref{lem:gu51} and \ref{lem:mu_to_f} respectively.
\end{proof}
%%%%%%%%%%%%%%%%%%%%%%%%%%%%%%%%%%%%%
\begin{proof}[Proof of Lemma \ref{lem:gammahat_conv}]
\looseness -1 We may repeat the recipe of the proof to Lemma \ref{lem:gammahat}. To make it applicable to the CNTK, one step has to be modified, and that is inequality (d) of Equation \ref{eq:Ihat_gamma}, which still holds due to Lemma \ref{lem:CNTK_gram}.
\end{proof}

\end{document}